\documentclass[acmtog]{acmart}

\copyrightyear{2024}
\acmYear{2024}
\setcopyright{acmlicensed}\acmConference[SIGGRAPH Conference Papers '24]{Special Interest Group on Computer Graphics and Interactive Techniques Conference Conference Papers '24}{July 27-August 1, 2024}{Denver, CO, USA}
\acmBooktitle{Special Interest Group on Computer Graphics and Interactive Techniques Conference Conference Papers '24 (SIGGRAPH Conference Papers '24), July 27-August 1, 2024, Denver, CO, USA}
\acmDOI{10.1145/3641519.3657526}
\acmISBN{979-8-4007-0525-0/24/07}

\acmSubmissionID{1267}
\usepackage{bm}
\usepackage{enumitem}
\usepackage{multirow}
\usepackage{graphicx}
\usepackage{caption}
\usepackage{subcaption}
\usepackage[ruled,linesnumbered]{algorithm2e}
\usepackage{rotating}
\usepackage{optidef}
\usepackage{soul}
\usepackage{calc}
\usepackage{tabularx}
\usepackage{booktabs}
\usepackage{ragged2e}
\usepackage{xcolor}
\usepackage{multirow}
\usepackage{wrapfig}
\usepackage{diagbox}
\usepackage{capt-of}
\usepackage{fixmath}
\usepackage{array, multirow, bigdelim, makecell, booktabs} 

\newcommand{\method}{ReFiNe}
\newcommand{\methodspace}{ReFiNe }

\newcommand{\latent}{\boldsymbol{z}}
\newcommand{\latentDim}{D}

\newcommand{\LoD}{m}
\newcommand{\maxLoD}{M}

\newcommand{\sdfPrediction}{s}

\newcommand{\xyz}{\boldsymbol{x}}

\newcommand{\numObjects}{K}

\newcommand{\realNum}{\mathbb{R}}
\newcommand{\fieldDim}{F}
\newcommand{\datasetIdx}{j}
\newcommand{\fieldVal}{f}
\newcommand{\numSupervisionPts}{N}
\newcommand{\dataset}{\mathcal{D}}

\newcommand{\latentFused}{\boldsymbol{\bar{z}}}
\newcommand{\latentDimFused}{\bar{D}}

\newcommand{\neuralMappingColor}{\xi}
\newcommand{\neuralMappingGeometry}{ \psi}

\newcommand{\occupancyDecoder}{\omega}
\newcommand{\latentSubdivision}{\phi}

\newcommand{\allObjects}{\mathcal{O}}
\newcommand{\object}{O}

\DeclareCaptionLabelFormat{andtable}{#1~#2  \&  \tablename~\thetable}
\begin{document}

\title{ReFiNe: Recursive Field Networks for Cross-Modal Multi-Scene Representation}

\author{Sergey Zakharov}
\affiliation{%
  \institution{Toyota Research Institute}
  \city{Los Altos}
  \country{USA}
}
\email{sergey.zakharov@tri.global}

\author{Katherine Liu}
\affiliation{%
  \institution{Toyota Research Institute}
  \city{Los Altos}
  \country{USA}
}
\email{katherine.liu@tri.global}

\author{Adrien Gaidon}
\affiliation{%
  \institution{Toyota Research Institute}
  \city{Los Altos}
  \country{USA}
}
\email{adrien.gaidon.ctr@tri.global}

\author{Rareș Ambruș}
\affiliation{%
  \institution{Toyota Research Institute}
  \city{Los Altos}
  \country{USA}
}
\email{rares.ambrus@tri.global}




\begin{abstract}
The common trade-offs of state-of-the-art methods for multi-shape representation (a single model "packing" multiple objects) involve trading modeling accuracy against memory and storage. We show how to encode multiple shapes represented as continuous neural fields with a higher degree of precision than previously possible and with low memory usage. Key to our approach is a recursive hierarchical formulation that exploits object self-similarity, leading to a highly compressed and efficient shape latent space. Thanks to the recursive formulation, our method supports spatial and global-to-local latent feature fusion without needing to initialize and maintain auxiliary data structures, while still allowing for continuous field queries to enable applications such as raytracing.
In experiments on a set of diverse datasets, we provide compelling qualitative results and demonstrate state-of-the-art multi-scene reconstruction and compression results with a single network per dataset.
Project page: 
\href{https://zakharos.github.io/projects/refine/}{\color{magenta}{https://zakharos.github.io/projects/refine/}}
\end{abstract}

\begin{teaserfigure}
    \centering
    \includegraphics[width=1\linewidth]{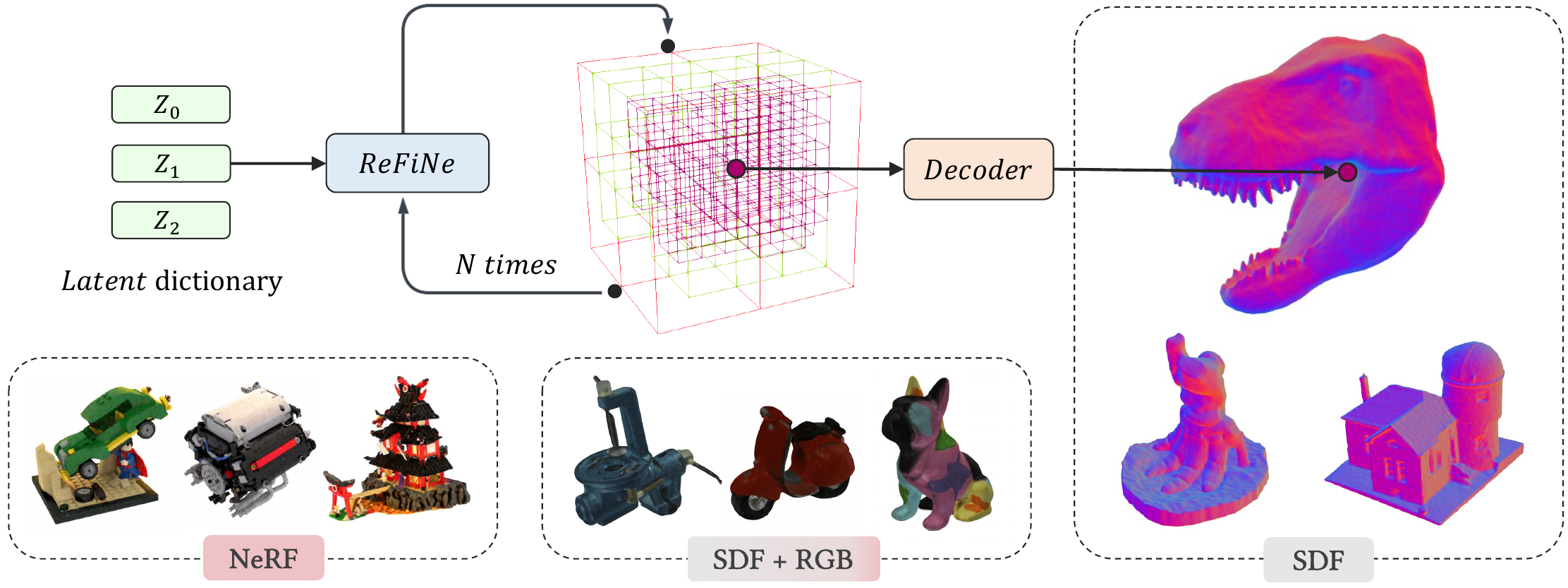}
    \caption{In this work, we propose Recursive Field Networks 
   (\textbf{\method}), a method for accurately representing a set of diverse 3D assets represented as fields within a single lightweight neural network. \textbf{\method} learns a hierarchically structured implicit shape representation that supports various output 3D geometry and color parameterizations. Shown here are example reconstructions on three different pairs of datasets and object representations: Thingi32 represented as SDFs, HomebrewedDB as colored SDFs, and RTMV as NeRFs.}
    \label{fig:teaser}
\end{teaserfigure}

\maketitle
\section{Introduction}
\label{sec:intro}

Neural fields that encode scene properties at arbitrary resolutions using neural networks have reached unprecedented levels of detail. Typically using fully-connected multi-layer perceptrons (MLPs) to predict continuous field values, they have been used to represent geometry and appearance with applications in computer vision~\cite{tancik2022block}, robotics~\cite{rashid2023language}, and computer graphics~\cite{graphics}. However, most high-fidelity methods are limited to single scenes~\cite{muller2022instant, takikawa2021neural, takikawa2022variable} and overfit to the target geometry or appearance~\cite{muller2022instant}, while methods that capture multiple shapes typically sacrifice high-frequency details~\cite{jang2021codenerf, park2019deepsdf, mescheder2019occupancy}, limiting utility for applications such as streaming and representation learning. We would like to enable the compression of multiple complex shapes into single vectors with a single neural network and while maintaining the ability reconstruct high frequency geometric and textural information. 

Global conditioning methods~\cite{sitzmann2019scene,jang2021codenerf, park2019deepsdf} (i.e. one latent vector per shape) are capable of learning latent spaces over large numbers of shapes but require ground truth 3D supervision and suffer when representing high frequency details. Conversely, locally-conditioned methods partition the implicit function by leveraging hybrid discrete-continuous neural scene representations, effectively blurring the line between classical data structures and neural representations and allowing for more precise reconstructions by handling scenes as collections of localized primitives. These methods typically encode single scenes and leverage a secondary data structure~\cite{takikawa2021neural,zakharov2022road,muller2022instant}, trading off additional memory for a less complex neural function mapping feature vectors to the target signal. Recently,~\cite{zakharov2022road} proposed to take advantage of both global and local conditioning via a recursive octree formulation, but the approach only captures geometry and outputs oriented point clouds that do not allow for continuous querying of the underlying implicit function, precluding the application of techniques such as ray-tracing.

In this work, we propose to encode many scenes represented as fields in a single network, where each scene is denoted by a single latent vector in a high dimensional space. We show how entire datasets of colored shapes can be encoded into a single neural network without sacrificing high frequency details (color or geometry) and without incurring a high memory cost. Key to our approach is a recursive formulation that allows us to effectively combine local and global conditioning. Our main motivation for a recursive structure comes from the observation that natural objects are self-similar~\cite{shechtman2007matching}, that is they are similar to a part of themselves at different scales. This property is famously used in the Fractal compression methods~\cite{jacquin1990fractal}. Our method effectively extends prior work to the continuous setting, which allows us to recover geometry and color information with a higher degree of fidelity than previously possible. Our novel formulation allows us to learn from direct 3D supervision (SDF plus optionally RGB), as well as from continuous valued fields (NeRFs). We also investigate the properties of the resulting latent space and our results suggest the emergence of structure based on shape and appearance similarity. We address the limitations of related methods for representing multiple 3D shapes through ReFiNe: Recursive Field Networks and our contributions are:

\begin{itemize}[nosep, wide, leftmargin=*]

\item \textbf{A novel implicit representation parameterized by a recursive function} that efficiently combines global and local conditioning, allowing \textbf{continuous} spatial interpolation and \textbf{multi-scale} feature aggregation. 

\item Thanks to its recursive formulation, \methodspace scales to \textbf{multiple 3D assets represented as fields without having to maintain auxiliary data structures}, leading to a compact and efficient network structure. We demonstrate a single network representing \textbf{more than 1000 objects} with high quality and \textbf{reducing the memory needed by 99.8\%}.

\item \methodspace is \textit{cross-modal}, i.e., it supports various output 3D geometry and color representations (e.g., \textbf{SDF}, SDF+Color, and \textbf{NeRF})  and its output can be rendered either with \textbf{sphere raytracing} (SDF), \textbf{iso-surface projection} (SDF) or \textbf{volumetric rendering} (NeRF).

\end{itemize}
\section{Related Work}
\label{sec:related}

\begin{figure*}[t]
    \centering
    \includegraphics[width=1\linewidth]{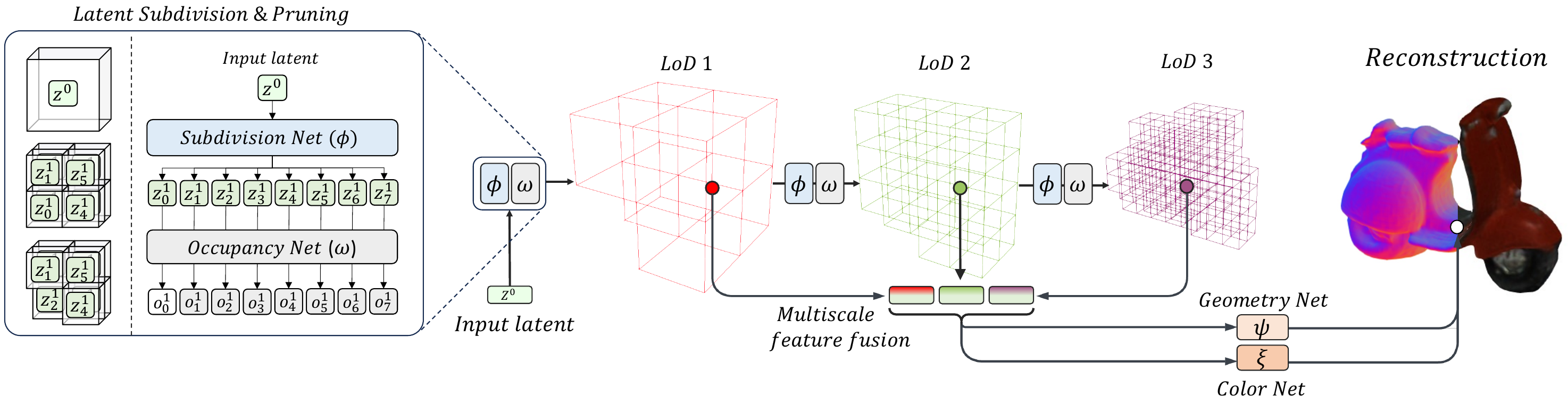}
    \caption{\textbf{\methodspace architecture}. \methodspace uses an implicit recursive hierarchical representation and a combination of spatial and global-to-local feature fusion to accurately reconstruct 3D assets. Given a single input feature corresponding to LoD 0, \methodspace recursively expands an octree to the desired LoD using the latent subdivision network $\latentSubdivision$. Unoccupied voxels at each LoD are pruned based on the output of $\occupancyDecoder$. To obtain a feature value at a specific spatial coordinate, we perform tri-linear interpolation within each individual LoD, then aggregate the features via multi-scale feature fusion. Finally, we use $\neuralMappingColor$ and $\neuralMappingGeometry$ to decode color and geometry respectively for the desired coordinate. Given the ability to query coordinates within the scene bounds, various methods including differentiable rendering can be applied for reconstruction. Importantly, \methodspace optimizes a single LoD 0 feature per 3D asset in the training dataset, enabling multiple assets to be reconstructed from a single trained \methodspace network. Voxel grids at LoDs not drawn to scale.}
    \label{fig:method}
\end{figure*}

\subsection{Neural Fields for Representing Shapes} Neural fields have emerged as powerful learners thanks to their ability to encode any continuous function up to an arbitrary level of resolution. For a survey of recent progress please refer to~\cite{xie2021neural}. Shapes are typically represented as Signed Distance Functions~\cite{park2019deepsdf,sitzmann2020metasdf,sitzmann2019siren} or by occupancy probabilities~\cite{mescheder2019occupancy,peng2020convolutional,chen2019learning}, with the encoded mesh extracted through methods such as sphere tracing~\cite{liu2020dist}. Hybrid discrete-continuous data structures have enabled encoding single objects to a very high degree of accuracy~\cite{takikawa2021neural,takikawa2022variable,muller2022instant, wang2022geometry, kim2024neuralvdb, yi2023canonical} and extensions have been proposed to model articulated~\cite{deng2019nasa,mu2021sdf} and deformable~\cite{deng2021deformed,palafox2021npms} objects. Alternatively, training on multiple shapes leads to disentangled latent spaces~\cite{park2019deepsdf,chen2019learning,tang2021octfield} which can be used for differentiable shape optimization~\cite{zakharov2021single,irshad2022shapo} shape generation~\cite{chen2019learning,yang2019pointflow,cai2020learning,zeng2022lion}, interpolation~\cite{williams2021neural} and completion~\cite{zhou20213d}. A number of methods have been proposed which continuously model and update scene geometry within the context of Simultaneous Localization and Mapping (SLAM)~\cite{sucar2021imap,ortiz2022isdf}. 
Some methods also leverage recursion to improve the reconstruction accuracy of neural fields \cite{yang2022recursive, zakharov2022road}. The recently proposed method ROAD~\cite{zakharov2022road} is most similar to ours as it also uses a recursive Octree structure and can represent the surface of multiple objects with a single network. However, it does not encode color and it outputs a discrete fixed-resolution reconstruction, making it unsuitable for applications that require volumetric rendering or ray-tracing. In contrast, ReFiNe outputs continuous feature fields that can be used to represent various continuous representations, such as (but not limited to) colored SDFs and NeRFs.

\subsection{Differentiable Rendering Advances}~\cite{kato2020differentiable,tewari2021advances} through techniques such as volume rendering~\cite{lombardi2019neural} or ray marching~\cite{niemeyer2020differentiable} have led to methods that learn to represent geometry, appearance and as well as other scene properties from image inputs and without needing direct 3D supervision. Leveraging ray marching,~\cite{sitzmann2019scene} regresses RGB colors at surface intersection allowing it to learn from multi-view images, while~\cite{niemeyer2020differentiable} couples an implicit shape representation with differentiable rendering. Building on~\cite{lombardi2019neural}, Neural Radiance Fields (NeRFs)~\cite{mildenhall2020nerf} regress density and color values along directed rays (5D coordinates) instead of of regressing SDF or RGB values at 3D coordinates. This simple and yet very convincingly effective representation boosted interest in implicit volumetric rendering and resulted in a multitude of works tackling problems from training and rendering time performance~\cite{rebain2020derf,lindell2021autoint,tancik2020learned,liu2020neural}, to covering dynamic scenes~\cite{park2020nerfies,pumarola2020d,xian2020space}, scene relighting~\cite{martinbrualla2020nerfw,bi2020neural,nerv2021}, and composition~\cite{ost2020neural,yuan2021star,niemeyer2020giraffe}. To achieve competitive results, NeRF-style methods require a large number of input views, with poor performance in the low data regime~\cite{zhang2020nerf++} which can be improved by leveraging external depth supervision~\cite{neff2021donerf,wei2021nerfingmvs,deng2022depth}. Image supervision has also been used to learn 3D object-centric models without any additional information~\cite{stelzner2021decomposing,yu2021unsupervised,sajjadi2022object}, through a combination of Slot Attention~\cite{locatello2020object} and volumetric rendering. Alternatively, a number of methods train generalizable priors over multiple scenes~\cite{sitzmann2019scene,yu2020pixelnerf,jang2021codenerf,sajjadi2022scene,guizilini2022depth}. In~\cite{jang2021codenerf} the authors learn a prior over objects that are represented as radiance fields via MLPs and parameterized by appearance and shape codes. 
As we show through experiments, the design of our recursive neural 3D representation leads to a latent space that promotes reusability of color and geometric primitives across shapes, enabling higher accuracy recostructions than previously possible.
\section{Methodology}
\label{sec:method}

\begin{figure*}[t]
	\centering
	\includegraphics[width=1\linewidth]{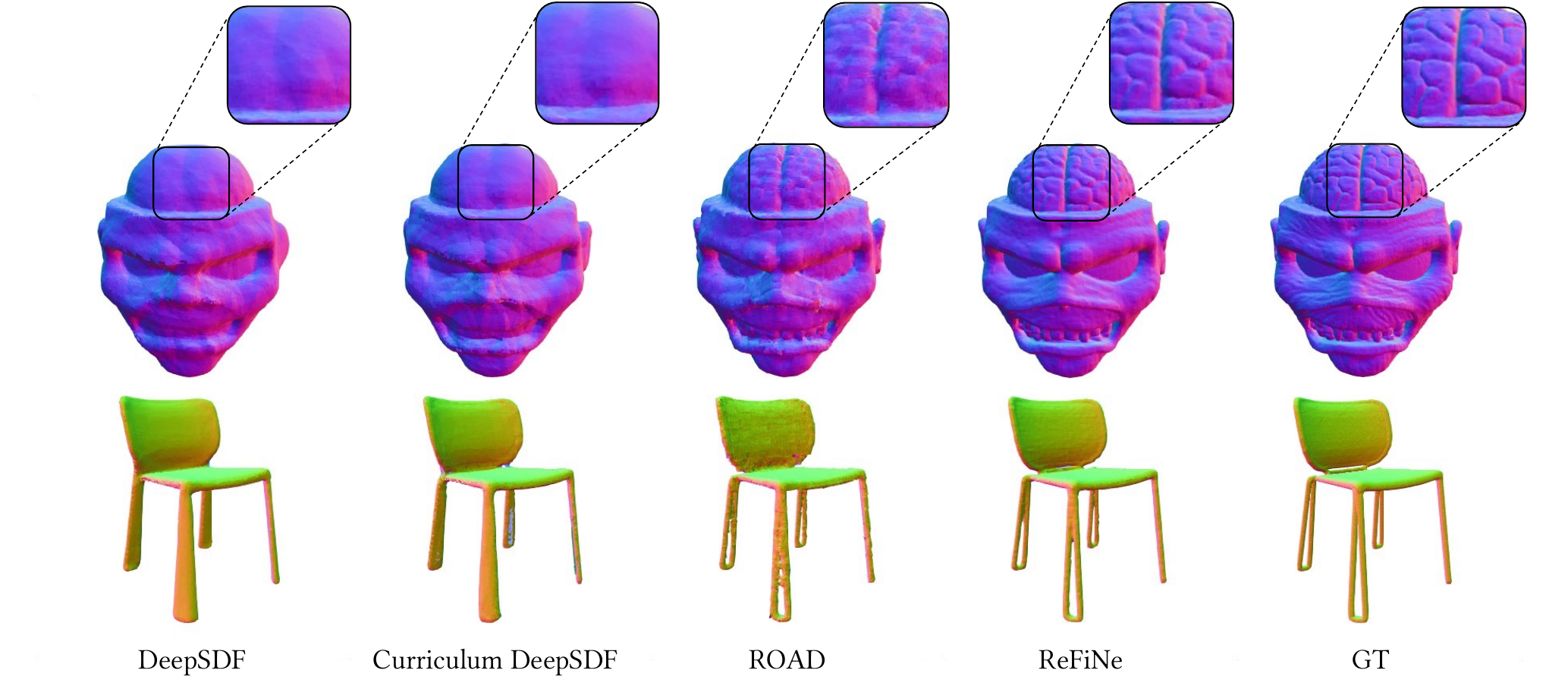}
	\caption{\textbf{SDF reconstruction comparisons on selected Thingi32 and ShapeNet150 objects}. DeepSDF and Curriculum DeepSDF capture the high-level geometry of visualized objects but fail to accurately model high frequency details such as the teeth in the top row and the chair legs in the bottom row. In this example, we also observe that ReFiNe is capable of representing geometry accurately while preserving overall shape smoothness better than ROAD (i.e., as seen in the teeth in the top row). Unlike ROAD, which also models objects recursively but discretely at each LoD, ReFiNe models objects as a continuous fields using multi-scale feature interpolation. In this visualization, ROAD and ReFiNe use nine and six LoDs, respectively. Quantitative results reported in Table \ref{tab:sdf}.}
	\label{fig:qualitative}
\end{figure*}

We would like to learn to represent a set of objects $\allObjects=\{\object_1, \dots, \object_\numObjects\}$. In particular, we are interested in representing objects as fields, where each object is a mapping from a 3D coordinate in space to a value of dimension $\fieldDim$, i.e., $\object_k:\realNum^3 \to \realNum^\fieldDim$. Examples of common fields are Signed Distance Fields (where $\fieldDim = 1$ and the value of the field indicates the distance to the nearest surface) and radiance fields (where $\fieldDim = 4$, representing RGB and density values). For each object, we assume supervision in the form of $\numSupervisionPts_k$ coordinate and field value tuples of $\{\xyz_\datasetIdx, \fieldVal_\datasetIdx \}_{\datasetIdx=0}^{\numSupervisionPts_k}$, where $\xyz \in \realNum^3$ and $\fieldVal \in \realNum^\fieldDim$ is the field value.

\subsection{ReFiNe}
Our method represents each shape $\object_k$ with a $\latentDim$-dimensional latent vector $\latent^0$ that is recursively expanded into an octree with a maximum Level-of-Detail (LoD) $\maxLoD$. Each level of the octree corresponds to a feature volume. We then perform both spatial and hierarchical feature aggregations before decoding into field values. Crucially, the expansion of each latent vector into an Octree-based neural field is achieved via the same simple MLP for each LoD and decoders are shared across all objects in $\allObjects$. Once optimized, ReFiNe represents all $\numObjects$ objects in a set of $\numObjects$ latent vectors, a recursive autodecoder for octree expansion, an occupancy prediction network, and field-specific decoders (i.e., for RGB, SDF, etc). Figure~\ref{fig:teaser} illustrates how after training, our method can extract neural fields given different optimized LoD 0 latents, where we have dropped the superscipt for readability. Figure~\ref{fig:method} shows a more detailed overview of a reconstruction given a single input latent.

\subsubsection{Recursive Subdivision \& Pruning} 
Given a latent vector $\latent^\LoD\in\mathbb{R}^\latentDim$ from LoD $\LoD$, our recursive autodecoder subdivision network $\latentSubdivision:\mathbb{R}^{\latentDim} \to \mathbb{R}^{8\latentDim}$ traverses an octree by latent subdivision:

\begin{equation}
    \latentSubdivision(\latent^\LoD) \to \{\latent^{\LoD+1}_i\}_{i=0}^{7}
\label{eq:phi}
\end{equation} 

Thus, a latent is divided into 8 cells, each with an associated child latent that is positioned at the cell's center. Cell locations are defined by the Morton space-filling curve~\cite{morton1966computer}.

Each child latent is then further decoded to occupancy values $o$ using occupancy network $\occupancyDecoder:\mathbb{R}^\latentDim\to\mathbb{R}^1$. Rather than continuing to expand the tree for all child latents, $\methodspace$ selects a subset based on the predicted occupancy value:

\begin{equation}
\label{eq:recursion}
    \mathcal{Z}^{m+1} = \{\latent^{m+1} \in \latentSubdivision(\latent^\LoD) \mid \occupancyDecoder(\latent^{m+1}) > 0.5\},
\end{equation}

where $\mathcal{Z}^{m+1}$ is the set of children latents from a particular parent latent $\latent^m$ having predicted occupancies above a threshold of $0.5$, from which the next set of children will be recursed. This process can be seen in the left inset of Fig.~\ref{fig:method}.
\noindent To supervise occupancy predictions, we further assume access to the structure of the ground-truth octree during training, i.e., annotations of which voxels at each LoD are occupied. If a voxel is predicted to be more likely unoccupied during reconstruction, we prune it from the octree structure. 

To build the set of latents at a particular LoD, the latent expansion process described by Equations \ref{eq:phi} and \ref{eq:recursion} for a single latent is applied to all unpruned children latents from the previous LoD. In this way, $\methodspace$ recursively expands a latent octree from a single root latent $\latent^0$ to a set of latents at the desired LoD.

\subsubsection{Multiscale Feature Fusion}
Once an octree is constructed, it can be decoded to various outputs depending on the desired field parametrization. As mentioned, we use $\occupancyDecoder$ and decode each recursively extracted latent vector to occupancy. However, to model more complex signals with high-frequency details (e.g. SDF or RGB) we found that directly decoding latents positioned at voxel centers results in coarse approximations at low octree LoDs and is directly tied to the voxel size, presenting challenges in scaling to high resolutions and/or complex scenes. 
Instead, we approximate latents at sampled locations by performing trilinear interpolation given spatially surrounding latents at the same LoD. We repeat this at every LoD except the first and then fuse resulting intermediate latents as shown in Fig.~\ref{fig:method} into a new latent $\latentFused \in \realNum^{\latentDimFused}$, where the dimension ${\latentDimFused}$ of the fused latent varies based on whether a concatenation or summation scheme is used. In the summation scheme, the latent size remains unchanged, i.e., $\latentDimFused = \latentDim$, whereas in the concatenation scheme, it is equal to the original latent size $\latentDim$ multiplied by the maximum LoD $\maxLoD$. 

\begin{figure*}[t]
	\centering
	\includegraphics[width=1\linewidth]{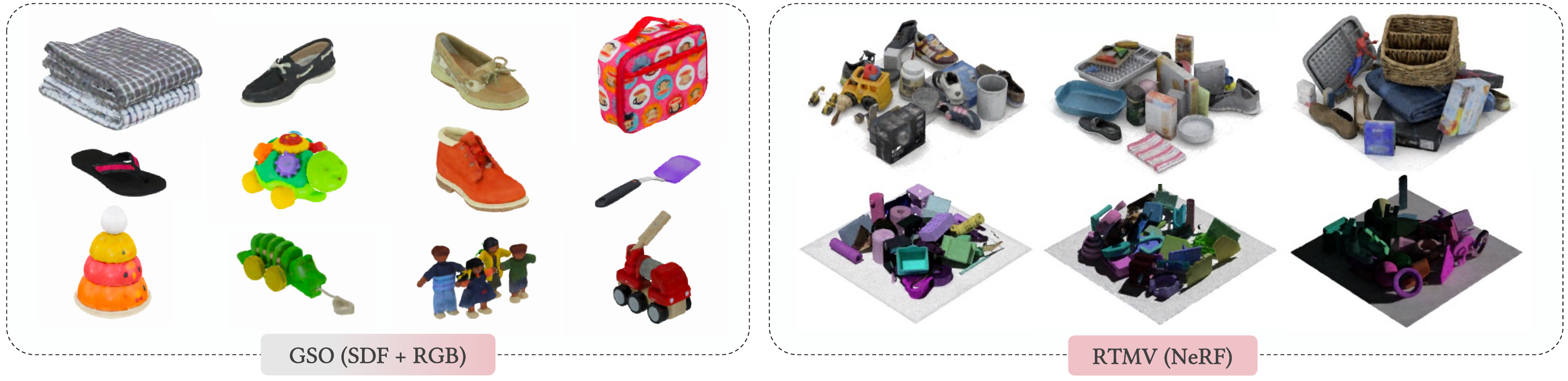}
	\caption{\textbf{Decoded Datasets}. \methodspace can encode Google Scanned Objects, a complex dataset of 1030 colored 3D objects within a single neural network of size 45.6 MB and a list of latent vectors of 1.05 MB (whereas the original meshes without texture require about 1.5 GB of storage). On the right, we show complex decoded reconstructions from our network trained on the RTMV dataset of 40 diverse scenes.}
	\label{fig:qualitative_sdf_color}
\end{figure*}

\subsubsection{Geometry Extraction and Rendering}
Similar to \cite{zakharov2022road}, once the feature octree has been extracted for a given object we can decode the voxel centers into field values. However, our resulting representation can also be used to differentiably render images via volumetric rendering. We first estimate AABB intersections with voxels at the highest LoD. Given enter and exit points for each voxel, we then sample points within the voxel volume, enabling rendering via methods such as sphere ray tracing and volumetric compositing.

\subsection{Field Specific Details}
To demonstrate the utility and flexibility of \method, we focus in this work on two popular choices of object fields: Signed Distance Fields (SDF)~\cite{park2019deepsdf} for representing surfaces and Neural Radiance Fields (NeRF)~\cite{mildenhall2020nerf} for volumetric rendering and view synthesis. \methodspace regresses field specific signals via neural mappings that map regressed latents, and optionally viewing direction to the desired output (e.g. SDF, SDF and RGB or density and RGB). We denote the neural mapping responsible for geometry as $\neuralMappingGeometry$ and the neural mapping responsible for appearance as $\neuralMappingColor$ and discuss specific instantiations below.

\subsubsection{SDF} Each fused latent $\latentFused$ regressed via spatial interpolation over the octree and fused over multiple LoDs is given to network $\neuralMappingGeometry:\mathbb{R}^{\latentDimFused}\to\mathbb{R}^1$ to estimate an SDF value $\sdfPrediction$ corresponding to a distance to the closest surface, with positive and negative values representing exterior and interior areas respectively. When dealing with colored objects, we introduce network $\neuralMappingColor:\mathbb{R}^{\latentDimFused}\to\mathbb{R}^3$ to estimate a 3D vector $\textbf{c}=(r,g,b)$ that represents RGB colors.

To quickly extract points on the surface of the object, we can simply decode $\sdfPrediction$ for the coordinate of each occupied voxel at the highest LoD and calculate the normal of the point by taking the derivative w.r.t to the spatial coordinates. If more points are desired, we can additionally sample within occupied voxels to obtain more surface points. Given further computation time, we may also render the encoded scene via sphere ray tracing, i.e. at each step querying a SDF value within voxels that defines a sphere radius for the next step. We repeat the process until we reach the surface. The latents at the surface points are then used to estimate color values. Figures \ref{fig:qualitative} and \ref{fig:qualitative_sdf_color} show qualitative examples of iso-surface projection and sphere ray tracing, respectively.

\subsubsection{NeRF} When representing neural radiance fields each fused multiscale feature is given to networks $\neuralMappingColor:\mathbb{R}^{{\latentDimFused}+3}\to\mathbb{R}^3$ and $\neuralMappingGeometry:\mathbb{R}^{\latentDimFused}\to\mathbb{R}^1$ to estimate a 4D vector $(\textbf{c},\sigma)$, where $\textbf{c}=(r,g,b)$ are RGB colors and $\sigma$ are densities per point. When trained on NeRF, our color network additionally takes a 3-channel view direction vector $d$ and the corresponding annotation $\dataset$ is augmented accordingly.

\begin{figure*}[t]
	\centering
	\includegraphics[width=1\linewidth]{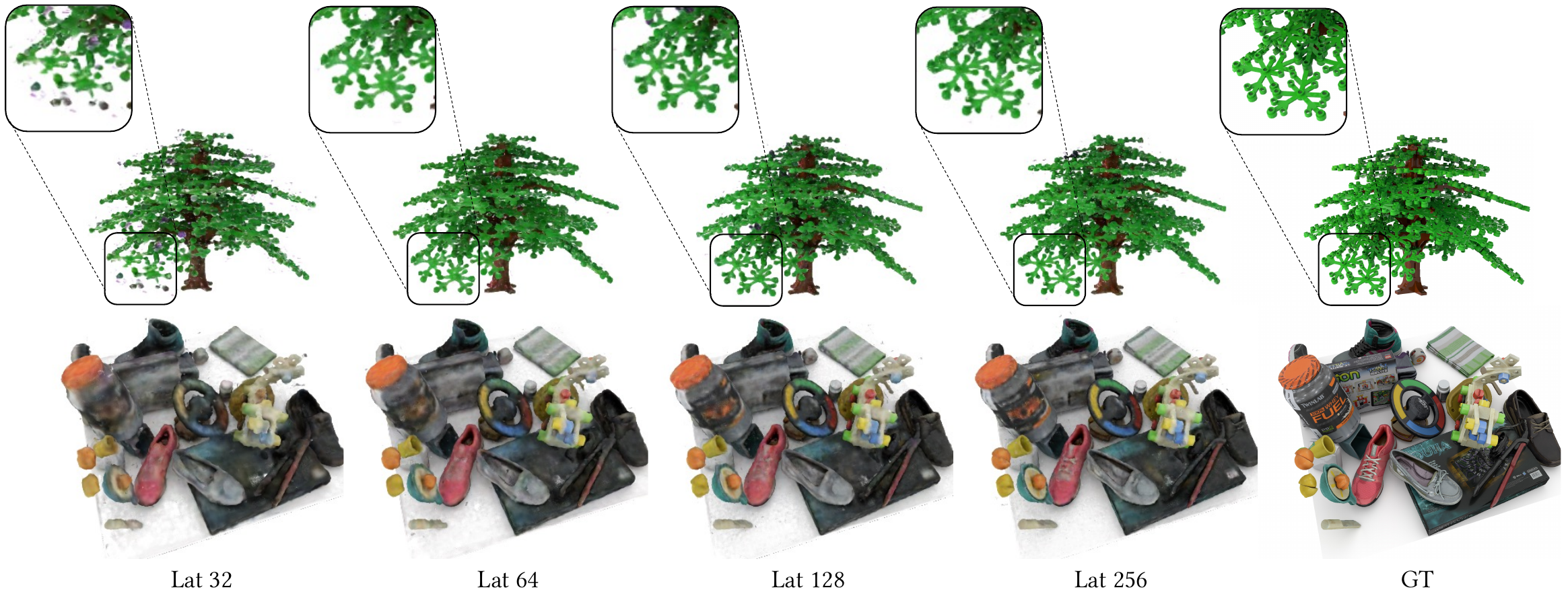}
	\caption{\textbf{Qualitative RTMV ablation results}. We observe that increasing the latent size increases the capacity of \methodspace for high fidelity reconstruction, and that more complex scenarios such as the cluttered scene in the bottom row may benefit more from larger latent spaces. Both objects are rendered from the same network. Quantitative results reported in Table \ref{tab:rtmv}.}
	\label{fig:rtmv_ablation}
\end{figure*}

To render an image, each pixel value in the desired image frame is generated by compositing $K$ color predictions along the viewing ray via:
\begin{equation}
\small
\hat{\textbf{c}}_{ij} = \sum_{k=1}^K w_k\hat{\textbf{c}}_k,
\label{eq:vol_pred}
\end{equation}
where weights $w_k$ and accumulated densities $T_k$, provided intervals $\delta_k = t_{k+1} - t_k$, are defined as follows:
\begin{align}
\small
w_k &= T_k \Big( 1-\exp(-\sigma_k\delta_k)
\Big) 
\\
T_k &= \exp \Big(
-\sum_{k'=1}^K \sigma_{k'}\delta_{k'} 
\Big)
\end{align}
and $\{t_k\}_{k=0}^{K-1}$ are sampled adaptive depth values. Example visualizations of NeRF-based volumetric rendering can be seen in Fig.~\ref{fig:rtmv_ablation}.

\subsection{Architecture and Training}
The functions $\latentSubdivision$, $\occupancyDecoder$, $\neuralMappingGeometry$,  and $\neuralMappingColor$ are parameterized with single SIREN-based~\cite{sitzmann2019siren} MLPs using periodic activation functions allowing to high-frequency details to be resolved. We refer to these components together as the \methodspace network.

Our supervision objective consists of three terms: a binary cross entropy occupancy loss $\mathcal{L}_{o}$, geometry loss $\mathcal{L}_{g}$ and color loss $\mathcal{L}_{c}$ minimizing the $l_2$ distance between respective predictions and ground truth values in each object's field annotation $\dataset$.

The final loss is formulated as:
\begin{equation}
    \mathcal{L} = w_{o} \mathcal{L}_o + w_{g} \mathcal{L}_g + w_c \mathcal{L}_c,
\end{equation}
where $w_o = 2, w_g = 10, w_c = 1$ for SDF, and $w_o = 2, w_g = 1, w_c = 1$ for NeRF. The color loss value is dropped entirely when training on purely geometric SDFs. During training, we optimize the parameters of the recursive autodecoder $\latentSubdivision$, occupancy prediction network $\occupancyDecoder$, decoding networks $\neuralMappingColor, \neuralMappingGeometry$ as well as the set of $\numObjects$ LoD 0 latent variables $\latent^0_{i\in \numObjects}$, where each latent represents a single object in $\allObjects$.

All our networks are trained on a single NVIDIA A100 GPU until convergence. The convergence time varies based on the number and complexity of objects to be encoded, as well as the network's configuration. It ranges from 10 hours for smaller datasets (Thingi32 and SRN Cars) to 40 hours for larger datasets (GSO and RTMV).

\section{Experiments}
\label{sec:experiments}

To demonstrate the utility of our method, we perform experiments across a variety of datasets (Thingi32, ShapeNet150, SRN Cars, GSO, and RTMV) and field representations (SDF, SDF+RGB, and NeRF). We highlight that our method encodes entire datasets within a single neural network, and thus we aim to compare with baselines that focus on the same task and require the same kind of supervision, as opposed to methods that overfit to single shapes or scenes.

\subsection{Network Details}
For experiments on Thingi32 and ShapeNet150 \method's recursive autodecoder network $\latentSubdivision$ consists of a single 1024-dimensional layer, and all decoding networks $\occupancyDecoder$,  $\neuralMappingGeometry$ and $\neuralMappingColor$ use two-layers of 256 fully connected units each. For the SRN Cars experiment we use a smaller capacity network featuring 128 two-layer decoding networks.
For GSO and RTMV, we increase the capacity of the \methodspace network, such that $\latentSubdivision$ consists of a single 4096-dimensional layer, and all decoding networks use two-layers of 512 fully connected units each.
We use the Adam solver~\cite{kingma2014adam} with a learning rate of $2\times10^{-5}$ to optimize the weights of our networks and a learning rate of $1\times10^{-4}$ for latent vectors. In general, when reporting network sizes we do not include the storage cost of latent vectors.

Throughout the experiments, we employ either concatenation (Tables~\ref{tab:sdf} and~\ref{tab:rtmv}) or summation latent fusion (Table~\ref{tab:srn}). The summation fusion scheme preserves the network size across different possible LoDs by keeping input sizes constant for decoder networks. On the other hand, the concatenation scheme comes at a higher storage cost as the corresponding decoding networks must have larger input layers, but it results in improved reconstruction quality. For an ablation comparing the fusion schemes, please refer to the supplemental material.

\subsection{Training Data Generation}
For object datasets represented as meshes, we normalize meshes to a unit sphere and additionally scale by a factor of 0.9. We first generate an octree of a desired LoD covering the mesh. We then perform dilation to secure a sufficient feature margin for trilinear interpolation. Finally, we sample points around the surface and compute respective SDF values. 
For colored shapes, we also sample points on the surface and store respective RGB values.

\setlength{\tabcolsep}{4pt}
\begin{table}[t]
  \caption{\textbf{Thingi32 and ShapeNet150 benchmarks}. \methodspace outperforms DeepSDF and Curriculum DeepSDF in terms of reconstruction accuracy, and for a given LoD provides reconstruction performance similar to that of ROAD at a higher LoD. \methodspace uses the concatenation latent fusion scheme.
}
\resizebox{\columnwidth}{!}{%
  \centering
    \begin{tabular}{lcccccccccc}
    \toprule
    \multirow{2}[4]{*}{\textbf{Method}} & \multicolumn{5}{c}{\textit{Thingi32}} & \multicolumn{5}{c}{\textit{ShapeNet150}} \\
\cmidrule(lr){2-6} \cmidrule(lr){7-11}         & CD$\downarrow$ & NC$\uparrow$ & gIoU$\uparrow$ & s$\downarrow$ & MB$\downarrow$ & CD$\downarrow$ & NC$\uparrow$ & gIoU$\uparrow$ & s$\downarrow$ & MB$\downarrow$ \\
    \midrule
    DeepSDF &   0.088  & 0.941  &   96.4   & 0.14 &  7.4    &   0.250    & 0.933 &   90.2  & 0.12  & 7.4 \\
    Curriculum DeepSDF &  0.102  & 0.941   &   96.3    & 0.14 &   7.4    &   0.214    & 0.903 &   93.3  & 0.12  & 7.4 \\
    ROAD / LoD6 & 0.138 & 0.959 & 96.4  & \textbf{0.03} & 3.2 & 0.175 & 0.928 & 86.3  & \textbf{0.01}   & 3.8 \\
    ROAD / LoD7  & 0.045 & 0.969 & 98.4   & 0.03 &  3.2     & 0.067  & 0.936 & 94.2 & 0.01  &  3.8 \\
    ROAD / LoD8  & 0.022 & 0.971 & 98.7  & 0.04 &  3.2    & 0.041 & 0.935 & 94.9 & 0.02  & 3.8 \\
    ROAD / LoD9  & \textbf{0.017} & 0.970 & 98.7   & 0.08  & 3.2 & 0.036 & 0.931 & 94.9  & 0.06  & 3.8 \\
    \midrule
    \methodspace / LoD4  &   0.023  & 0.980  &  98.8     & 0.07 & \textbf{3.1}    &   0.041 & 0.945   &  96.6   & 0.04  &  \textbf{3.7} \\
    \methodspace / LoD5  &   0.022  & \textbf{0.981}  & 99.1     & 0.07 & 3.1    &   0.036  & 0.944 &  96.5   & 0.05  &  3.8 \\
    \methodspace / LoD6  &   0.019  & \textbf{0.981}  & \textbf{99.4}   &  0.07  &   3.2    &   \textbf{0.027}  & \textbf{0.954} &  \textbf{97.4}    & 0.05  &  3.8 \\
    \bottomrule
    \end{tabular}%
    }
    \label{tab:sdf}%
\end{table}%

To efficiently train \methodspace on NeRFs, we first overfit single-scene NeRFs~\cite{muller2022instant} on separate scenes. 
Each neural field can be constructed from a collection of RGB images $\{I_i\}_{i=0}^{N-1}$, where camera intrinsic parameters $\textbf{K}_i \in \mathbb{R}^{3\times 3}$ as well as extrinsics $\mathbb{R}^{4 \times 4}$ are assumed to be known. 
If ground truth depth maps are provided (RTMV), then the octree structure for each scene is computed and subsequently used to supervise our recursive autodecoder $\phi$. If depth maps are not available (SRN Cars), we instead use adaptive pruning as implemented in~\cite{muller2022instant}. Then, we also densely sample points augmented with viewpoints inside the octree to store groundtruth density and color values for later supervision of geometry network $\neuralMappingGeometry$ and color network $\neuralMappingColor$.

\subsection{Reconstruction Benchmarks}
\subsubsection{Thingi32 / ShapeNet150 (SDF)}
In the first benchmark we evaluate our method's ability to represent and reconstruct object surfaces in the form of a SDF. We follow the experimental setup of~\cite{takikawa2021neural,zakharov2022road} and train two networks: one on a subset of 32 objects from Thingi10K~\cite{zhou2016thingi10k} denoted Thingi32, and another on a subset of 150 objects from ShapeNet~\cite{chang2015shapenet} denoted ShapeNet150. We use a latent dimension of 64 for Thingi32 and a latent dimension of 80 for ShapeNet150. We compute the commonly used Chamfer (CD), gIoU, and normal consistency (NC) metrics to evaluate surface reconstruction and we also record a memory footprint and inference time for each baseline. To extract a pointcloud from \method, we utilize the zero isosurface projection discussed in Section \ref{sec:method}. Following ROAD's~\cite{zakharov2022road} setup, gIoU is computed by recovering the object mesh using Poisson surface reconstruction~\cite{kazhdan2006poisson}. We compare to DeepSDF~\cite{park2019deepsdf}, Curriculum DeepSDF~\cite{duan2020curriculum}, using both methods' open-sourced implementations for data generation and training with some minor hyper-parameter tuning to improve performance. Further details can be found in the supplemental material.

\begin{figure}[b]
	\centering	\includegraphics[width=1\linewidth]{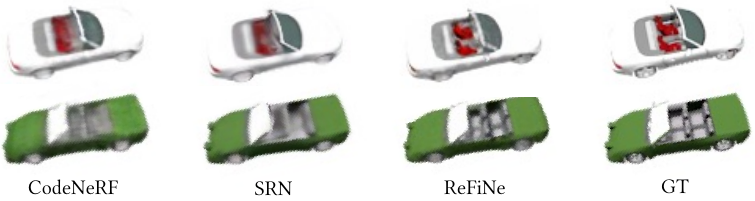}
    \caption{
	\textbf{SRN Cars benchmark}. Our method outperforms other methods at reconstructing high-frequency details. 
 \label{fig:srn}}
\end{figure}
\setlength{\tabcolsep}{14pt}
\begin{table}[t]
 \caption{\textbf{SRN Cars benchmark}. \methodspace outperforms baselines at reconstructing high-frequency details. The summation latent fusion scheme is used preserving the net size across different LoDs.}
\resizebox{\columnwidth}{!}{
  \centering
    \begin{tabular}{lccccc}
    \toprule
    \multicolumn{1}{l}{\multirow{2}[4]{*}{\textbf{Method}}} & \multicolumn{3}{c}{\textit{View Synthesis}} & \textit{Runtime} &\textit{Size} \\
\cmidrule(lr){2-4} \cmidrule(lr){5-5} \cmidrule(lr){6-6}       & PSNR$\uparrow$  & SSIM$\uparrow$ & LPIPS$\downarrow$ & s$\downarrow$ & MB$\downarrow$ \\
    \midrule
    SRN   &   28.02    &   0.95  & \textbf{0.06} & \textbf{0.03} & 198 \\
    CodeNeRF &   27.87    &   0.95  & 0.08 & 0.17  & 2.8 \\
    \midrule
    ReFiNe / LoD4 &   28.19    &   0.95  & 0.08  & \textbf{0.03} & \textbf{2.6} \\
    ReFiNe / LoD5 &    29.80   &   0.96 & \textbf{0.06} & 0.04  & \textbf{2.6} \\
    ReFiNe / LoD6 &   \textbf{30.19}   &   \textbf{0.96} & \textbf{0.06} & 0.04  & \textbf{2.6} \\
    \bottomrule
    \end{tabular}
    }
  \label{tab:srn}
\end{table}

Our results are summarized in Table~\ref{tab:sdf} and we note that our method outperforms other SDF-based baselines with respect to Chamfer and gIoU, while having the smallest storage requirements. Figure~\ref{fig:qualitative} qualitatively shows that DeepSDF~\cite{park2019deepsdf} and Curriculum DeepSDF~\cite{duan2020curriculum} have difficulties reconstructing high frequency details. ROAD~\cite{zakharov2022road}, on the other hand, can recover high frequency details but is discrete and outputs oriented point clouds with a fixed number of points at each level of detail. While \methodspace has an analogous recursive backbone, it also performs multi-scale spatial feature interpolation, and instead models the object as a continuous field. \methodspace outperforms ROAD on the ShapeNet150 dataset and performs on par on Thingi32 while only needing to traverse the octree to LoD6 as opposed to the expensive traversal to LoD9 for ROAD.
Additionally, we compare the average surface extraction times for all baselines. Notably, both us and ROAD are significantly faster than DeepSDF and Curriculum DeepSDF. While ROAD demonstrates faster runtimes per the same LoD, it can’t sample values continuously limiting it to the extracted discrete cell centers. ReFiNe shows competitive extraction times with ROAD and already at LoD6 it outperforms ROAD’s LoD9 thanks to the ability to sample values continuously.

\setlength{\tabcolsep}{11pt}
\begin{table}[t]
    \caption{\textbf{RTMV benchmark}. Comparison of baseline methods and \methodspace for various latent dimensions on the RTMV dataset at 400 × 400 resolution. \methodspace utilizes the concatenation latent scheme and trains to LoD 6.}
\resizebox{\columnwidth}{!}{%
  \centering
    \begin{tabular}[b]{lccccc}
    \toprule
    \multicolumn{1}{l}{\multirow{2}[4]{*}{\textbf{Method}}} & \multicolumn{3}{c}{\textit{View Synthesis}}  & \textit{Runtime} & \textit{Size} \\
\cmidrule(lr){2-4} \cmidrule(lr){5-5} \cmidrule(lr){6-6}          & PSNR$\uparrow$ & SSIM$\uparrow$ & LPIPS$\downarrow$ & s$\downarrow$ & MB$\downarrow$ \\
    \midrule
    \methodspace / Lat 32  &    24.18    &   0.83  & 0.23  & 1.19 &  8.4\\
    \methodspace / Lat 64  &  25.29    &   0.85    & 0.21  & 1.57 & 13.7  \\
    \methodspace / Lat 128  &  25.96    &   0.86   & 0.20  & 2.34
&  24.3 \\
    \methodspace / Lat 256  &  26.72    &   0.87   & 0.19 & 3.89
 &  45.6 \\
    \bottomrule
    \end{tabular}%
    }
    \label{tab:rtmv}%
  \end{table}

\begin{figure}[b]
	\centering	\includegraphics[width=0.9\linewidth]{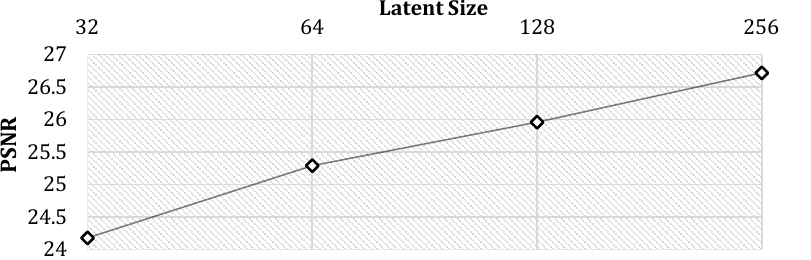}
    \caption{
	\textbf{RTMV benchmark}. Latent Size vs Reconstruction curve. 
 \label{fig:rtmv}}
\end{figure}

% \setlength{\tabcolsep}{2pt}
%   \begin{figure}[b]
%     \centering
%     \includegraphics[width=0.46\columnwidth]{figures_pdf/size_quality_lat.pdf}
%     \quad
%   \resizebox{0.46\columnwidth}{!}{%
%     \begin{tabular}[b]{lccccc}
%     \toprule
%     \multicolumn{1}{l}{\multirow{2}[4]{*}{\textbf{Method}}} & \multicolumn{3}{c}{\textit{View Synthesis}}  & \textit{Runtime} & \textit{Size} \\
% \cmidrule(lr){2-4} \cmidrule(lr){5-5} \cmidrule(lr){6-6}          & PSNR$\uparrow$ & SSIM$\uparrow$ & LPIPS$\downarrow$ & s$\downarrow$ & MB$\downarrow$ \\
%     \midrule
%     \methodspace / Lat 32  &    24.18    &   0.83  & 0.23  & 1.19 &  8.4\\
%     \methodspace / Lat 64  &  25.29    &   0.85    & 0.21  & 1.57 & 13.7  \\
%     \methodspace / Lat 128  &  25.96    &   0.86   & 0.20  & 2.34
% &  24.3 \\
%     \methodspace / Lat 256  &  26.72    &   0.87   & 0.19 & 3.89
%  &  45.6 \\
%     \bottomrule
%     \end{tabular}%
%     }
%     \captionlistentry[table]{}
%     \label{tab:rtmv}%
%     \captionsetup{labelformat=andtable}
%     \caption{Comparison of baseline methods and \methodspace for various latent dimensions on the 40-scene RTMV dataset at 400 × 400 resolution. \methodspace utilizes the concatenation latent scheme and trains to LoD 6.}
%     \label{fig:rtmv}%
%   \end{figure}

\subsubsection{SRN Cars (NeRF)}
In the next benchmark, we evaluate \methodspace on another popular representation - Neural Radiance Fields (NeRFs). We use a feature dimension of 64 and compare our method against CodeNeRF~\cite{jang2021codenerf} and SRN~\cite{sitzmann2019scene} on a subset of the SRN dataset consisting of 32 cars. We use 45 images for training and 5 non-overlapping images for testing on the task of novel view synthesis. As seen in Table~\ref{tab:srn}, our representation outperforms both SRN and CodeNeRF baselines. Fig.~\ref{fig:srn} shows that \methodspace does better when it comes to reconstructing high-frequency details. 
To compare inference time for NeRF-based baselines, we compute the average rendering time over the test images of the SRN benchmark. Our method demonstrates runtimes similar to those of SRN, with both significantly faster than CodeNeRF.

\subsection{Scaling to Larger Datasets}
Next, we demonstrate our model's ability to scale to larger multi-modal datasets. For the experiments in this section, we use a latent size of 256.

\subsubsection{GSO (SDF+RGB)}
In the first experiment, we train \methodspace to output a colored SDF field on the large Google Scanned Objects (GSO) dataset~\cite{downs2022google} containing 1030 diverse colored household objects targeting robotics applications. Despite the high complexity both in terms of geometry as well as color, our method achieves 0.044 Chamfer and 25.36 3D PSNR using a single network of size 45.6 MB together with a list of 256 dimensional latent vectors of 1.05 MB. Our method achieves a compression rate above $99.8\%$ compared to storing the original meshes (1.5 GB) and corresponding textures (24.2 GB). Qualitative results are shown in Fig.~\ref{fig:qualitative_sdf_color} and demonstrate the reconstruction quality of our approach. 

\subsubsection{RTMV (NeRF)}
In this experiment we want to demonstrate that our method is not limited to reconstructing objects and is able to cover diverse scenes of a much higher complexity. We evaluate \methodspace on the RTMV view synthesis benchmark~\cite{tremblay2022rtmv} which  consists of 40 scenes from 4 different environments (10 scenes each). Each scene comprises 150 unique views, with 100 views used for training, 5 views for validation, and 45 for testing.

As results in Fig.~\ref{fig:rtmv_ablation} show, \methodspace is able to faithfully reconstruct the encoded scenes while storing all of them within a single network with low storage requirements and without specifically optimizing for compression. We attribute this to the recursive nature of our method splitting scene space into primitives at each recursive step. As we show in Table~\ref{tab:rtmv}, our most lightweight network is only 8.36 MB, resulting in an average storage requirement of 210 KB per scene while still achieving an acceptable reconstruction quality of 24.2 PSNR. Similar to the SRN benchmark, we also compute the average rendering time over the test images, observing a gradual increase in runtime with larger latent sizes. Additionally we perform an ablation testing the effect of changing the latent size on the final reconstruction. We report results in Table~\ref{tab:rtmv} and Fig.~\ref{fig:rtmv} and note that performance gradually degrades when lowering the latent size, while at the same time decreasing storage requirements. 
\section{Limitations and Future Work}
Our representation is currently limited to bounded scenes. This limitation can potentially be resolved by introducing an inverted sphere scene model for backgrounds from ~\cite{zhang2020nerf++}. We would also like to leverage diffusion-based generative models to explore the task of 3D synthesis conditioned on various modalities such as text, images, and depth maps.

\begin{acks}

We would like to thank Prof. Greg Shakhnarovich for his valuable feedback and help with reviewing the draft for this paper.

\end{acks}

% Bibliography
\bibliographystyle{ACM-Reference-Format}
\bibliography{egbib}


\begin{thebibliography}{84}

%%% the Web address.

\ifx \showCODEN    \undefined \def \showCODEN     #1{\unskip}     \fi
\ifx \showDOI      \undefined \def \showDOI       #1{#1}\fi
\ifx \showISBNx    \undefined \def \showISBNx     #1{\unskip}     \fi
\ifx \showISBNxiii \undefined \def \showISBNxiii  #1{\unskip}     \fi
\ifx \showISSN     \undefined \def \showISSN      #1{\unskip}     \fi
\ifx \showLCCN     \undefined \def \showLCCN      #1{\unskip}     \fi
\ifx \shownote     \undefined \def \shownote      #1{#1}          \fi
\ifx \showarticletitle \undefined \def \showarticletitle #1{#1}   \fi
\ifx \showURL      \undefined \def \showURL       {\relax}        \fi
% invisible to TeX
\providecommand\bibfield[2]{#2}
\providecommand\bibinfo[2]{#2}
\providecommand\natexlab[1]{#1}
\providecommand\showeprint[2][]{arXiv:#2}

\bibitem[sit(2019)]%
        {sitzmann2019scene}
 \bibinfo{year}{2019}\natexlab{}.
\newblock \showarticletitle{Scene Representation Networks: Continuous 3D-Structure-Aware Neural Scene Representations}, In \bibinfo{booktitle}{Sitzmann, Vincent and Zollhoefer, Michael and Wetzstein, Gordon}.
\newblock \bibinfo{journal}{\emph{NeurIPS}}.
\newblock


\bibitem[Adamkiewicz et~al\mbox{.}(2022)]%
        {adamkiewicz2022vision}
\bibfield{author}{\bibinfo{person}{Michal Adamkiewicz}, \bibinfo{person}{Timothy Chen}, \bibinfo{person}{Adam Caccavale}, \bibinfo{person}{Rachel Gardner}, \bibinfo{person}{Preston Culbertson}, \bibinfo{person}{Jeannette Bohg}, {and} \bibinfo{person}{Mac Schwager}.} \bibinfo{year}{2022}\natexlab{}.
\newblock \showarticletitle{Vision-only robot navigation in a neural radiance world}.
\newblock \bibinfo{journal}{\emph{RA-L}} (\bibinfo{year}{2022}).
\newblock


\bibitem[Barron et~al\mbox{.}(2021)]%
        {barron2021mip}
\bibfield{author}{\bibinfo{person}{Jonathan~T Barron}, \bibinfo{person}{Ben Mildenhall}, \bibinfo{person}{Matthew Tancik}, \bibinfo{person}{Peter Hedman}, \bibinfo{person}{Ricardo Martin-Brualla}, {and} \bibinfo{person}{Pratul~P Srinivasan}.} \bibinfo{year}{2021}\natexlab{}.
\newblock \showarticletitle{Mip-nerf: A multiscale representation for anti-aliasing neural radiance fields}. In \bibinfo{booktitle}{\emph{ICCV}}.
\newblock


\bibitem[Bi et~al\mbox{.}(2020)]%
        {bi2020neural}
\bibfield{author}{\bibinfo{person}{Sai Bi}, \bibinfo{person}{Zexiang Xu}, \bibinfo{person}{Pratul Srinivasan}, \bibinfo{person}{Ben Mildenhall}, \bibinfo{person}{Kalyan Sunkavalli}, \bibinfo{person}{Milo{\v{s}} Ha{\v{s}}an}, \bibinfo{person}{Yannick Hold-Geoffroy}, \bibinfo{person}{David Kriegman}, {and} \bibinfo{person}{Ravi Ramamoorthi}.} \bibinfo{year}{2020}\natexlab{}.
\newblock \showarticletitle{Neural reflectance fields for appearance acquisition}.
\newblock \bibinfo{journal}{\emph{arXiv}} (\bibinfo{year}{2020}).
\newblock


\bibitem[Breyer et~al\mbox{.}(2021)]%
        {breyer2021volumetric}
\bibfield{author}{\bibinfo{person}{Michel Breyer}, \bibinfo{person}{Jen~Jen Chung}, \bibinfo{person}{Lionel Ott}, \bibinfo{person}{Roland Siegwart}, {and} \bibinfo{person}{Juan Nieto}.} \bibinfo{year}{2021}\natexlab{}.
\newblock \showarticletitle{Volumetric grasping network: Real-time 6 dof grasp detection in clutter}. In \bibinfo{booktitle}{\emph{CoRL}}.
\newblock


\bibitem[Cai et~al\mbox{.}(2020)]%
        {cai2020learning}
\bibfield{author}{\bibinfo{person}{Ruojin Cai}, \bibinfo{person}{Guandao Yang}, \bibinfo{person}{Hadar Averbuch-Elor}, \bibinfo{person}{Zekun Hao}, \bibinfo{person}{Serge Belongie}, \bibinfo{person}{Noah Snavely}, {and} \bibinfo{person}{Bharath Hariharan}.} \bibinfo{year}{2020}\natexlab{}.
\newblock \showarticletitle{Learning gradient fields for shape generation}. In \bibinfo{booktitle}{\emph{ECCV}}.
\newblock


\bibitem[Chang et~al\mbox{.}(2015)]%
        {chang2015shapenet}
\bibfield{author}{\bibinfo{person}{Angel~X Chang}, \bibinfo{person}{Thomas Funkhouser}, \bibinfo{person}{Leonidas Guibas}, \bibinfo{person}{Pat Hanrahan}, \bibinfo{person}{Qixing Huang}, \bibinfo{person}{Zimo Li}, \bibinfo{person}{Silvio Savarese}, \bibinfo{person}{Manolis Savva}, \bibinfo{person}{Shuran Song}, \bibinfo{person}{Hao Su}, {et~al\mbox{.}}} \bibinfo{year}{2015}\natexlab{}.
\newblock \showarticletitle{Shape{N}et: An information-rich 3D model repository}.
\newblock \bibinfo{journal}{\emph{arXiv}} (\bibinfo{year}{2015}).
\newblock


\bibitem[Chen and Zhang(2019)]%
        {chen2019learning}
\bibfield{author}{\bibinfo{person}{Zhiqin Chen} {and} \bibinfo{person}{Hao Zhang}.} \bibinfo{year}{2019}\natexlab{}.
\newblock \showarticletitle{Learning implicit fields for generative shape modeling}. In \bibinfo{booktitle}{\emph{CVPR}}.
\newblock


\bibitem[Davies et~al\mbox{.}(2020)]%
        {davies2020overfit}
\bibfield{author}{\bibinfo{person}{Thomas Davies}, \bibinfo{person}{Derek Nowrouzezahrai}, {and} \bibinfo{person}{Alec Jacobson}.} \bibinfo{year}{2020}\natexlab{}.
\newblock \showarticletitle{Overfit neural networks as a compact shape representation}.
\newblock \bibinfo{journal}{\emph{arXiv}} (\bibinfo{year}{2020}).
\newblock


\bibitem[Deng et~al\mbox{.}(2020)]%
        {deng2019nasa}
\bibfield{author}{\bibinfo{person}{Boyang Deng}, \bibinfo{person}{JP Lewis}, \bibinfo{person}{Timothy Jeruzalski}, \bibinfo{person}{Gerard Pons-Moll}, \bibinfo{person}{Geoffrey Hinton}, \bibinfo{person}{Mohammad Norouzi}, {and} \bibinfo{person}{Andrea Tagliasacchi}.} \bibinfo{year}{2020}\natexlab{}.
\newblock \showarticletitle{Neural Articulated Shape Approximation}. In \bibinfo{booktitle}{\emph{ECCV}}.
\newblock


\bibitem[Deng et~al\mbox{.}(2022)]%
        {deng2022depth}
\bibfield{author}{\bibinfo{person}{Kangle Deng}, \bibinfo{person}{Andrew Liu}, \bibinfo{person}{Jun-Yan Zhu}, {and} \bibinfo{person}{Deva Ramanan}.} \bibinfo{year}{2022}\natexlab{}.
\newblock \showarticletitle{Depth-supervised nerf: Fewer views and faster training for free}. In \bibinfo{booktitle}{\emph{CVPR}}.
\newblock


\bibitem[Deng et~al\mbox{.}(2021)]%
        {deng2021deformed}
\bibfield{author}{\bibinfo{person}{Yu Deng}, \bibinfo{person}{Jiaolong Yang}, {and} \bibinfo{person}{Xin Tong}.} \bibinfo{year}{2021}\natexlab{}.
\newblock \showarticletitle{Deformed implicit field: Modeling 3d shapes with learned dense correspondence}. In \bibinfo{booktitle}{\emph{CVPR}}.
\newblock


\bibitem[Downs et~al\mbox{.}(2022)]%
        {downs2022google}
\bibfield{author}{\bibinfo{person}{Laura Downs}, \bibinfo{person}{Anthony Francis}, \bibinfo{person}{Nate Koenig}, \bibinfo{person}{Brandon Kinman}, \bibinfo{person}{Ryan Hickman}, \bibinfo{person}{Krista Reymann}, \bibinfo{person}{Thomas~B McHugh}, {and} \bibinfo{person}{Vincent Vanhoucke}.} \bibinfo{year}{2022}\natexlab{}.
\newblock \showarticletitle{Google Scanned Objects: A High-Quality Dataset of 3D Scanned Household Items}.
\newblock \bibinfo{journal}{\emph{arXiv}} (\bibinfo{year}{2022}).
\newblock


\bibitem[Duan et~al\mbox{.}(2020)]%
        {duan2020curriculum}
\bibfield{author}{\bibinfo{person}{Yueqi Duan}, \bibinfo{person}{Haidong Zhu}, \bibinfo{person}{He Wang}, \bibinfo{person}{Li Yi}, \bibinfo{person}{Ram Nevatia}, {and} \bibinfo{person}{Leonidas~J Guibas}.} \bibinfo{year}{2020}\natexlab{}.
\newblock \showarticletitle{Curriculum deepsdf}. In \bibinfo{booktitle}{\emph{ECCV}}.
\newblock


\bibitem[Fuji~Tsang et~al\mbox{.}(2022)]%
        {KaolinLibrary}
\bibfield{author}{\bibinfo{person}{Clement Fuji~Tsang}, \bibinfo{person}{Maria Shugrina}, \bibinfo{person}{Jean~Francois Lafleche}, \bibinfo{person}{Towaki Takikawa}, \bibinfo{person}{Jiehan Wang}, \bibinfo{person}{Charles Loop}, \bibinfo{person}{Wenzheng Chen}, \bibinfo{person}{Krishna~Murthy Jatavallabhula}, \bibinfo{person}{Edward Smith}, \bibinfo{person}{Artem Rozantsev}, \bibinfo{person}{Or Perel}, \bibinfo{person}{Tianchang Shen}, \bibinfo{person}{Jun Gao}, \bibinfo{person}{Sanja Fidler}, \bibinfo{person}{Gavriel State}, \bibinfo{person}{Jason Gorski}, \bibinfo{person}{Tommy Xiang}, \bibinfo{person}{Jianing Li}, \bibinfo{person}{Michael Li}, {and} \bibinfo{person}{Rev Lebaredian}.} \bibinfo{year}{2022}\natexlab{}.
\newblock \bibinfo{title}{Kaolin: A Pytorch Library for Accelerating 3D Deep Learning Research}.
\newblock \bibinfo{howpublished}{\url{https://github.com/NVIDIAGameWorks/kaolin}}.
\newblock


\bibitem[Guizilini et~al\mbox{.}(2022)]%
        {guizilini2022depth}
\bibfield{author}{\bibinfo{person}{Vitor Guizilini}, \bibinfo{person}{Igor Vasiljevic}, \bibinfo{person}{Jiading Fang}, \bibinfo{person}{Rares Ambrus}, \bibinfo{person}{Greg Shakhnarovich}, \bibinfo{person}{Matthew~R Walter}, {and} \bibinfo{person}{Adrien Gaidon}.} \bibinfo{year}{2022}\natexlab{}.
\newblock \showarticletitle{Depth field networks for generalizable multi-view scene representation}. In \bibinfo{booktitle}{\emph{ECCV}}.
\newblock


\bibitem[Hodan et~al\mbox{.}(2018)]%
        {hodan2018bop}
\bibfield{author}{\bibinfo{person}{Tomas Hodan}, \bibinfo{person}{Frank Michel}, \bibinfo{person}{Eric Brachmann}, \bibinfo{person}{Wadim Kehl}, \bibinfo{person}{Anders GlentBuch}, \bibinfo{person}{Dirk Kraft}, \bibinfo{person}{Bertram Drost}, \bibinfo{person}{Joel Vidal}, \bibinfo{person}{Stephan Ihrke}, \bibinfo{person}{Xenophon Zabulis}, {et~al\mbox{.}}} \bibinfo{year}{2018}\natexlab{}.
\newblock \showarticletitle{Bop: Benchmark for 6d object pose estimation}. In \bibinfo{booktitle}{\emph{ECCV}}.
\newblock


\bibitem[Ichnowski et~al\mbox{.}(2022)]%
        {ichnowski2022dex}
\bibfield{author}{\bibinfo{person}{Jeffrey Ichnowski}, \bibinfo{person}{Yahav Avigal}, \bibinfo{person}{Justin Kerr}, {and} \bibinfo{person}{Ken Goldberg}.} \bibinfo{year}{2022}\natexlab{}.
\newblock \showarticletitle{Dex-NeRF: Using a Neural Radiance Field to Grasp Transparent Objects}. In \bibinfo{booktitle}{\emph{CoRL}}.
\newblock


\bibitem[Irshad et~al\mbox{.}(2022)]%
        {irshad2022shapo}
\bibfield{author}{\bibinfo{person}{Muhammad~Zubair Irshad}, \bibinfo{person}{Sergey Zakharov}, \bibinfo{person}{Rares Ambrus}, \bibinfo{person}{Thomas Kollar}, \bibinfo{person}{Zsolt Kira}, {and} \bibinfo{person}{Adrien Gaidon}.} \bibinfo{year}{2022}\natexlab{}.
\newblock \showarticletitle{ShAPO: Implicit Representations for Multi-Object Shape Appearance and Pose Optimization}. In \bibinfo{booktitle}{\emph{ECCV}}.
\newblock


\bibitem[Jacquin(1990)]%
        {jacquin1990fractal}
\bibfield{author}{\bibinfo{person}{Arnaud~E Jacquin}.} \bibinfo{year}{1990}\natexlab{}.
\newblock \showarticletitle{Fractal image coding based on a theory of iterated contractive image transformations}. In \bibinfo{booktitle}{\emph{VCIP}}.
\newblock


\bibitem[Jang and Agapito(2021)]%
        {jang2021codenerf}
\bibfield{author}{\bibinfo{person}{Wonbong Jang} {and} \bibinfo{person}{Lourdes Agapito}.} \bibinfo{year}{2021}\natexlab{}.
\newblock \showarticletitle{Codenerf: Disentangled neural radiance fields for object categories}. In \bibinfo{booktitle}{\emph{ICCV}}.
\newblock


\bibitem[Kaskman et~al\mbox{.}(2019)]%
        {kaskman2019homebreweddb}
\bibfield{author}{\bibinfo{person}{Roman Kaskman}, \bibinfo{person}{Sergey Zakharov}, \bibinfo{person}{Ivan Shugurov}, {and} \bibinfo{person}{Slobodan Ilic}.} \bibinfo{year}{2019}\natexlab{}.
\newblock \showarticletitle{Homebreweddb: Rgb-d dataset for 6d pose estimation of 3d objects}. In \bibinfo{booktitle}{\emph{ICCV Workshops}}.
\newblock


\bibitem[Kato et~al\mbox{.}(2020)]%
        {kato2020differentiable}
\bibfield{author}{\bibinfo{person}{Hiroharu Kato}, \bibinfo{person}{Deniz Beker}, \bibinfo{person}{Mihai Morariu}, \bibinfo{person}{Takahiro Ando}, \bibinfo{person}{Toru Matsuoka}, \bibinfo{person}{Wadim Kehl}, {and} \bibinfo{person}{Adrien Gaidon}.} \bibinfo{year}{2020}\natexlab{}.
\newblock \showarticletitle{Differentiable rendering: A survey}.
\newblock \bibinfo{journal}{\emph{arXiv}} (\bibinfo{year}{2020}).
\newblock


\bibitem[Kazhdan et~al\mbox{.}(2006)]%
        {kazhdan2006poisson}
\bibfield{author}{\bibinfo{person}{Michael Kazhdan}, \bibinfo{person}{Matthew Bolitho}, {and} \bibinfo{person}{Hugues Hoppe}.} \bibinfo{year}{2006}\natexlab{}.
\newblock \showarticletitle{Poisson surface reconstruction}. In \bibinfo{booktitle}{\emph{SGP}}.
\newblock


\bibitem[Kim et~al\mbox{.}(2024)]%
        {kim2024neuralvdb}
\bibfield{author}{\bibinfo{person}{Doyub Kim}, \bibinfo{person}{Minjae Lee}, {and} \bibinfo{person}{Ken Museth}.} \bibinfo{year}{2024}\natexlab{}.
\newblock \showarticletitle{Neuralvdb: High-resolution sparse volume representation using hierarchical neural networks}.
\newblock \bibinfo{journal}{\emph{TOG}} (\bibinfo{year}{2024}).
\newblock


\bibitem[Kingma and Ba(2014)]%
        {kingma2014adam}
\bibfield{author}{\bibinfo{person}{Diederik~P Kingma} {and} \bibinfo{person}{Jimmy Ba}.} \bibinfo{year}{2014}\natexlab{}.
\newblock \showarticletitle{Adam: A method for stochastic optimization}.
\newblock \bibinfo{journal}{\emph{arXiv}} (\bibinfo{year}{2014}).
\newblock


\bibitem[Lindell et~al\mbox{.}(2021)]%
        {lindell2021autoint}
\bibfield{author}{\bibinfo{person}{David~B. Lindell}, \bibinfo{person}{Julien~N.P. Martel}, {and} \bibinfo{person}{Gordon Wetzstein}.} \bibinfo{year}{2021}\natexlab{}.
\newblock \showarticletitle{AutoInt: Automatic Integration for Fast Neural Volume Rendering}. In \bibinfo{booktitle}{\emph{CVPR}}.
\newblock


\bibitem[Liu et~al\mbox{.}(2020a)]%
        {liu2020neural}
\bibfield{author}{\bibinfo{person}{Lingjie Liu}, \bibinfo{person}{Jiatao Gu}, \bibinfo{person}{Kyaw~Zaw Lin}, \bibinfo{person}{Tat-Seng Chua}, {and} \bibinfo{person}{Christian Theobalt}.} \bibinfo{year}{2020}\natexlab{a}.
\newblock \showarticletitle{Neural Sparse Voxel Fields}. In \bibinfo{booktitle}{\emph{NeurIPS}}.
\newblock


\bibitem[Liu et~al\mbox{.}(2020b)]%
        {liu2020dist}
\bibfield{author}{\bibinfo{person}{Shaohui Liu}, \bibinfo{person}{Yinda Zhang}, \bibinfo{person}{Songyou Peng}, \bibinfo{person}{Boxin Shi}, \bibinfo{person}{Marc Pollefeys}, {and} \bibinfo{person}{Zhaopeng Cui}.} \bibinfo{year}{2020}\natexlab{b}.
\newblock \showarticletitle{Dist: Rendering deep implicit signed distance function with differentiable sphere tracing}. In \bibinfo{booktitle}{\emph{CVPR}}.
\newblock


\bibitem[Locatello et~al\mbox{.}(2020)]%
        {locatello2020object}
\bibfield{author}{\bibinfo{person}{Francesco Locatello}, \bibinfo{person}{Dirk Weissenborn}, \bibinfo{person}{Thomas Unterthiner}, \bibinfo{person}{Aravindh Mahendran}, \bibinfo{person}{Georg Heigold}, \bibinfo{person}{Jakob Uszkoreit}, \bibinfo{person}{Alexey Dosovitskiy}, {and} \bibinfo{person}{Thomas Kipf}.} \bibinfo{year}{2020}\natexlab{}.
\newblock \showarticletitle{Object-centric learning with slot attention}. In \bibinfo{booktitle}{\emph{NeurIPS}}.
\newblock


\bibitem[Lombardi et~al\mbox{.}(2019)]%
        {lombardi2019neural}
\bibfield{author}{\bibinfo{person}{Stephen Lombardi}, \bibinfo{person}{Tomas Simon}, \bibinfo{person}{Jason Saragih}, \bibinfo{person}{Gabriel Schwartz}, \bibinfo{person}{Andreas Lehrmann}, {and} \bibinfo{person}{Yaser Sheikh}.} \bibinfo{year}{2019}\natexlab{}.
\newblock \showarticletitle{Neural volumes: learning dynamic renderable volumes from images}.
\newblock \bibinfo{journal}{\emph{TOG}} (\bibinfo{year}{2019}).
\newblock


\bibitem[Martin-Brualla et~al\mbox{.}(2021)]%
        {martinbrualla2020nerfw}
\bibfield{author}{\bibinfo{person}{Ricardo Martin-Brualla}, \bibinfo{person}{Noha Radwan}, \bibinfo{person}{Mehdi S.~M. Sajjadi}, \bibinfo{person}{Jonathan~T. Barron}, \bibinfo{person}{Alexey Dosovitskiy}, {and} \bibinfo{person}{Daniel Duckworth}.} \bibinfo{year}{2021}\natexlab{}.
\newblock \showarticletitle{{NeRF in the Wild: Neural Radiance Fields for Unconstrained Photo Collections}}. In \bibinfo{booktitle}{\emph{CVPR}}.
\newblock


\bibitem[Mescheder et~al\mbox{.}(2019)]%
        {mescheder2019occupancy}
\bibfield{author}{\bibinfo{person}{Lars Mescheder}, \bibinfo{person}{Michael Oechsle}, \bibinfo{person}{Michael Niemeyer}, \bibinfo{person}{Sebastian Nowozin}, {and} \bibinfo{person}{Andreas Geiger}.} \bibinfo{year}{2019}\natexlab{}.
\newblock \showarticletitle{Occupancy networks: Learning 3d reconstruction in function space}. In \bibinfo{booktitle}{\emph{CVPR}}.
\newblock


\bibitem[Mildenhall et~al\mbox{.}(2020)]%
        {mildenhall2020nerf}
\bibfield{author}{\bibinfo{person}{Ben Mildenhall}, \bibinfo{person}{Pratul~P Srinivasan}, \bibinfo{person}{Matthew Tancik}, \bibinfo{person}{Jonathan~T Barron}, \bibinfo{person}{Ravi Ramamoorthi}, {and} \bibinfo{person}{Ren Ng}.} \bibinfo{year}{2020}\natexlab{}.
\newblock \showarticletitle{Nerf: Representing scenes as neural radiance fields for view synthesis}. In \bibinfo{booktitle}{\emph{ECCV}}.
\newblock


\bibitem[Mitra et~al\mbox{.}(2019)]%
        {graphics}
\bibfield{author}{\bibinfo{person}{Niloy~J. Mitra}, \bibinfo{person}{Iasonas Kokkinos}, \bibinfo{person}{Paul Guerrero}, \bibinfo{person}{Nils Thuerey}, \bibinfo{person}{Vladimir Kim}, {and} \bibinfo{person}{Leonidas Guibas}.} \bibinfo{year}{2019}\natexlab{}.
\newblock \showarticletitle{CreativeAI: Deep Learning for Graphics}. In \bibinfo{booktitle}{\emph{SIGGRAPH 2019 Courses}}.
\newblock


\bibitem[Morton(1966)]%
        {morton1966computer}
\bibfield{author}{\bibinfo{person}{Guy~M Morton}.} \bibinfo{year}{1966}\natexlab{}.
\newblock \showarticletitle{A computer oriented geodetic data base and a new technique in file sequencing}.
\newblock  (\bibinfo{year}{1966}).
\newblock


\bibitem[Mu et~al\mbox{.}(2021)]%
        {mu2021sdf}
\bibfield{author}{\bibinfo{person}{Jiteng Mu}, \bibinfo{person}{Weichao Qiu}, \bibinfo{person}{Adam Kortylewski}, \bibinfo{person}{Alan Yuille}, \bibinfo{person}{Nuno Vasconcelos}, {and} \bibinfo{person}{Xiaolong Wang}.} \bibinfo{year}{2021}\natexlab{}.
\newblock \showarticletitle{A-sdf: Learning disentangled signed distance functions for articulated shape representation}. In \bibinfo{booktitle}{\emph{ICCV}}.
\newblock


\bibitem[M\"uller et~al\mbox{.}(2022)]%
        {muller2022instant}
\bibfield{author}{\bibinfo{person}{Thomas M\"uller}, \bibinfo{person}{Alex Evans}, \bibinfo{person}{Christoph Schied}, {and} \bibinfo{person}{Alexander Keller}.} \bibinfo{year}{2022}\natexlab{}.
\newblock \showarticletitle{Instant Neural Graphics Primitives with a Multiresolution Hash Encoding}.
\newblock \bibinfo{journal}{\emph{TOG}} (\bibinfo{year}{2022}).
\newblock


\bibitem[Neff et~al\mbox{.}(2021)]%
        {neff2021donerf}
\bibfield{author}{\bibinfo{person}{Thomas Neff}, \bibinfo{person}{Pascal Stadlbauer}, \bibinfo{person}{Mathias Parger}, \bibinfo{person}{Andreas Kurz}, \bibinfo{person}{Joerg~H Mueller}, \bibinfo{person}{Chakravarty R~Alla Chaitanya}, \bibinfo{person}{Anton Kaplanyan}, {and} \bibinfo{person}{Markus Steinberger}.} \bibinfo{year}{2021}\natexlab{}.
\newblock \showarticletitle{DONeRF: Towards Real-Time Rendering of Compact Neural Radiance Fields using Depth Oracle Networks}. In \bibinfo{booktitle}{\emph{Computer Graphics Forum}}.
\newblock


\bibitem[Niemeyer and Geiger(2021)]%
        {niemeyer2020giraffe}
\bibfield{author}{\bibinfo{person}{Michael Niemeyer} {and} \bibinfo{person}{Andreas Geiger}.} \bibinfo{year}{2021}\natexlab{}.
\newblock \showarticletitle{Giraffe: Representing scenes as compositional generative neural feature fields}. In \bibinfo{booktitle}{\emph{CVPR}}.
\newblock


\bibitem[Niemeyer et~al\mbox{.}(2020)]%
        {niemeyer2020differentiable}
\bibfield{author}{\bibinfo{person}{Michael Niemeyer}, \bibinfo{person}{Lars Mescheder}, \bibinfo{person}{Michael Oechsle}, {and} \bibinfo{person}{Andreas Geiger}.} \bibinfo{year}{2020}\natexlab{}.
\newblock \showarticletitle{Differentiable volumetric rendering: Learning implicit 3d representations without 3d supervision}. In \bibinfo{booktitle}{\emph{CVPR}}.
\newblock


\bibitem[Ortiz et~al\mbox{.}(2022)]%
        {ortiz2022isdf}
\bibfield{author}{\bibinfo{person}{Joseph Ortiz}, \bibinfo{person}{Alexander Clegg}, \bibinfo{person}{Jing Dong}, \bibinfo{person}{Edgar Sucar}, \bibinfo{person}{David Novotny}, \bibinfo{person}{Michael Zollhoefer}, {and} \bibinfo{person}{Mustafa Mukadam}.} \bibinfo{year}{2022}\natexlab{}.
\newblock \showarticletitle{iSDF: Real-Time Neural Signed Distance Fields for Robot Perception}. In \bibinfo{booktitle}{\emph{RSS}}.
\newblock


\bibitem[Ost et~al\mbox{.}(2021)]%
        {ost2020neural}
\bibfield{author}{\bibinfo{person}{Julian Ost}, \bibinfo{person}{Fahim Mannan}, \bibinfo{person}{Nils Thuerey}, \bibinfo{person}{Julian Knodt}, {and} \bibinfo{person}{Felix Heide}.} \bibinfo{year}{2021}\natexlab{}.
\newblock \showarticletitle{Neural scene graphs for dynamic scenes}. In \bibinfo{booktitle}{\emph{CVPR}}.
\newblock


\bibitem[Palafox et~al\mbox{.}(2021)]%
        {palafox2021npms}
\bibfield{author}{\bibinfo{person}{Pablo Palafox}, \bibinfo{person}{Alja{\v{z}} Bo{\v{z}}i{\v{c}}}, \bibinfo{person}{Justus Thies}, \bibinfo{person}{Matthias Nie{\ss}ner}, {and} \bibinfo{person}{Angela Dai}.} \bibinfo{year}{2021}\natexlab{}.
\newblock \showarticletitle{Npms: Neural parametric models for 3d deformable shapes}. In \bibinfo{booktitle}{\emph{ICCV}}.
\newblock


\bibitem[Park et~al\mbox{.}(2019)]%
        {park2019deepsdf}
\bibfield{author}{\bibinfo{person}{Jeong~Joon Park}, \bibinfo{person}{Peter Florence}, \bibinfo{person}{Julian Straub}, \bibinfo{person}{Richard Newcombe}, {and} \bibinfo{person}{Steven Lovegrove}.} \bibinfo{year}{2019}\natexlab{}.
\newblock \showarticletitle{DeepSDF: Learning Continuous Signed Distance Functions for Shape Representation}. In \bibinfo{booktitle}{\emph{CVPR}}.
\newblock


\bibitem[Park et~al\mbox{.}(2021)]%
        {park2020nerfies}
\bibfield{author}{\bibinfo{person}{Keunhong Park}, \bibinfo{person}{Utkarsh Sinha}, \bibinfo{person}{Jonathan~T. Barron}, \bibinfo{person}{Sofien Bouaziz}, \bibinfo{person}{Dan~B Goldman}, \bibinfo{person}{Steven~M. Seitz}, {and} \bibinfo{person}{Ricardo Martin-Brualla}.} \bibinfo{year}{2021}\natexlab{}.
\newblock \showarticletitle{Nerfies: Deformable Neural Radiance Fields}. In \bibinfo{booktitle}{\emph{ICCV}}.
\newblock


\bibitem[Peng et~al\mbox{.}(2020)]%
        {peng2020convolutional}
\bibfield{author}{\bibinfo{person}{Songyou Peng}, \bibinfo{person}{Michael Niemeyer}, \bibinfo{person}{Lars Mescheder}, \bibinfo{person}{Marc Pollefeys}, {and} \bibinfo{person}{Andreas Geiger}.} \bibinfo{year}{2020}\natexlab{}.
\newblock \showarticletitle{Convolutional occupancy networks}. In \bibinfo{booktitle}{\emph{ECCV}}.
\newblock


\bibitem[Pumarola et~al\mbox{.}(2021)]%
        {pumarola2020d}
\bibfield{author}{\bibinfo{person}{Albert Pumarola}, \bibinfo{person}{Enric Corona}, \bibinfo{person}{Gerard Pons-Moll}, {and} \bibinfo{person}{Francesc Moreno-Noguer}.} \bibinfo{year}{2021}\natexlab{}.
\newblock \showarticletitle{{D-NeRF: Neural Radiance Fields for Dynamic Scenes}}. In \bibinfo{booktitle}{\emph{CVPR}}.
\newblock


\bibitem[Rashid et~al\mbox{.}(2023)]%
        {rashid2023language}
\bibfield{author}{\bibinfo{person}{Adam Rashid}, \bibinfo{person}{Satvik Sharma}, \bibinfo{person}{Chung~Min Kim}, \bibinfo{person}{Justin Kerr}, \bibinfo{person}{Lawrence~Yunliang Chen}, \bibinfo{person}{Angjoo Kanazawa}, {and} \bibinfo{person}{Ken Goldberg}.} \bibinfo{year}{2023}\natexlab{}.
\newblock \showarticletitle{Language embedded radiance fields for zero-shot task-oriented grasping}. In \bibinfo{booktitle}{\emph{CoRL}}.
\newblock


\bibitem[Rebain et~al\mbox{.}(2021)]%
        {rebain2020derf}
\bibfield{author}{\bibinfo{person}{Daniel Rebain}, \bibinfo{person}{Wei Jiang}, \bibinfo{person}{Soroosh Yazdani}, \bibinfo{person}{Ke Li}, \bibinfo{person}{Kwang~Moo Yi}, {and} \bibinfo{person}{Andrea Tagliasacchi}.} \bibinfo{year}{2021}\natexlab{}.
\newblock \showarticletitle{Derf: Decomposed radiance fields}. In \bibinfo{booktitle}{\emph{CVPR}}.
\newblock


\bibitem[Sajjadi et~al\mbox{.}(2022a)]%
        {sajjadi2022object}
\bibfield{author}{\bibinfo{person}{Mehdi~SM Sajjadi}, \bibinfo{person}{Daniel Duckworth}, \bibinfo{person}{Aravindh Mahendran}, \bibinfo{person}{Sjoerd Van~Steenkiste}, \bibinfo{person}{Filip Pavetic}, \bibinfo{person}{Mario Lucic}, \bibinfo{person}{Leonidas~J Guibas}, \bibinfo{person}{Klaus Greff}, {and} \bibinfo{person}{Thomas Kipf}.} \bibinfo{year}{2022}\natexlab{a}.
\newblock \showarticletitle{Object scene representation transformer}. In \bibinfo{booktitle}{\emph{NeurIPS}}.
\newblock


\bibitem[Sajjadi et~al\mbox{.}(2022b)]%
        {sajjadi2022scene}
\bibfield{author}{\bibinfo{person}{Mehdi~SM Sajjadi}, \bibinfo{person}{Henning Meyer}, \bibinfo{person}{Etienne Pot}, \bibinfo{person}{Urs Bergmann}, \bibinfo{person}{Klaus Greff}, \bibinfo{person}{Noha Radwan}, \bibinfo{person}{Suhani Vora}, \bibinfo{person}{Mario Lu{\v{c}}i{\'c}}, \bibinfo{person}{Daniel Duckworth}, \bibinfo{person}{Alexey Dosovitskiy}, {et~al\mbox{.}}} \bibinfo{year}{2022}\natexlab{b}.
\newblock \showarticletitle{Scene representation transformer: Geometry-free novel view synthesis through set-latent scene representations}. In \bibinfo{booktitle}{\emph{CVPR}}.
\newblock


\bibitem[Shechtman and Irani(2007)]%
        {shechtman2007matching}
\bibfield{author}{\bibinfo{person}{Eli Shechtman} {and} \bibinfo{person}{Michal Irani}.} \bibinfo{year}{2007}\natexlab{}.
\newblock \showarticletitle{Matching local self-similarities across images and videos}. In \bibinfo{booktitle}{\emph{CVPR}}.
\newblock


\bibitem[Sitzmann et~al\mbox{.}(2020a)]%
        {sitzmann2020metasdf}
\bibfield{author}{\bibinfo{person}{Vincent Sitzmann}, \bibinfo{person}{Eric Chan}, \bibinfo{person}{Richard Tucker}, \bibinfo{person}{Noah Snavely}, {and} \bibinfo{person}{Gordon Wetzstein}.} \bibinfo{year}{2020}\natexlab{a}.
\newblock \showarticletitle{Metasdf: Meta-learning signed distance functions}. In \bibinfo{booktitle}{\emph{NeurIPS}}.
\newblock


\bibitem[Sitzmann et~al\mbox{.}(2020b)]%
        {sitzmann2019siren}
\bibfield{author}{\bibinfo{person}{Vincent Sitzmann}, \bibinfo{person}{Julien Martel}, \bibinfo{person}{Alexander Bergman}, \bibinfo{person}{David Lindell}, {and} \bibinfo{person}{Gordon Wetzstein}.} \bibinfo{year}{2020}\natexlab{b}.
\newblock \showarticletitle{Implicit neural representations with periodic activation functions}. In \bibinfo{booktitle}{\emph{NeurIPS}}.
\newblock


\bibitem[Srinivasan et~al\mbox{.}(2021)]%
        {nerv2021}
\bibfield{author}{\bibinfo{person}{Pratul~P. Srinivasan}, \bibinfo{person}{Boyang Deng}, \bibinfo{person}{Xiuming Zhang}, \bibinfo{person}{Matthew Tancik}, \bibinfo{person}{Ben Mildenhall}, {and} \bibinfo{person}{Jonathan~T. Barron}.} \bibinfo{year}{2021}\natexlab{}.
\newblock \showarticletitle{NeRV: Neural Reflectance and Visibility Fields for Relighting and View Synthesis}. In \bibinfo{booktitle}{\emph{CVPR}}.
\newblock


\bibitem[Stelzner et~al\mbox{.}(2021)]%
        {stelzner2021decomposing}
\bibfield{author}{\bibinfo{person}{Karl Stelzner}, \bibinfo{person}{Kristian Kersting}, {and} \bibinfo{person}{Adam~R Kosiorek}.} \bibinfo{year}{2021}\natexlab{}.
\newblock \showarticletitle{Decomposing 3d scenes into objects via unsupervised volume segmentation}.
\newblock \bibinfo{journal}{\emph{arXiv}} (\bibinfo{year}{2021}).
\newblock


\bibitem[Sucar et~al\mbox{.}(2021)]%
        {sucar2021imap}
\bibfield{author}{\bibinfo{person}{Edgar Sucar}, \bibinfo{person}{Shikun Liu}, \bibinfo{person}{Joseph Ortiz}, {and} \bibinfo{person}{Andrew~J Davison}.} \bibinfo{year}{2021}\natexlab{}.
\newblock \showarticletitle{iMAP: Implicit mapping and positioning in real-time}. In \bibinfo{booktitle}{\emph{ICCV}}.
\newblock


\bibitem[Takikawa et~al\mbox{.}(2022a)]%
        {takikawa2022variable}
\bibfield{author}{\bibinfo{person}{Towaki Takikawa}, \bibinfo{person}{Alex Evans}, \bibinfo{person}{Jonathan Tremblay}, \bibinfo{person}{Thomas M{\"u}ller}, \bibinfo{person}{Morgan McGuire}, \bibinfo{person}{Alec Jacobson}, {and} \bibinfo{person}{Sanja Fidler}.} \bibinfo{year}{2022}\natexlab{a}.
\newblock \showarticletitle{Variable bitrate neural fields}. In \bibinfo{booktitle}{\emph{SIGGRAPH}}.
\newblock


\bibitem[Takikawa et~al\mbox{.}(2021)]%
        {takikawa2021neural}
\bibfield{author}{\bibinfo{person}{Towaki Takikawa}, \bibinfo{person}{Joey Litalien}, \bibinfo{person}{Kangxue Yin}, \bibinfo{person}{Karsten Kreis}, \bibinfo{person}{Charles Loop}, \bibinfo{person}{Derek Nowrouzezahrai}, \bibinfo{person}{Alec Jacobson}, \bibinfo{person}{Morgan McGuire}, {and} \bibinfo{person}{Sanja Fidler}.} \bibinfo{year}{2021}\natexlab{}.
\newblock \showarticletitle{Neural geometric level of detail: Real-time rendering with implicit 3D shapes}. In \bibinfo{booktitle}{\emph{CVPR}}.
\newblock


\bibitem[Takikawa et~al\mbox{.}(2022b)]%
        {KaolinWispLibrary}
\bibfield{author}{\bibinfo{person}{Towaki Takikawa}, \bibinfo{person}{Or Perel}, \bibinfo{person}{Clement~Fuji Tsang}, \bibinfo{person}{Charles Loop}, \bibinfo{person}{Joey Litalien}, \bibinfo{person}{Jonathan Tremblay}, \bibinfo{person}{Sanja Fidler}, {and} \bibinfo{person}{Maria Shugrina}.} \bibinfo{year}{2022}\natexlab{b}.
\newblock \bibinfo{title}{Kaolin Wisp: A PyTorch Library and Engine for Neural Fields Research}.
\newblock \bibinfo{howpublished}{\url{https://github.com/NVIDIAGameWorks/kaolin-wisp}}.
\newblock


\bibitem[Tancik et~al\mbox{.}(2022)]%
        {tancik2022block}
\bibfield{author}{\bibinfo{person}{Matthew Tancik}, \bibinfo{person}{Vincent Casser}, \bibinfo{person}{Xinchen Yan}, \bibinfo{person}{Sabeek Pradhan}, \bibinfo{person}{Ben Mildenhall}, \bibinfo{person}{Pratul~P Srinivasan}, \bibinfo{person}{Jonathan~T Barron}, {and} \bibinfo{person}{Henrik Kretzschmar}.} \bibinfo{year}{2022}\natexlab{}.
\newblock \showarticletitle{Block-nerf: Scalable large scene neural view synthesis}. In \bibinfo{booktitle}{\emph{CVPR}}.
\newblock


\bibitem[Tancik et~al\mbox{.}(2021)]%
        {tancik2020learned}
\bibfield{author}{\bibinfo{person}{Matthew Tancik}, \bibinfo{person}{Ben Mildenhall}, \bibinfo{person}{Terrance Wang}, \bibinfo{person}{Divi Schmidt}, \bibinfo{person}{Pratul~P Srinivasan}, \bibinfo{person}{Jonathan~T Barron}, {and} \bibinfo{person}{Ren Ng}.} \bibinfo{year}{2021}\natexlab{}.
\newblock \showarticletitle{Learned initializations for optimizing coordinate-based neural representations}. In \bibinfo{booktitle}{\emph{CVPR}}.
\newblock


\bibitem[Tang et~al\mbox{.}(2021)]%
        {tang2021octfield}
\bibfield{author}{\bibinfo{person}{Jia-Heng Tang}, \bibinfo{person}{Weikai Chen}, \bibinfo{person}{Jie Yang}, \bibinfo{person}{Bo Wang}, \bibinfo{person}{Songrun Liu}, \bibinfo{person}{Bo Yang}, {and} \bibinfo{person}{Lin Gao}.} \bibinfo{year}{2021}\natexlab{}.
\newblock \showarticletitle{OctField: Hierarchical Implicit Functions for 3D Modeling}. In \bibinfo{booktitle}{\emph{NeurIPS}}.
\newblock


\bibitem[Tewari et~al\mbox{.}(2021)]%
        {tewari2021advances}
\bibfield{author}{\bibinfo{person}{Ayush Tewari}, \bibinfo{person}{Justus Thies}, \bibinfo{person}{Ben Mildenhall}, \bibinfo{person}{Pratul Srinivasan}, \bibinfo{person}{Edgar Tretschk}, \bibinfo{person}{Yifan Wang}, \bibinfo{person}{Christoph Lassner}, \bibinfo{person}{Vincent Sitzmann}, \bibinfo{person}{Ricardo Martin-Brualla}, \bibinfo{person}{Stephen Lombardi}, {et~al\mbox{.}}} \bibinfo{year}{2021}\natexlab{}.
\newblock \showarticletitle{Advances in neural rendering}.
\newblock \bibinfo{journal}{\emph{arXiv}} (\bibinfo{year}{2021}).
\newblock


\bibitem[Tremblay et~al\mbox{.}(2022)]%
        {tremblay2022rtmv}
\bibfield{author}{\bibinfo{person}{Jonathan Tremblay}, \bibinfo{person}{Moustafa Meshry}, \bibinfo{person}{Alex Evans}, \bibinfo{person}{Jan Kautz}, \bibinfo{person}{Alexander Keller}, \bibinfo{person}{Sameh Khamis}, \bibinfo{person}{Charles Loop}, \bibinfo{person}{Nathan Morrical}, \bibinfo{person}{Koki Nagano}, \bibinfo{person}{Towaki Takikawa}, {and} \bibinfo{person}{Stan Birchfield}.} \bibinfo{year}{2022}\natexlab{}.
\newblock \showarticletitle{RTMV: A Ray-Traced Multi-View Synthetic Dataset for Novel View Synthesis}.
\newblock \bibinfo{journal}{\emph{ECCV Workshops}}.
\newblock


\bibitem[Wang et~al\mbox{.}(2022)]%
        {wang2022geometry}
\bibfield{author}{\bibinfo{person}{Yifan Wang}, \bibinfo{person}{Lukas Rahmann}, {and} \bibinfo{person}{Olga Sorkine-Hornung}.} \bibinfo{year}{2022}\natexlab{}.
\newblock \showarticletitle{Geometry-consistent neural shape representation with implicit displacement fields}. In \bibinfo{booktitle}{\emph{ICLR}}.
\newblock


\bibitem[Wei et~al\mbox{.}(2021)]%
        {wei2021nerfingmvs}
\bibfield{author}{\bibinfo{person}{Yi Wei}, \bibinfo{person}{Shaohui Liu}, \bibinfo{person}{Yongming Rao}, \bibinfo{person}{Wang Zhao}, \bibinfo{person}{Jiwen Lu}, {and} \bibinfo{person}{Jie Zhou}.} \bibinfo{year}{2021}\natexlab{}.
\newblock \showarticletitle{Nerfingmvs: Guided optimization of neural radiance fields for indoor multi-view stereo}. In \bibinfo{booktitle}{\emph{ICCV}}.
\newblock


\bibitem[Williams et~al\mbox{.}(2022)]%
        {williams2021neural}
\bibfield{author}{\bibinfo{person}{Francis Williams}, \bibinfo{person}{Zan Gojcic}, \bibinfo{person}{Sameh Khamis}, \bibinfo{person}{Denis Zorin}, \bibinfo{person}{Joan Bruna}, \bibinfo{person}{Sanja Fidler}, {and} \bibinfo{person}{Or Litany}.} \bibinfo{year}{2022}\natexlab{}.
\newblock \showarticletitle{Neural fields as learnable kernels for 3d reconstruction}. In \bibinfo{booktitle}{\emph{CVPR}}.
\newblock


\bibitem[Xian et~al\mbox{.}(2021)]%
        {xian2020space}
\bibfield{author}{\bibinfo{person}{Wenqi Xian}, \bibinfo{person}{Jia-Bin Huang}, \bibinfo{person}{Johannes Kopf}, {and} \bibinfo{person}{Changil Kim}.} \bibinfo{year}{2021}\natexlab{}.
\newblock \showarticletitle{Space-time neural irradiance fields for free-viewpoint video}. In \bibinfo{booktitle}{\emph{CVPR}}.
\newblock


\bibitem[Xie et~al\mbox{.}(2021)]%
        {xie2021neural}
\bibfield{author}{\bibinfo{person}{Yiheng Xie}, \bibinfo{person}{Towaki Takikawa}, \bibinfo{person}{Shunsuke Saito}, \bibinfo{person}{Or Litany}, \bibinfo{person}{Shiqin Yan}, \bibinfo{person}{Numair Khan}, \bibinfo{person}{Federico Tombari}, \bibinfo{person}{James Tompkin}, \bibinfo{person}{Vincent Sitzmann}, {and} \bibinfo{person}{Srinath Sridhar}.} \bibinfo{year}{2021}\natexlab{}.
\newblock \showarticletitle{Neural Fields in Visual Computing and Beyond}.
\newblock \bibinfo{journal}{\emph{arXiv}} (\bibinfo{year}{2021}).
\newblock


\bibitem[Yang et~al\mbox{.}(2019)]%
        {yang2019pointflow}
\bibfield{author}{\bibinfo{person}{Guandao Yang}, \bibinfo{person}{Xun Huang}, \bibinfo{person}{Zekun Hao}, \bibinfo{person}{Ming-Yu Liu}, \bibinfo{person}{Serge Belongie}, {and} \bibinfo{person}{Bharath Hariharan}.} \bibinfo{year}{2019}\natexlab{}.
\newblock \showarticletitle{Pointflow: 3d point cloud generation with continuous normalizing flows}. In \bibinfo{booktitle}{\emph{ICCV}}.
\newblock


\bibitem[Yang et~al\mbox{.}(2022)]%
        {yang2022recursive}
\bibfield{author}{\bibinfo{person}{Guo-Wei Yang}, \bibinfo{person}{Wen-Yang Zhou}, \bibinfo{person}{Hao-Yang Peng}, \bibinfo{person}{Dun Liang}, \bibinfo{person}{Tai-Jiang Mu}, {and} \bibinfo{person}{Shi-Min Hu}.} \bibinfo{year}{2022}\natexlab{}.
\newblock \showarticletitle{Recursive-nerf: An efficient and dynamically growing nerf}.
\newblock \bibinfo{journal}{\emph{TVCG}} (\bibinfo{year}{2022}).
\newblock


\bibitem[Yi et~al\mbox{.}(2023)]%
        {yi2023canonical}
\bibfield{author}{\bibinfo{person}{Brent Yi}, \bibinfo{person}{Weijia Zeng}, \bibinfo{person}{Sam Buchanan}, {and} \bibinfo{person}{Yi Ma}.} \bibinfo{year}{2023}\natexlab{}.
\newblock \showarticletitle{Canonical factors for hybrid neural fields}. In \bibinfo{booktitle}{\emph{ICCV}}.
\newblock


\bibitem[Yu et~al\mbox{.}(2021)]%
        {yu2020pixelnerf}
\bibfield{author}{\bibinfo{person}{Alex Yu}, \bibinfo{person}{Vickie Ye}, \bibinfo{person}{Matthew Tancik}, {and} \bibinfo{person}{Angjoo Kanazawa}.} \bibinfo{year}{2021}\natexlab{}.
\newblock \showarticletitle{pixelnerf: Neural radiance fields from one or few images}. In \bibinfo{booktitle}{\emph{CVPR}}.
\newblock


\bibitem[Yu et~al\mbox{.}(2022)]%
        {yu2021unsupervised}
\bibfield{author}{\bibinfo{person}{Hong-Xing Yu}, \bibinfo{person}{Leonidas~J. Guibas}, {and} \bibinfo{person}{Jiajun Wu}.} \bibinfo{year}{2022}\natexlab{}.
\newblock \showarticletitle{Unsupervised Discovery of Object Radiance Fields}. In \bibinfo{booktitle}{\emph{ICLR}}.
\newblock


\bibitem[Yuan et~al\mbox{.}(2021)]%
        {yuan2021star}
\bibfield{author}{\bibinfo{person}{Wentao Yuan}, \bibinfo{person}{Zhaoyang Lv}, \bibinfo{person}{Tanner Schmidt}, {and} \bibinfo{person}{Steven Lovegrove}.} \bibinfo{year}{2021}\natexlab{}.
\newblock \showarticletitle{STaR: Self-supervised Tracking and Reconstruction of Rigid Objects in Motion with Neural Rendering}. In \bibinfo{booktitle}{\emph{CVPR}}.
\newblock


\bibitem[Zakharov et~al\mbox{.}(2022)]%
        {zakharov2022road}
\bibfield{author}{\bibinfo{person}{Sergey Zakharov}, \bibinfo{person}{Rares Ambrus}, \bibinfo{person}{Katherine Liu}, {and} \bibinfo{person}{Adrien Gaidon}.} \bibinfo{year}{2022}\natexlab{}.
\newblock \showarticletitle{ROAD: Learning an Implicit Recursive Octree Auto-Decoder to Efficiently Encode 3D Shapes}. In \bibinfo{booktitle}{\emph{CoRL}}.
\newblock


\bibitem[Zakharov et~al\mbox{.}(2021)]%
        {zakharov2021single}
\bibfield{author}{\bibinfo{person}{Sergey Zakharov}, \bibinfo{person}{Rares~Andrei Ambrus}, \bibinfo{person}{Vitor~Campagnolo Guizilini}, \bibinfo{person}{Dennis Park}, \bibinfo{person}{Wadim Kehl}, \bibinfo{person}{Fredo Durand}, \bibinfo{person}{Joshua~B Tenenbaum}, \bibinfo{person}{Vincent Sitzmann}, \bibinfo{person}{Jiajun Wu}, {and} \bibinfo{person}{Adrien Gaidon}.} \bibinfo{year}{2021}\natexlab{}.
\newblock \showarticletitle{Single-Shot Scene Reconstruction}. In \bibinfo{booktitle}{\emph{CoRL}}.
\newblock


\bibitem[Zakharov et~al\mbox{.}(2020)]%
        {zakharov2020autolabeling}
\bibfield{author}{\bibinfo{person}{Sergey Zakharov}, \bibinfo{person}{Wadim Kehl}, \bibinfo{person}{Arjun Bhargava}, {and} \bibinfo{person}{Adrien Gaidon}.} \bibinfo{year}{2020}\natexlab{}.
\newblock \showarticletitle{Autolabeling 3D Objects with Differentiable Rendering of SDF Shape Priors}. In \bibinfo{booktitle}{\emph{CVPR}}.
\newblock


\bibitem[Zeng et~al\mbox{.}(2022)]%
        {zeng2022lion}
\bibfield{author}{\bibinfo{person}{Xiaohui Zeng}, \bibinfo{person}{Arash Vahdat}, \bibinfo{person}{Francis Williams}, \bibinfo{person}{Zan Gojcic}, \bibinfo{person}{Or Litany}, \bibinfo{person}{Sanja Fidler}, {and} \bibinfo{person}{Karsten Kreis}.} \bibinfo{year}{2022}\natexlab{}.
\newblock \showarticletitle{LION: Latent Point Diffusion Models for 3D Shape Generation}. In \bibinfo{booktitle}{\emph{NeurIPS}}.
\newblock


\bibitem[Zhang et~al\mbox{.}(2020)]%
        {zhang2020nerf++}
\bibfield{author}{\bibinfo{person}{Kai Zhang}, \bibinfo{person}{Gernot Riegler}, \bibinfo{person}{Noah Snavely}, {and} \bibinfo{person}{Vladlen Koltun}.} \bibinfo{year}{2020}\natexlab{}.
\newblock \showarticletitle{Nerf++: Analyzing and improving neural radiance fields}.
\newblock \bibinfo{journal}{\emph{arXiv}} (\bibinfo{year}{2020}).
\newblock


\bibitem[Zhou et~al\mbox{.}(2021)]%
        {zhou20213d}
\bibfield{author}{\bibinfo{person}{Linqi Zhou}, \bibinfo{person}{Yilun Du}, {and} \bibinfo{person}{Jiajun Wu}.} \bibinfo{year}{2021}\natexlab{}.
\newblock \showarticletitle{3d shape generation and completion through point-voxel diffusion}. In \bibinfo{booktitle}{\emph{ICCV}}.
\newblock


\bibitem[Zhou and Jacobson(2016)]%
        {zhou2016thingi10k}
\bibfield{author}{\bibinfo{person}{Qingnan Zhou} {and} \bibinfo{person}{Alec Jacobson}.} \bibinfo{year}{2016}\natexlab{}.
\newblock \showarticletitle{Thingi10k: A dataset of 10,000 3d-printing models}.
\newblock \bibinfo{journal}{\emph{arXiv}} (\bibinfo{year}{2016}).
\newblock


\end{thebibliography}

\appendix
\clearpage
% \chapter{Supplementary Material}

\noindent{\huge Supplementary Material}
\section{Training Details}
\subsection{ReFiNe Training Data}
\method's training data consists of a ground truth octree structure covering the mesh at a desired LoD and densely sampled coordinates together with respective GT values (SDF, RGB, density). We sample $10^6$ points within 2 bands - a smaller one (LoD-1) and a larger one (LoD+1) to ensure sufficient coverage for recovering high frequency details and store respective supervision values (e.g. SDF, RGB, density).

Following~\cite{KaolinLibrary}, our octree is represented as a tensor of bytes, where each bit stands for the binary occupancy sorted in Morton order. The Morton order defines a space-filling curve, which provides a bijective mapping to 3D coordinates from 1D coordinates. As a result, this frees us from storing indirection pointers and allows efficient tree access. We additionally dilate our octree using a simple $3\times3\times3$ dilation kernel to secure a sufficient feature margin for trilinear interpolation.

All our networks are trained on a single NVIDIA A100 GPU.

\subsection{Baseline Method Details}

\paragraph{DeepSDF}
We use the open-source implementation of DeepSDF \cite{park2019deepsdf}. To generate training data, we preprocess models from Thingi32 and ShapeNet150 via the provided code and parameters, which aims to generate approximately 500k training points. We improve results on the overfitting scenario by setting the dropout rate to zero and removing the latent regularization. We use a learning rate of $0.001$ for the decoder network parameters and $0.002$ for latents as well as a decay factor of $0.75$ every 500 steps, training the methods until convergence (about 20k epochs). For the experiments on Thingi32 we use a batch size of 32 objects and for ShapeNet150 we use a batch size of 64 objects. All other parameters we leave as provided by the example implementations (i.e., we used a code length of 256 and keep the neural network architecture unchanged).

\paragraph{Curriculum DeepSDF}
We also use the open-source implementation of Curriculum-DeepSDF \cite{duan2020curriculum}. We duplicate the parameter changes made to DeepSDF for consistency, and use the same training data input. We do not modify the curriculum proposed in \cite{duan2020curriculum} other than lengthening the last stage of training. We observe that the proposed curriculum provided quantitative reconstruction gains for ShapeNet150 and not Thingi32, suggesting that a different curriculum may improve results for the latter dataset. However, searching for the optimal curriculum is expensive and we choose to report results based on the baseline curriculum given in the open-source implementation.

\paragraph{SRN \& CodeNeRF} We use the open-source implementations with default configurations for SRN~\cite{sitzmann2019scene} and CodeNeRF~\cite{jang2021codenerf} and train both methods on our subset of the SRN dataset as described in Section 4.3 of the main paper. Both baselines use a default latent code size of 256, whereas CodeNeRF uses 2 latent codes of 256 to represent an object - one for geometry, another for appearance. In Table 2 and Fig. 6 of the main paper we demonstrate that our method outperforms both baselines, while using a more lightweight architecture and a latent code size of 64.

\setlength{\tabcolsep}{12pt}
\begin{table}[b]
    \caption{The multiscale feature fusion scheme vs reconstruction quality on HB dataset. At the cost of storage, concatenating features before decoding increases reconstruction accuracy.}
\resizebox{\columnwidth}{!}{%
  \centering
    \begin{tabular}{lcccc}
    \toprule
    \multirow{2}[4]{*}{\textbf{Fusion}} & \multicolumn{2}{c}{\textit{Reconstruction}} & \textit{Runtime} & \textit{Size} \\
\cmidrule(lr){2-3} \cmidrule(lr){4-4}  \cmidrule(lr){5-5}        & CD$\downarrow$ & 3D PSNR$\uparrow$ & s$\downarrow$ & MB$\downarrow$ \\
    \midrule
    \textit{Sum}   & 0.046 & 33.61 & 0.11 & 3.2 \\
    \textit{Concatenate} & 0.046 & 34.89 & 0.12 & 3.8 \\
    \bottomrule
    \end{tabular}%
    }
  \label{tab:fusion}%
\end{table}%

\begin{figure*}[t]
	\centering
	\includegraphics[width=1\linewidth]{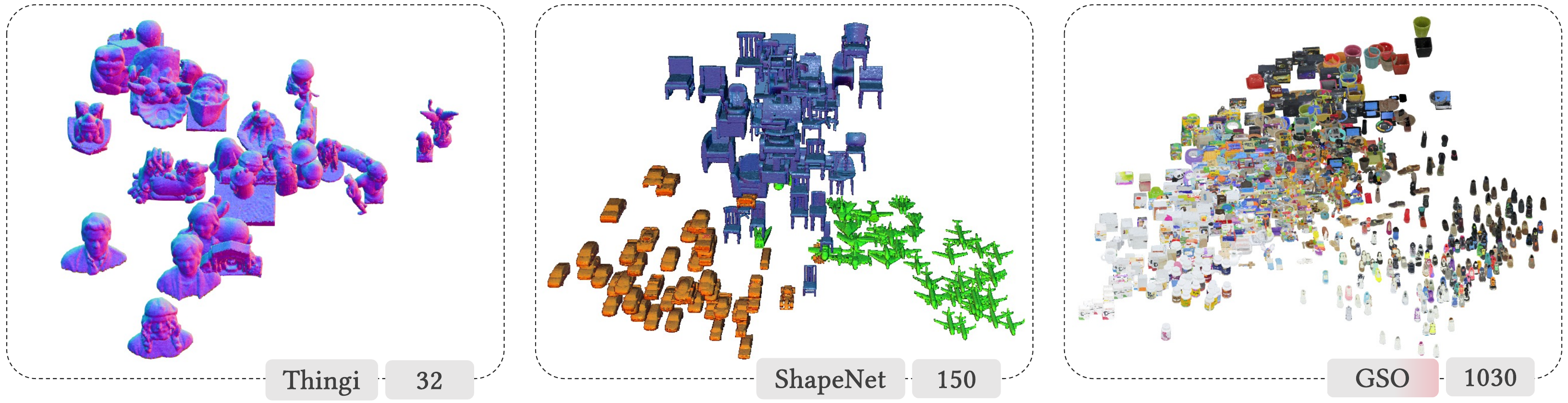}
	\caption{\textbf{Latent space visualization}. We use PCA to visualize the encoded shapes of Thingi32, ShapeNet150, and Google Scanned Objects in two dimensions. The qualitative results suggest that \methodspace clusters similar objects  together by color and or geometry, pointing to potential classification utility. We observe that the latent space clustering is more pronounced in the considered datasets with more encoded objects as in GSO or less unique object classes as in ShapeNet150.}
 % defined by geometry (Thingi32, ShapeNet150) or geometry and color (Google Scanned Objects)
	\label{fig:latent_spaces}
\end{figure*}

\section{Evaluation Details}

To calculate the Chamfer distance for DeepSDF and Curriculum DeepSDF we first extract surface points following the protocol of~\cite{irshad2022shapo}. In particular, we define a coarse voxel grid of LoD 2 and estimate SDF values for each
of the points using a pretrained SDF network. The voxels whose SDF values are larger than their size are pruned and the remaining voxels are propagated to the next level via subdivision. When the desired LoD is reached, we use zero isosurface projection to extract surface points using predicted SDF values and estimated surface normals. Finally, we use the Chamfer distance implementation from~\cite{KaolinWispLibrary} to compare our prediction against a ground truth point cloud of $2^{17}$ points sampled from the original mesh. When reconstructing SDF + Color, we additionally use PSNR to evaluate RGB values regressed from the same $2^{17}$ points. 
To compute gIoU, we first reconstruct a mesh using Poisson surface reconstruction from~\cite{kazhdan2006poisson} and then compare against $2^{17}$ ground truth values randomly computed using the original mesh. 
% When evaluating NeRF for novel view synthesis, we first render the images given a test pose and compare RGB values against GT using common PSNR and SSIM metrics.

\begin{figure}[b]
	\centering
	\includegraphics[width=0.8\linewidth]{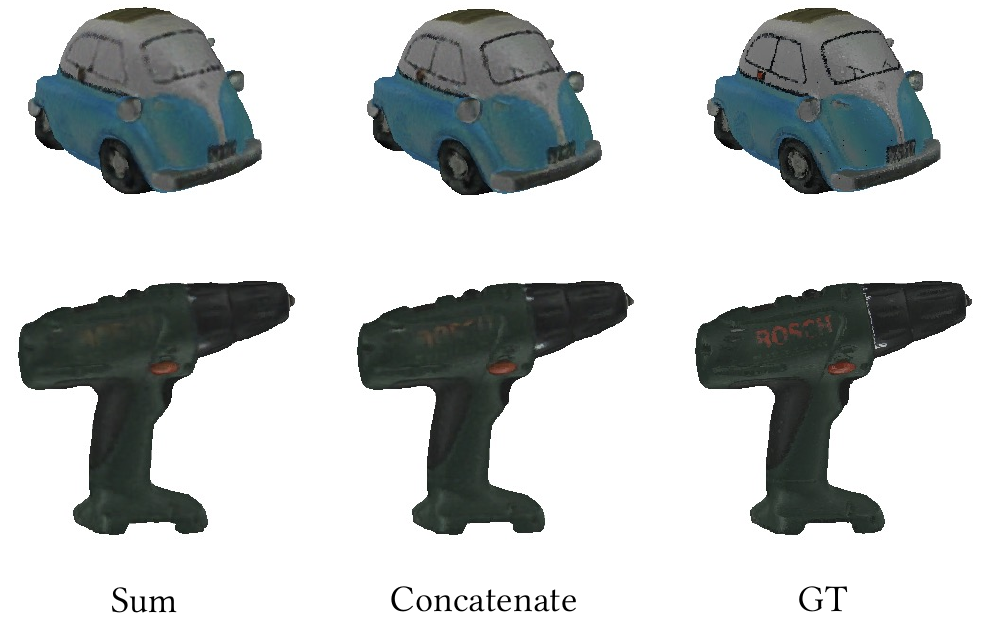}
	\caption{
	\textbf{Effect of the latent fusion scheme on reconstruction quality of the HB dataset.} The concatenation scheme better preserves object details at the cost of storage.
 \label{figure:hb_fusion}}
\end{figure}

\section{Additional Results}

\paragraph{Multiscale Feature Interpolation} In Table~\ref{tab:fusion} we perform an ablation studying how the multi-scale feature fusion scheme affects the final reconstruction quality. For this experiment, we use the latent size of 64, our recursive autodecoder network $\latentSubdivision$ consists of a single 1024-dimensional layer, and all decoding networks $\occupancyDecoder$,  $\neuralMappingGeometry$ and $\neuralMappingColor$ use two-layers of 256 fully connected units each. We use HomebrewedDB~\cite{kaskman2019homebreweddb} - a 6D pose estimation dataset from BOP benchmark~\cite{hodan2018bop} comprising 33 colored meshes (17 toy, 8 household and 8 industry-relevant) of various complexity in terms of both geometry and color. Two methods of feature fusion to combine interpolated features from multiple LoDs are considered: \textit{Sum}, where the latents are simply added together, and \textit{Concatenate}, where the interpolated latents from each LoD are concatenated together. Both modalities are trained to encode the full dataset consisting of 33 objects. As was shown in Table 2 of the main paper, the \textit{Sum} fusion scheme preserves the network size across different possible LoDs, because it doesn't change the input size for the respective decoder networks and we have a single recursive network $\latentSubdivision$ by design. On the other hand, the \textit{Concatenate} scheme comes at a higher storage cost as the corresponding decoding networks must have larger input layers, but results in an improved 3D PSNR value as shown in Table~\ref{tab:fusion}. As can be seen in Fig.~\ref{figure:hb_fusion}, while both schemes manage to faithfully represent object geometry the \textit{Concatenate} scheme does better when it comes to preserving high-frequency color details.

\begin{figure}[b]
% \vspace{-2mm}
	\centering	\includegraphics[width=1\linewidth]{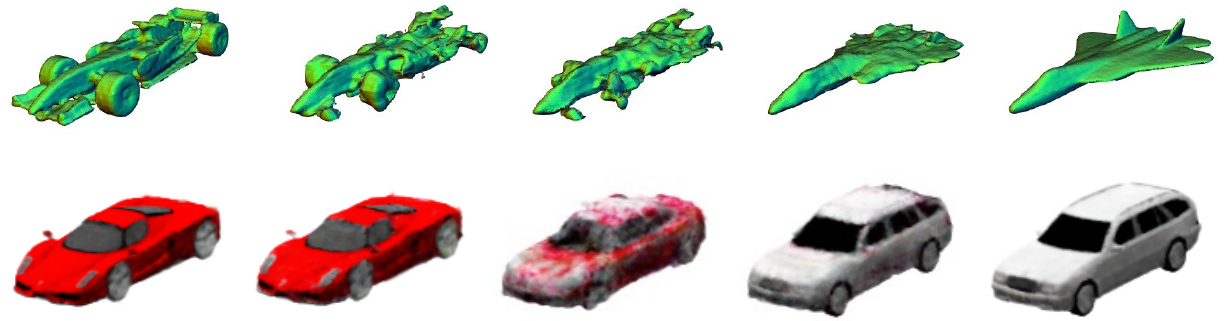}
    \caption{
	\textbf{Latent space interpolation}. Our latent space allows for smooth interpolation between nearby objects, both for SDFs from ShapeNet150 (top) and NeRFs from SRN (bottom). 
 \label{fig:latent}}
% \vspace{-2mm}
\end{figure}

\paragraph{Latent Space Interpolation and Clustering}
We present a qualitative analysis of our latent space conducted on the ShapeNet150 and SRN Cars datasets. As our method outputs a continuous feature field, it can be used for interpolation in the latent space between objects of similar geometry. Figure~\ref{fig:latent} shows an example of such interpolation between two objects of different classes. 
In addition, we plot latent spaces of Thingi32, ShapeNet150, and Google Scanned Objects represented by respective networks using the principal component analysis (see Fig.~\ref{fig:latent_spaces}). Projected latent spaces suggest that the structure of ReFiNe's latent space clusters similar objects defined either by geometry (Thingi32, ShapeNet150) or geometry and color (Google Scanned Objects), pointing to potential classification utility. 

% \clearpage
% \newpage

\paragraph{Single-Scene Baselines}
Our main paper features baselines that are carefully selected to adhere to our key paradigm of representing the entire dataset with a single network, where each object or scene is represented by a single compact latent vector. However, it is also useful to evaluate how our results fare against single-scene methods that use a single network per object or scene. In this section, we compare our results with single-scene methods on SDF and NeRF benchmarks. All storage sizes include both network and latent vector sizes. 
Table~\ref{tab:sdf_single} shows the results on the ShapeNet150 and Thingi32 SDF benchmark. We compare our method against two baselines: mip-Neural Implicits~\cite{davies2020overfit}, and NGLOD~\cite{takikawa2021neural}. Our method outperforms single-scene baselines on ShapeNet150 both in terms of reconstruction quality and storage and demonstrates a comparable performance on Thingi32.
Similarly, Table~\ref{tab:rtmv_single} shows the results on the RTMV benchmark. We compare our method against two baselines: mip-NeRF~\cite{barron2021mip}, and SVLF~\cite{tremblay2022rtmv}. As results show, \methodspace is able to approach the performance of single-scene methods while storing all 40 scenes within a single network providing substantially lower storage requirements without specifically optimizing for compression. We attribute this to the recursive nature of our method splitting scene space into primitives at each recursive step.

\setlength{\tabcolsep}{4pt}
\begin{table}[t]
  \centering
     \caption{Per-dataset vs per-scene methods on SDF benchmarks.}
    \resizebox{\columnwidth}{!}{
    \begin{tabular}{lccccccc}
    \toprule
    \multirow{2}[4]{*}{\textbf{Method}} & \multirow{2}[4]{*}{Type} & \multicolumn{3}{c}{ShapeNet150} & \multicolumn{3}{c}{Thingi32} \\
\cmidrule(lr){3-5} \cmidrule(lr){6-8}           &       & CD$\downarrow$    & gIoU$\uparrow$  & MB$\downarrow$    & CD$\downarrow$    & gIoU$\uparrow$  & MB$\downarrow$ \\
    \midrule
    Neural Implicits & \multirow{2}[0]{*}{Per-Scene} & 0.500 & 82.2  & 4.4   & 0.092 & 96.0  & 0.9 \\
    NGLOD &       & 0.062 & 91.7  & 185.4 & 0.027 & 99.4  & 39.6 \\
    \midrule
    ReFiNe/LoD6 & Per-Dataset & 0.019 & 99.4  & 3.9   & 0.027 & 97.4  & 3.2 \\
    \bottomrule
    \end{tabular}%
    }
  \label{tab:sdf_single}%
\end{table}%

\setlength{\tabcolsep}{8pt}
\begin{table}[t]
  \centering
      \caption{Per-dataset vs per-scene methods on RTMV (NeRF).}
      \resizebox{\columnwidth}{!}{
    \begin{tabular}{lccccc}
    \toprule
    \multirow{2}[3]{*}{\textbf{Method}} & \multirow{2}[3]{*}{Type} & \multicolumn{3}{c}{\textit{View synthesis}} & \textit{Storage} \\
\cmidrule(lr){3-5} \cmidrule(lr){6-6}          &       & PSNR$\uparrow$  & SSIM$\uparrow$  & LPIPS$\downarrow$ & MB$\downarrow$ \\
    \midrule
    mip-NeRF & \multirow{2}[0]{*}{Per-Scene} & 30.53 & 0.91  & 0.06  & 7.4 * 40 \\
    SVLF  &       & 28.83 & 0.91  & 0.06  & 947 * 40 \\
    \midrule
    ReFiNe/LoD6 & Per-Dataset & 26.72 & 0.87  & 0.19  & 45.6 \\
    \bottomrule
    \end{tabular}%
    }
  \label{tab:rtmv_single}%
\end{table}%

\paragraph{SIREN vs ReLU}
Our recursive subdivision network $\latentSubdivision$ and all decoding networks $\occupancyDecoder$, $\neuralMappingGeometry$ and $\neuralMappingColor$ are parametrized with SIREN-based MLPs using periodic activation functions. In this ablation, we evaluate how replacing SIREN-based MLPs with standard vanilla ReLU-based MLPs affects the reconstruction metrics for scenes using different field representations. To accomplish this, we select a single object from each modality (T-Rex from Thingi32 for SDF, Dog from HB for SDF+RGB, car from SRNCars for NeRF) and overfit a single MLP to each of the modalities. All baselines use a latent size of 64, a single 1024-dimensional layer for recursive subdivision network $\latentSubdivision$, and 256-dimensional two-layer decoding networks $\occupancyDecoder$, $\neuralMappingGeometry$, and $\neuralMappingColor$.
Our results shown in Table~\ref{tab:siren_relu} demonstrate that a naive ReLU-based MLP implementation performs worse overall and especially suffers when it comes to reconstructing high frequency details and colors.

\begin{figure}[b]
% \vspace{-2mm}
	\centering	\includegraphics[width=0.9\linewidth]{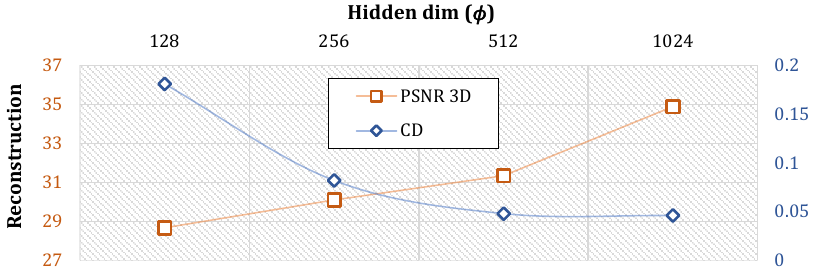}
    \caption{
	\textbf{Network size vs Reconstruction quality}. We observe a graceful degradation of reconstruction quality with decreasing the hidden dimension of the recursive subdivision network $\latentSubdivision$. 
 \label{fig:nsize_quality}
 }
% \vspace{-2mm}
\end{figure}

\paragraph{Network Size vs Reconstruction Quality}
Similar to our latent size experiments in Table 3 of the main paper, in this ablation we study how changing the hidden dimension of our recursive subdivision network $\latentSubdivision$ affects reconstruction quality. We train four baselines with different sizes for the hidden dimension of the recursive subdivision network, $\latentSubdivision$: 128, 256, 512, and 1024. The remaining parameters are consistent across all four networks: a latent size of 64, and each of the decoding networks $\occupancyDecoder$, $\neuralMappingGeometry$, and $\neuralMappingColor$ utilizes two layers of 256 fully connected units. All the baselines are trained on the HB dataset (SDF+RGB). As shown in Fig.~\ref{fig:nsize_quality}, we observe a graceful degradation of quality with decreasing network capacity.

\begin{figure*}[t]
    \centering
    \vspace{-5pt}
    
    \begin{subfigure}[t]{0.2\linewidth}
        \centering
        \includegraphics[width=\textwidth]{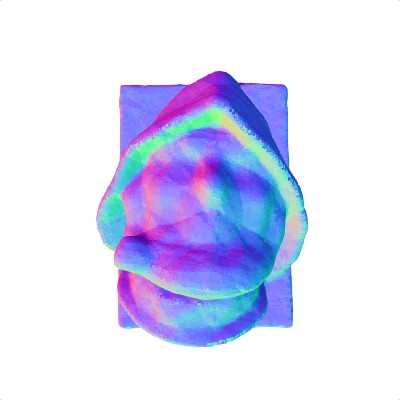}\vspace{-5pt}
        \caption*{DeepSDF}
    \end{subfigure}%
    \begin{subfigure}[t]{0.2\linewidth}
        \centering
        \includegraphics[width=\textwidth]{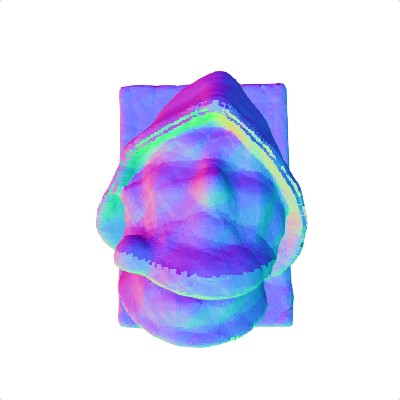}\vspace{-5pt}
        \caption*{Curriculum DeepSDF}
    \end{subfigure}%
    \begin{subfigure}[t]{0.2\linewidth}
        \centering
        \includegraphics[width=\textwidth]{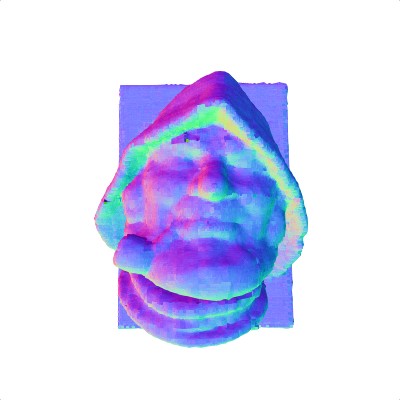}\vspace{-5pt}
        \caption*{ROAD / LoD9}
    \end{subfigure}%
    \begin{subfigure}[t]{0.2\linewidth}
        \centering
        \includegraphics[width=\textwidth]{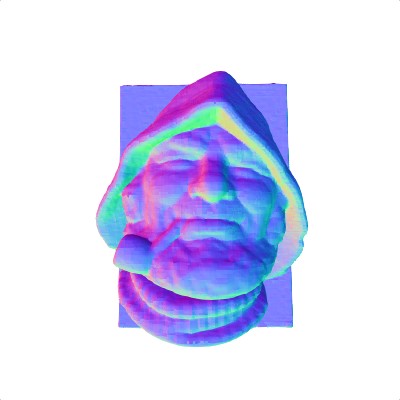}\vspace{-5pt}
        \caption*{Ours / LoD6}
    \end{subfigure}%
    \begin{subfigure}[t]{0.2\linewidth}
        \centering
        \includegraphics[width=\textwidth]{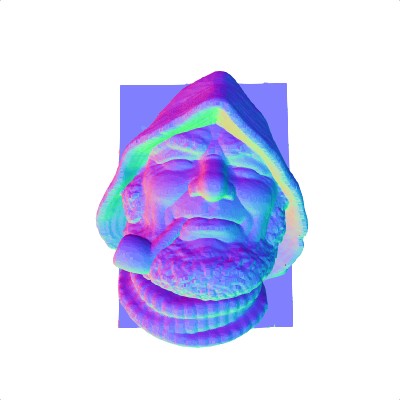}\vspace{-5pt}
        \caption*{GT}
    \end{subfigure}

    \begin{subfigure}[t]{0.2\linewidth}
        \centering
        \includegraphics[width=\textwidth]{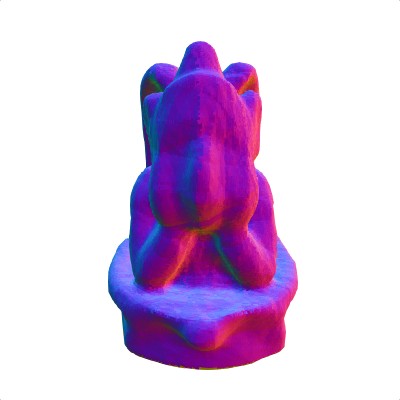}\vspace{-5pt}
        \caption*{DeepSDF }
    \end{subfigure}%
    \begin{subfigure}[t]{0.2\linewidth}
        \centering
        \includegraphics[width=\textwidth]{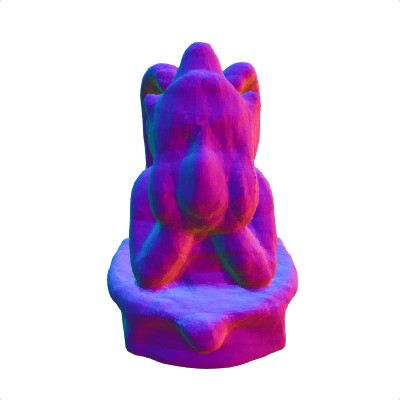}\vspace{-5pt}
        \caption*{Curriculum DeepSDF}
    \end{subfigure}%
    \begin{subfigure}[t]{0.2\linewidth}
        \centering
        \includegraphics[width=\textwidth]{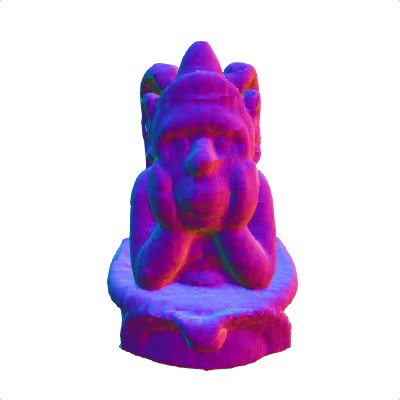}\vspace{-5pt}
        \caption*{ROAD / LoD9}
    \end{subfigure}%
    \begin{subfigure}[t]{0.2\linewidth}
        \centering
        \includegraphics[width=\textwidth]{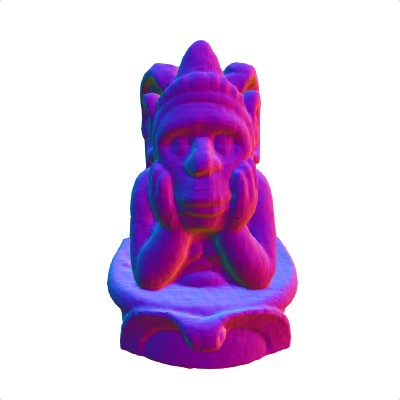}\vspace{-5pt}
        \caption*{Ours / LoD6}
    \end{subfigure}%
    \begin{subfigure}[t]{0.2\linewidth}
        \centering
        \includegraphics[width=\textwidth]{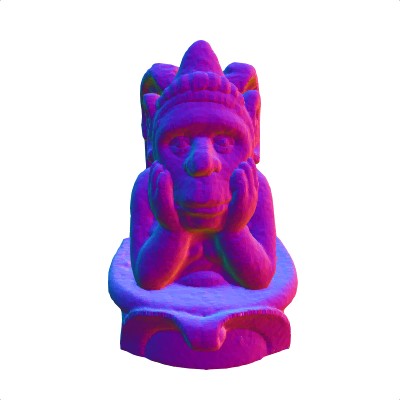}\vspace{-5pt}
        \caption*{GT}
    \end{subfigure}

    \begin{subfigure}[t]{0.2\linewidth}
        \centering
        \includegraphics[width=\textwidth]{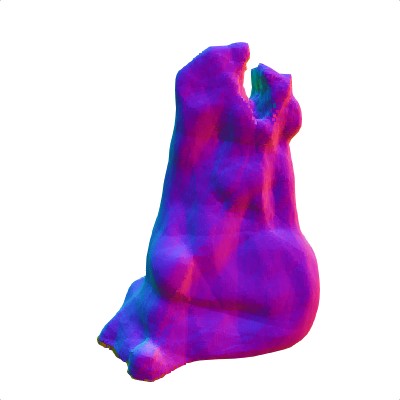}\vspace{-5pt}
        \caption*{DeepSDF }
    \end{subfigure}%
    \begin{subfigure}[t]{0.2\linewidth}
        \centering
        \includegraphics[width=\textwidth]{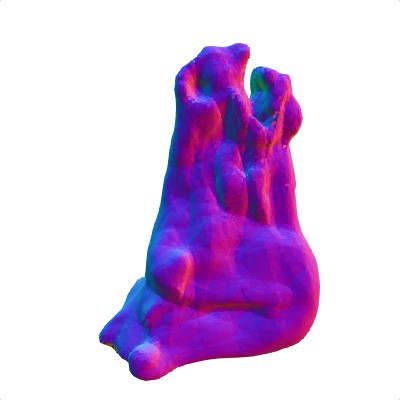}\vspace{-5pt}
        \caption*{Curriculum DeepSDF}
    \end{subfigure}%
    \begin{subfigure}[t]{0.2\linewidth}
        \centering
        \includegraphics[width=\textwidth]{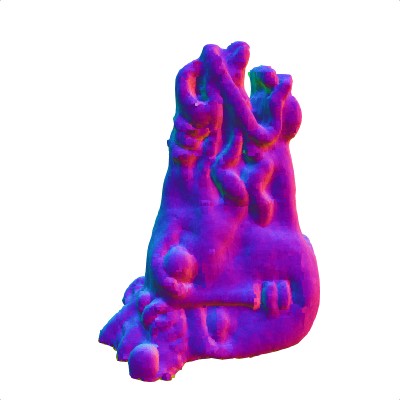}\vspace{-5pt}
        \caption*{ROAD / LoD9}
    \end{subfigure}%
    \begin{subfigure}[t]{0.2\linewidth}
        \centering
        \includegraphics[width=\textwidth]{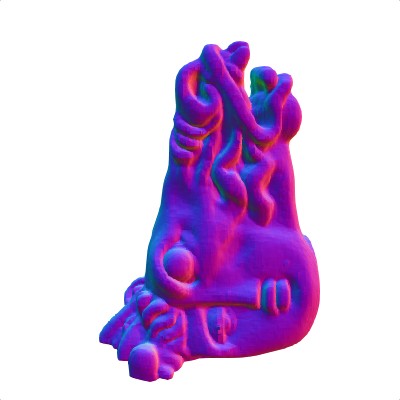}\vspace{-5pt}
        \caption*{Ours / LoD6}
    \end{subfigure}%
    \begin{subfigure}[t]{0.2\linewidth}
        \centering
        \includegraphics[width=\textwidth]{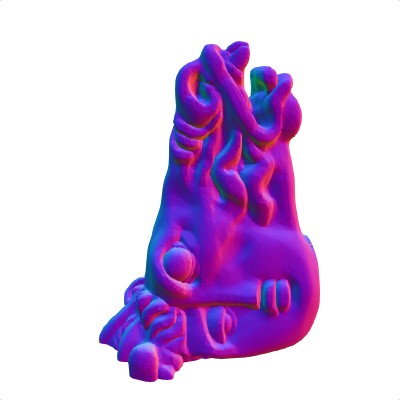}\vspace{-5pt}
        \caption*{GT}
    \end{subfigure}   

    \begin{subfigure}[t]{0.2\linewidth}
        \centering
        \includegraphics[width=\textwidth]{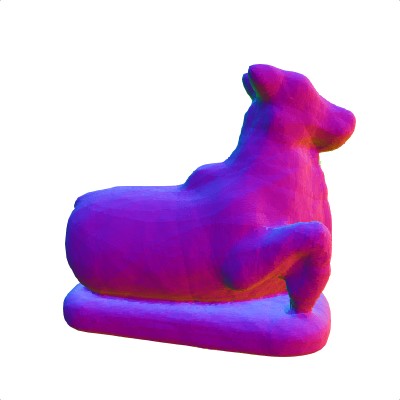}\vspace{-5pt}
        \caption*{DeepSDF }
    \end{subfigure}%
    \begin{subfigure}[t]{0.2\linewidth}
        \centering
        \includegraphics[width=\textwidth]{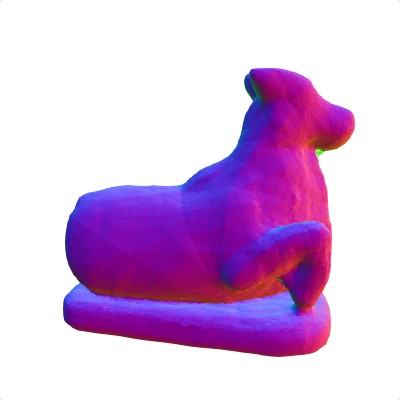}\vspace{-5pt}
        \caption*{Curriculum DeepSDF}
    \end{subfigure}%
    \begin{subfigure}[t]{0.2\linewidth}
        \centering
        \includegraphics[width=\textwidth]{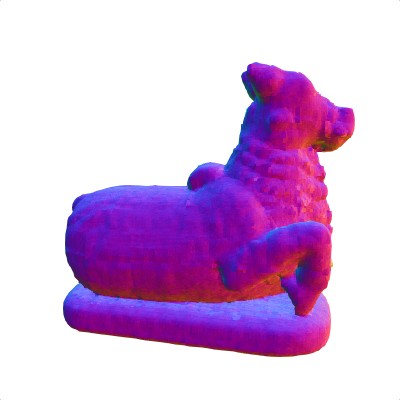}\vspace{-5pt}
        \caption*{ROAD / LoD9}
    \end{subfigure}%
    \begin{subfigure}[t]{0.2\linewidth}
        \centering
        \includegraphics[width=\textwidth]{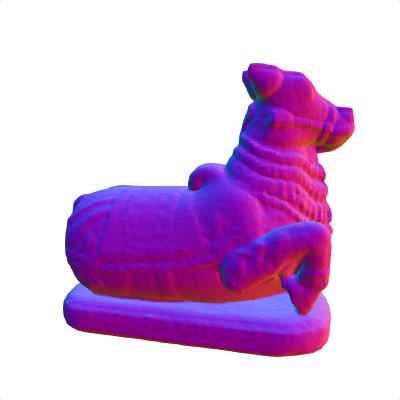}\vspace{-5pt}
        \caption*{Ours / LoD6}
    \end{subfigure}%
    \begin{subfigure}[t]{0.2\linewidth}
        \centering
        \includegraphics[width=\textwidth]{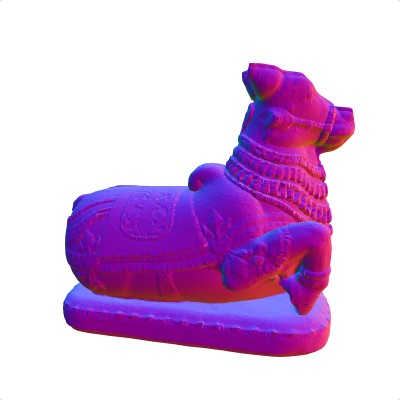}\vspace{-5pt}
        \caption*{GT}
    \end{subfigure}    

    \begin{subfigure}[t]{0.2\linewidth}
        \centering
        \includegraphics[width=\textwidth]{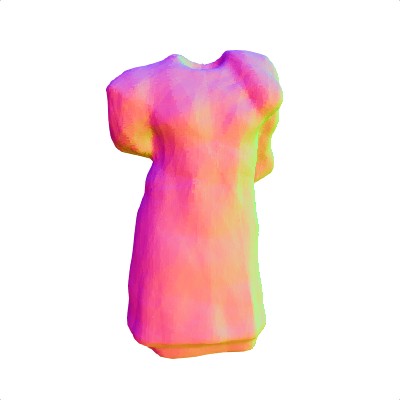}\vspace{-5pt}
        \caption*{DeepSDF }
    \end{subfigure}%
    \begin{subfigure}[t]{0.2\linewidth}
        \centering
        \includegraphics[width=\textwidth]{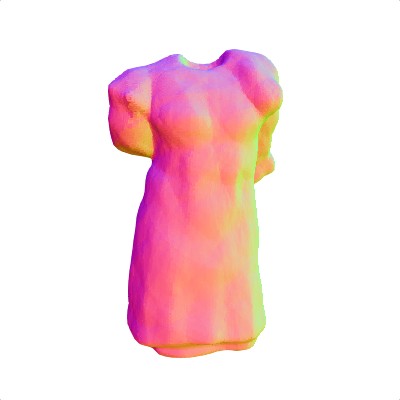}\vspace{-5pt}
        \caption*{Curriculum DeepSDF}
    \end{subfigure}%
    \begin{subfigure}[t]{0.2\linewidth}
        \centering
        \includegraphics[width=\textwidth]{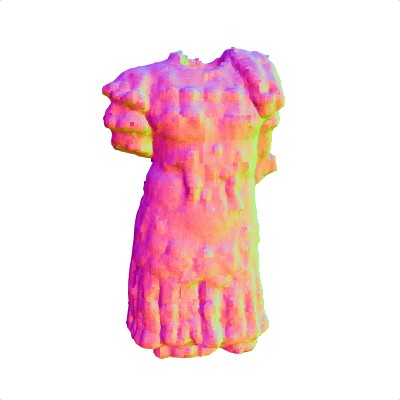}\vspace{-5pt}
        \caption*{ROAD / LoD9}
    \end{subfigure}%
    \begin{subfigure}[t]{0.2\linewidth}
        \centering
        \includegraphics[width=\textwidth]{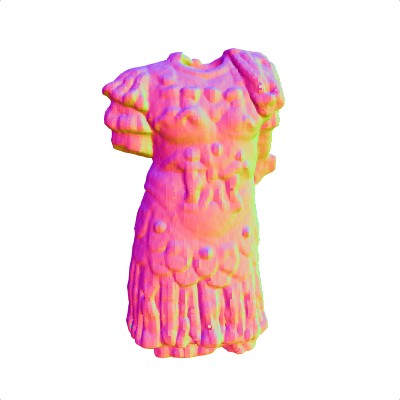}\vspace{-5pt}
        \caption*{Ours / LoD6}
    \end{subfigure}%
    \begin{subfigure}[t]{0.2\linewidth}
        \centering
        \includegraphics[width=\textwidth]{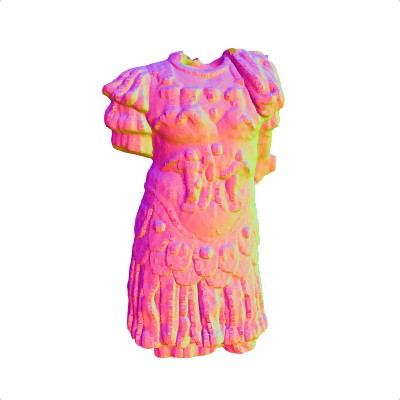}\vspace{-5pt}
        \caption*{GT}
    \end{subfigure}

    \caption{\textbf{Additional Thingi32 Results.} Best viewed zoomed in.}
\label{fig:supp_thingi32}
\vspace{-9pt}
\end{figure*}

\begin{figure*}[t]
    \centering
    \vspace{-5pt}
    
    \begin{subfigure}[t]{0.2\linewidth}
        \centering
        \includegraphics[width=\textwidth]{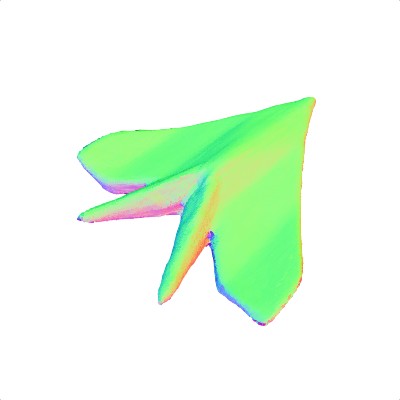}\vspace{-5pt}
        \caption*{DeepSDF}
    \end{subfigure}%
    \begin{subfigure}[t]{0.2\linewidth}
        \centering
        \includegraphics[width=\textwidth]{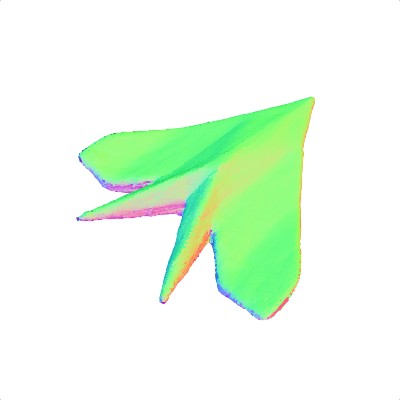}\vspace{-5pt}
        \caption*{Curriculum DeepSDF}
    \end{subfigure}%
    \begin{subfigure}[t]{0.2\linewidth}
        \centering
        \includegraphics[width=\textwidth]{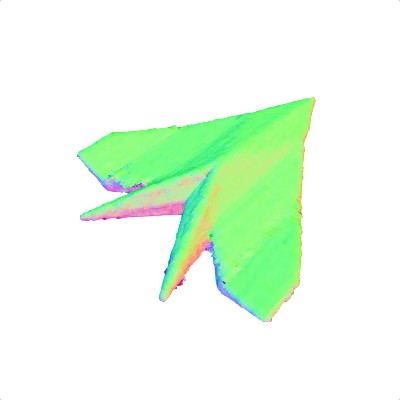}\vspace{-5pt}
        \caption*{ROAD / LoD9}
    \end{subfigure}%
    \begin{subfigure}[t]{0.2\linewidth}
        \centering
        \includegraphics[width=\textwidth]{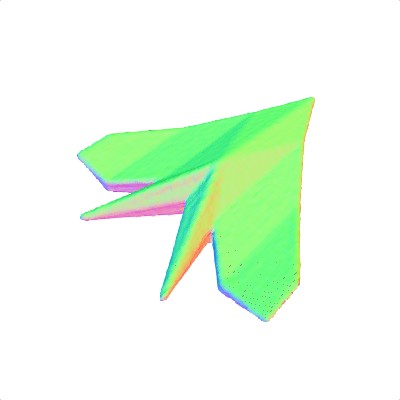}\vspace{-5pt}
        \caption*{Ours / LoD6}
    \end{subfigure}%
    \begin{subfigure}[t]{0.2\linewidth}
        \centering
        \includegraphics[width=\textwidth]{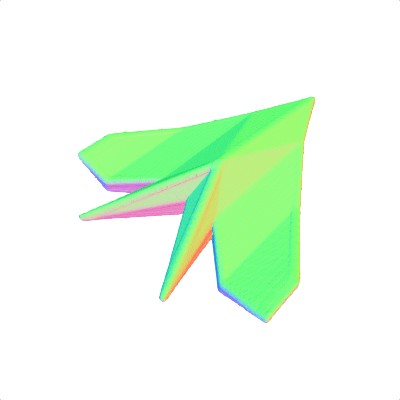}\vspace{-5pt}
        \caption*{GT}
    \end{subfigure}

    \begin{subfigure}[t]{0.2\linewidth}
        \centering
        \includegraphics[width=\textwidth]{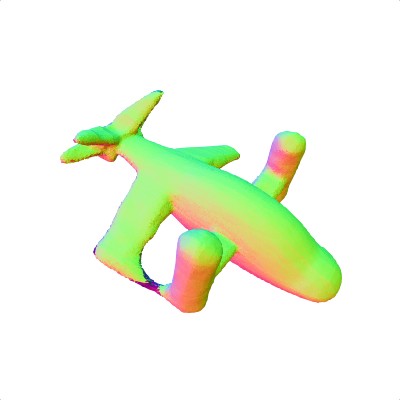}\vspace{-5pt}
        \caption*{DeepSDF}
    \end{subfigure}%
    \begin{subfigure}[t]{0.2\linewidth}
        \centering
        \includegraphics[width=\textwidth]{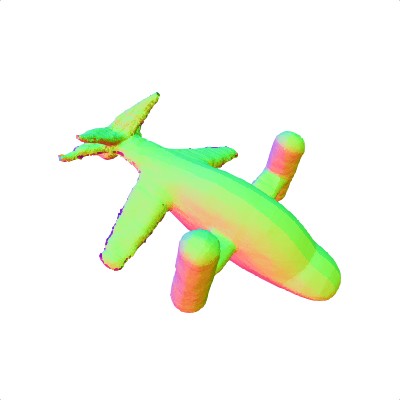}\vspace{-5pt}
        \caption*{Curriculum DeepSDF}
    \end{subfigure}%
    \begin{subfigure}[t]{0.2\linewidth}
        \centering
        \includegraphics[width=\textwidth]{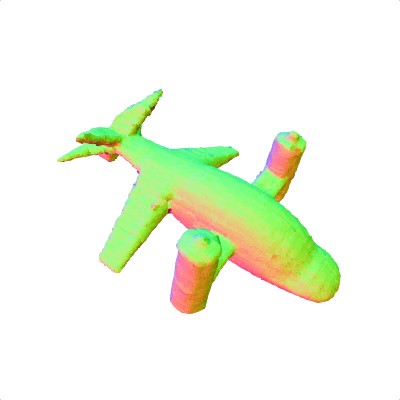}\vspace{-5pt}
        \caption*{ROAD / LoD9}
    \end{subfigure}%
    \begin{subfigure}[t]{0.2\linewidth}
        \centering
        \includegraphics[width=\textwidth]{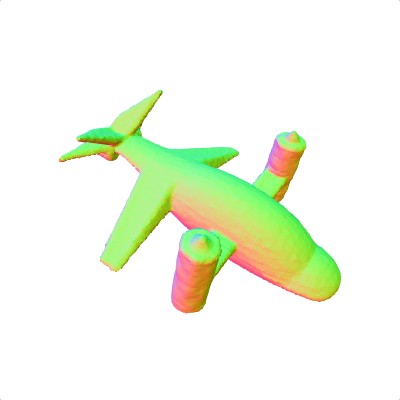}\vspace{-5pt}
        \caption*{Ours / LoD6}
    \end{subfigure}%
    \begin{subfigure}[t]{0.2\linewidth}
        \centering
        \includegraphics[width=\textwidth]{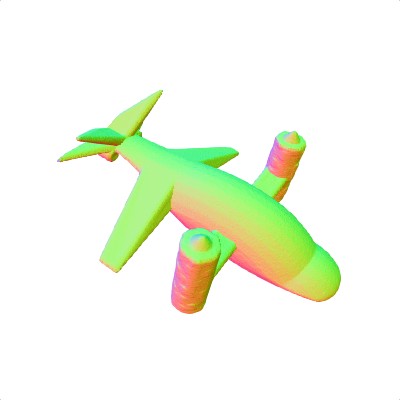}\vspace{-5pt}
        \caption*{GT}
    \end{subfigure}    

    \begin{subfigure}[t]{0.2\linewidth}
        \centering
        \includegraphics[width=\textwidth]{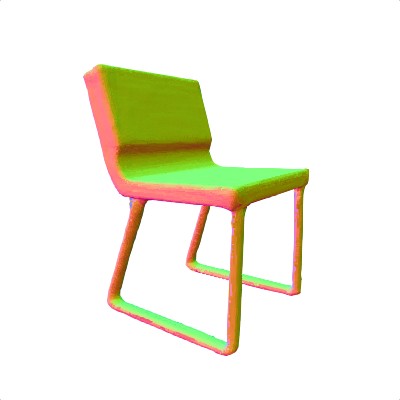}\vspace{-5pt}
        \caption*{DeepSDF}
    \end{subfigure}%
    \begin{subfigure}[t]{0.2\linewidth}
        \centering
        \includegraphics[width=\textwidth]{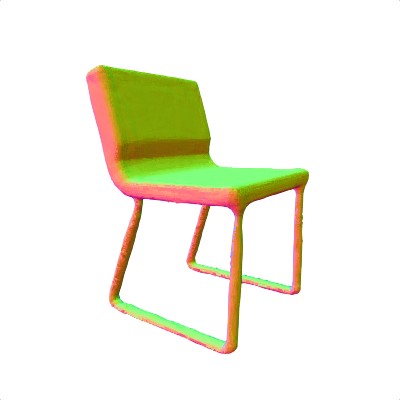}\vspace{-5pt}
        \caption*{Curriculum DeepSDF}
    \end{subfigure}%
    \begin{subfigure}[t]{0.2\linewidth}
        \centering
        \includegraphics[width=\textwidth]{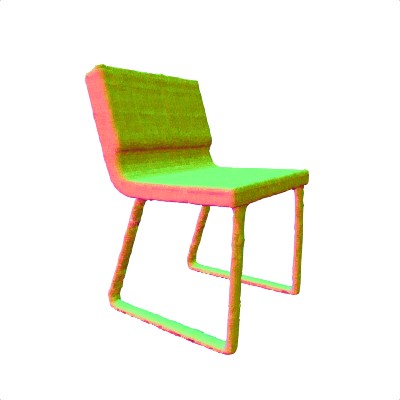}\vspace{-5pt}
        \caption*{ROAD / LoD9}
    \end{subfigure}%
    \begin{subfigure}[t]{0.2\linewidth}
        \centering
        \includegraphics[width=\textwidth]{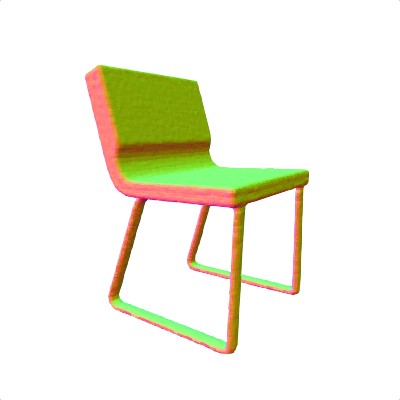}\vspace{-5pt}
        \caption*{Ours / LoD6}
    \end{subfigure}%
    \begin{subfigure}[t]{0.2\linewidth}
        \centering
        \includegraphics[width=\textwidth]{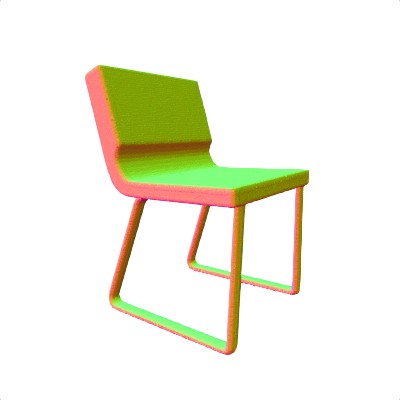}\vspace{-5pt}
        \caption*{GT}
    \end{subfigure}   

    \begin{subfigure}[t]{0.2\linewidth}
        \centering
        \includegraphics[width=\textwidth]{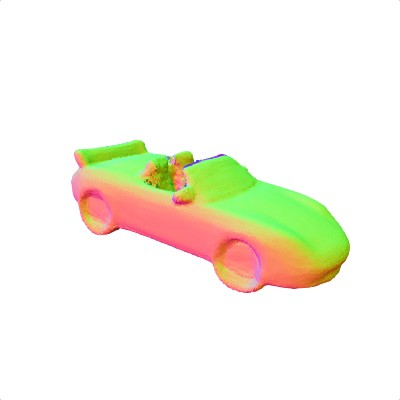}\vspace{-5pt}
        \caption*{DeepSDF}
    \end{subfigure}%
    \begin{subfigure}[t]{0.2\linewidth}
        \centering
        \includegraphics[width=\textwidth]{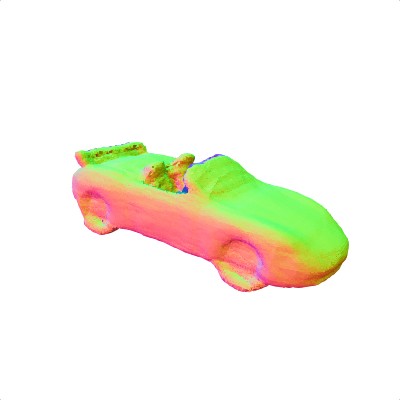}\vspace{-5pt}
        \caption*{Curriculum DeepSDF}
    \end{subfigure}%
    \begin{subfigure}[t]{0.2\linewidth}
        \centering
        \includegraphics[width=\textwidth]{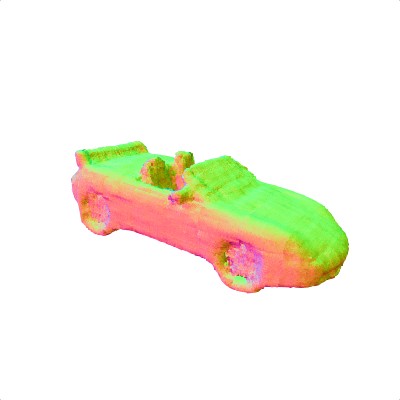}\vspace{-5pt}
        \caption*{ROAD / LoD9}
    \end{subfigure}%
    \begin{subfigure}[t]{0.2\linewidth}
        \centering
        \includegraphics[width=\textwidth]{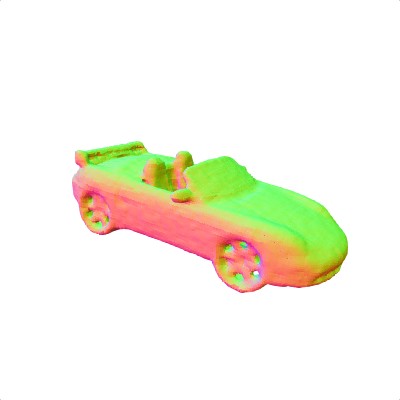}\vspace{-5pt}
        \caption*{Ours / LoD6}
    \end{subfigure}%
    \begin{subfigure}[t]{0.2\linewidth}
        \centering
        \includegraphics[width=\textwidth]{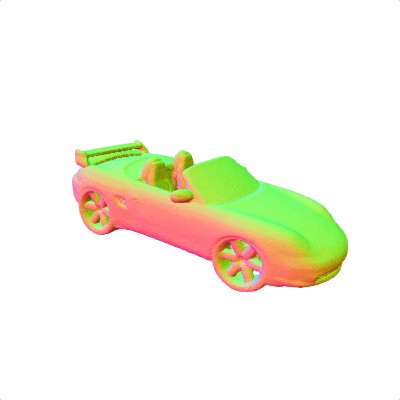}\vspace{-5pt}
        \caption*{GT}
    \end{subfigure}

    \begin{subfigure}[t]{0.2\linewidth}
        \centering
        \includegraphics[width=\textwidth]{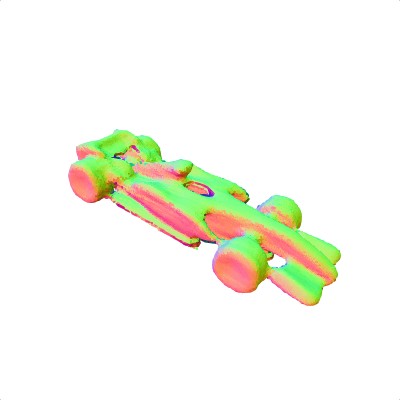}\vspace{-5pt}
        \caption*{DeepSDF}
    \end{subfigure}%
    \begin{subfigure}[t]{0.2\linewidth}
        \centering
        \includegraphics[width=\textwidth]{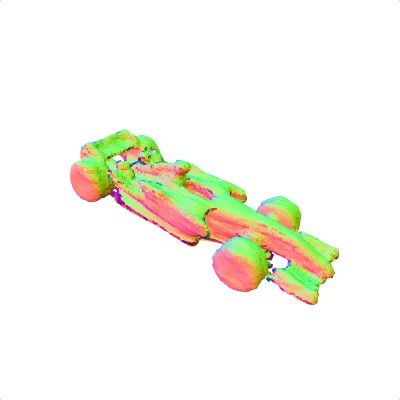}\vspace{-5pt}
        \caption*{Curriculum DeepSDF}
    \end{subfigure}%
    \begin{subfigure}[t]{0.2\linewidth}
        \centering
        \includegraphics[width=\textwidth]{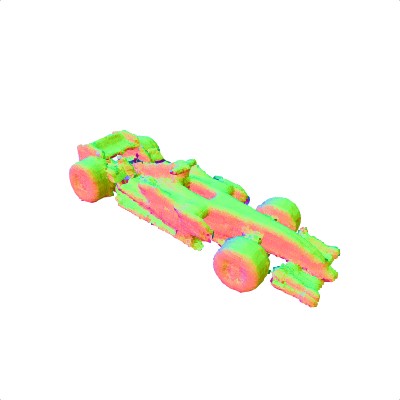}\vspace{-5pt}
        \caption*{ROAD / LoD9}
    \end{subfigure}%
    \begin{subfigure}[t]{0.2\linewidth}
        \centering
        \includegraphics[width=\textwidth]{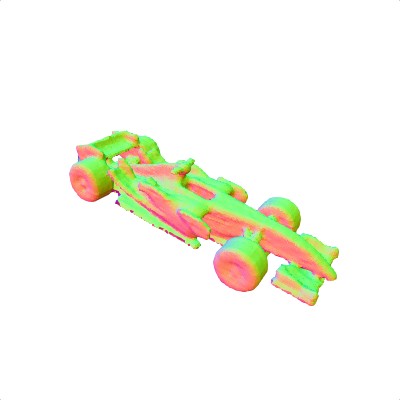}\vspace{-5pt}
        \caption*{Ours / LoD6}
    \end{subfigure}%
    \begin{subfigure}[t]{0.2\linewidth}
        \centering
        \includegraphics[width=\textwidth]{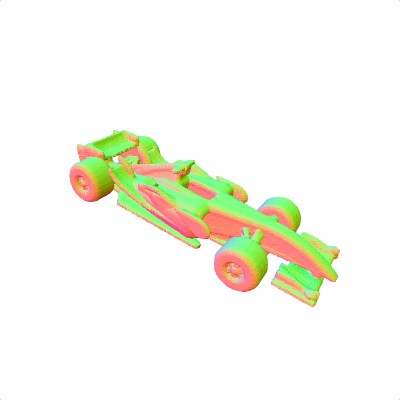}\vspace{-5pt}
        \caption*{GT}
    \end{subfigure}

    \caption{\textbf{Additional ShapeNet150 Results.} Best viewed zoomed in.}
\label{fig:supp_shapenet150}
\vspace{-9pt}
\end{figure*}

\begin{figure*}[t]
    \centering
    \vspace{-5pt}
    
    \begin{subfigure}[t]{0.2\linewidth}
        \centering
        \includegraphics[width=\textwidth]{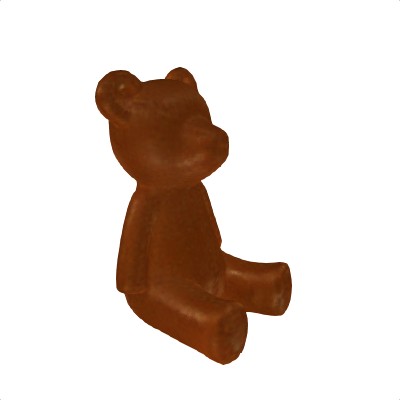}\vspace{-5pt}
    \end{subfigure}%
    \begin{subfigure}[t]{0.2\linewidth}
        \centering
        \includegraphics[width=\textwidth]{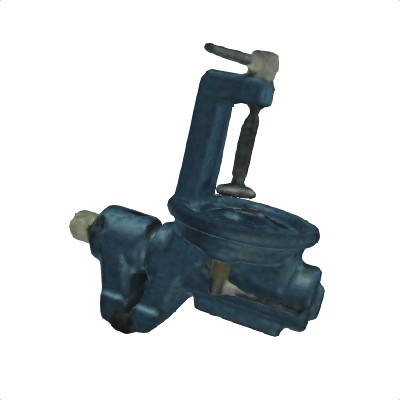}\vspace{-5pt}
    \end{subfigure}%
    \begin{subfigure}[t]{0.2\linewidth}
        \centering
        \includegraphics[width=\textwidth]{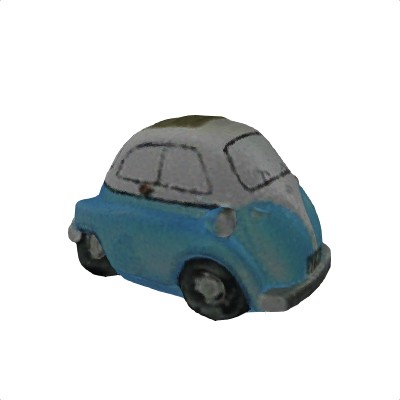}\vspace{-5pt}
    \end{subfigure}%
    \begin{subfigure}[t]{0.2\linewidth}
        \centering
        \includegraphics[width=\textwidth]{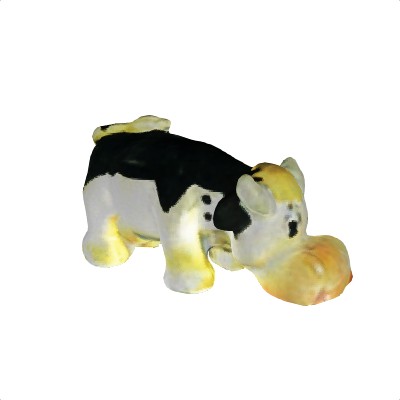}\vspace{-5pt}
    \end{subfigure}%
    \begin{subfigure}[t]{0.2\linewidth}
        \centering
        \includegraphics[width=\textwidth]{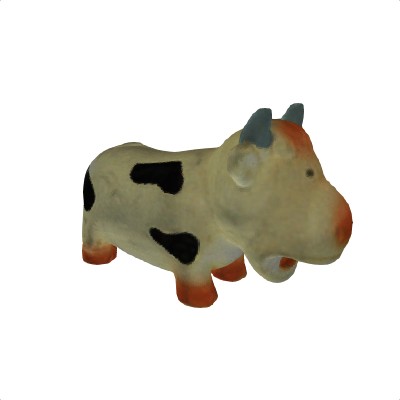}\vspace{-5pt}
    \end{subfigure}

    \begin{subfigure}[t]{0.2\linewidth}
        \centering
        \includegraphics[width=\textwidth]{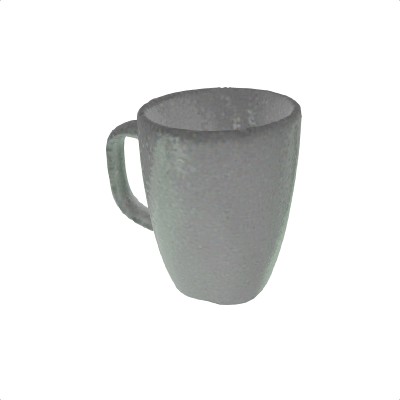}\vspace{-5pt}
    \end{subfigure}%
    \begin{subfigure}[t]{0.2\linewidth}
        \centering
        \includegraphics[width=\textwidth]{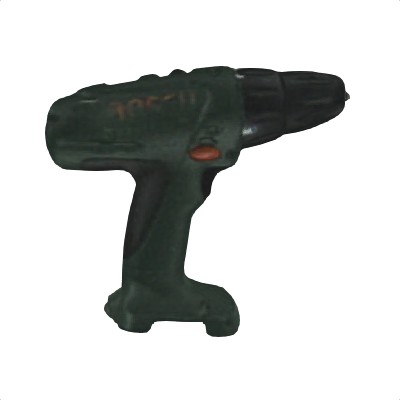}\vspace{-5pt}
    \end{subfigure}%
    \begin{subfigure}[t]{0.2\linewidth}
        \centering
        \includegraphics[width=\textwidth]{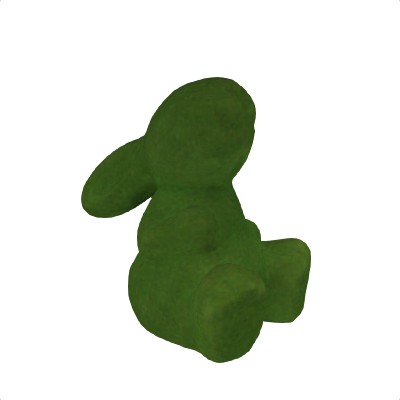}\vspace{-5pt}
    \end{subfigure}%
    \begin{subfigure}[t]{0.2\linewidth}
        \centering
        \includegraphics[width=\textwidth]{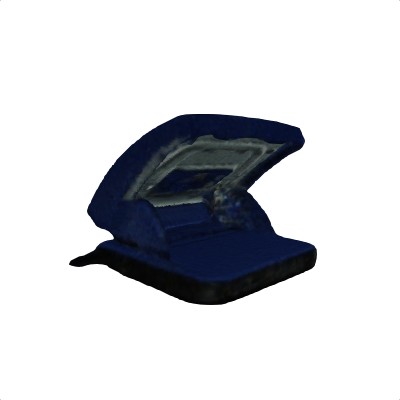}\vspace{-5pt}
    \end{subfigure}%
    \begin{subfigure}[t]{0.2\linewidth}
        \centering
        \includegraphics[width=\textwidth]{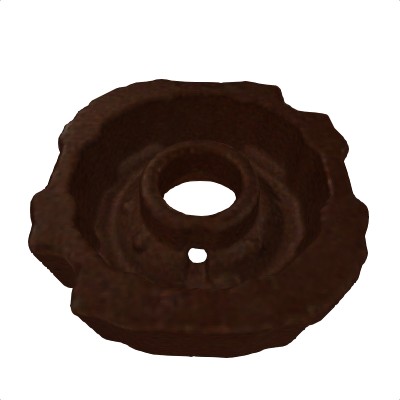}\vspace{-5pt}
    \end{subfigure}

    \begin{subfigure}[t]{0.2\linewidth}
        \centering
        \includegraphics[width=\textwidth]{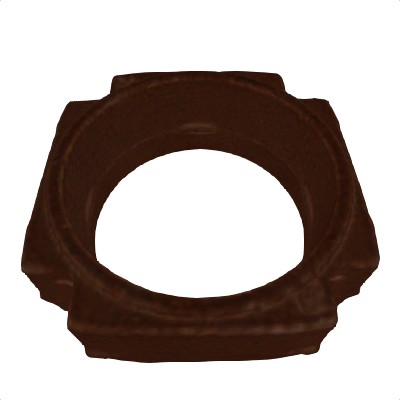}\vspace{-5pt}
    \end{subfigure}%
    \begin{subfigure}[t]{0.2\linewidth}
        \centering
        \includegraphics[width=\textwidth]{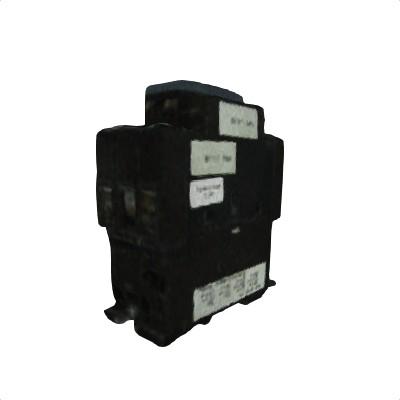}\vspace{-5pt}
    \end{subfigure}%
    \begin{subfigure}[t]{0.2\linewidth}
        \centering
        \includegraphics[width=\textwidth]{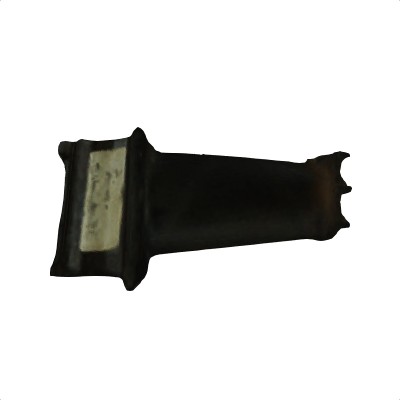}\vspace{-5pt}
    \end{subfigure}%
    \begin{subfigure}[t]{0.2\linewidth}
        \centering
        \includegraphics[width=\textwidth]{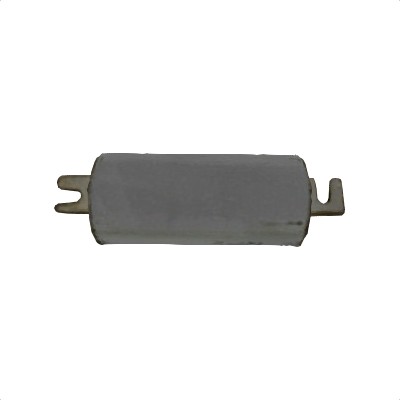}\vspace{-5pt}
    \end{subfigure}%
    \begin{subfigure}[t]{0.2\linewidth}
        \centering
        \includegraphics[width=\textwidth]{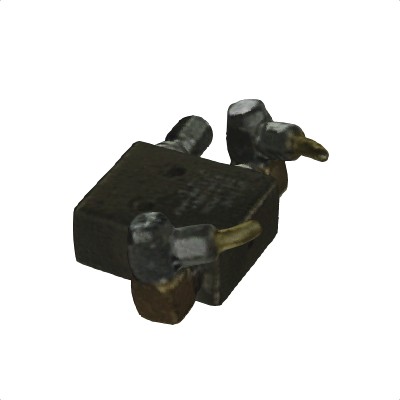}\vspace{-5pt}
    \end{subfigure}

    \begin{subfigure}[t]{0.2\linewidth}
        \centering
        \includegraphics[width=\textwidth]{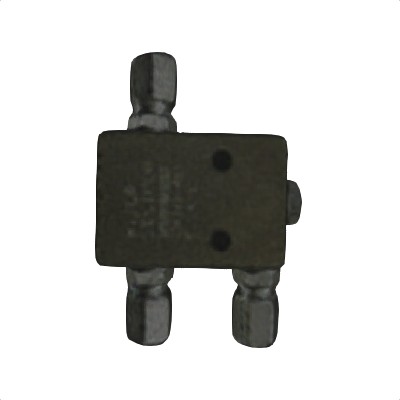}\vspace{-5pt}
    \end{subfigure}%
    \begin{subfigure}[t]{0.2\linewidth}
        \centering
        \includegraphics[width=\textwidth]{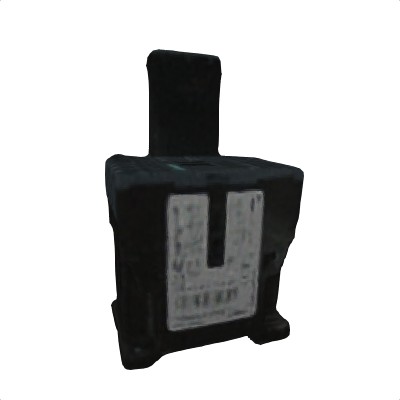}\vspace{-5pt}
    \end{subfigure}%
    \begin{subfigure}[t]{0.2\linewidth}
        \centering
        \includegraphics[width=\textwidth]{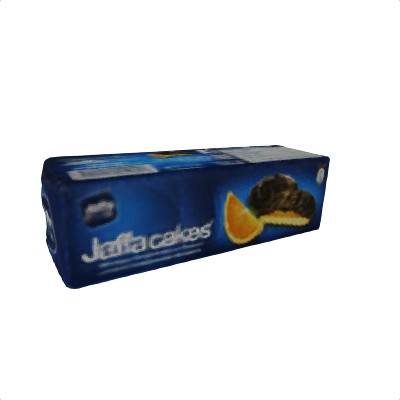}\vspace{-5pt}
    \end{subfigure}%
    \begin{subfigure}[t]{0.2\linewidth}
        \centering
        \includegraphics[width=\textwidth]{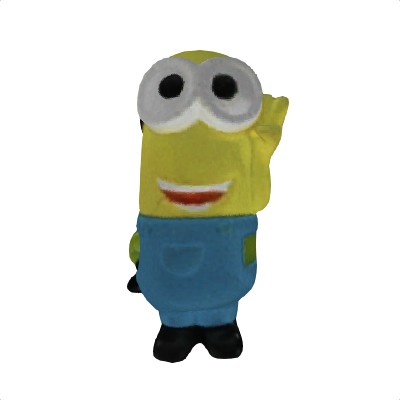}\vspace{-5pt}
    \end{subfigure}%
    \begin{subfigure}[t]{0.2\linewidth}
        \centering
        \includegraphics[width=\textwidth]{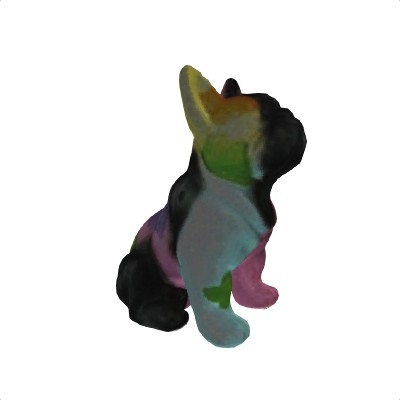}\vspace{-5pt}
    \end{subfigure}

    \begin{subfigure}[t]{0.2\linewidth}
        \centering
        \includegraphics[width=\textwidth]{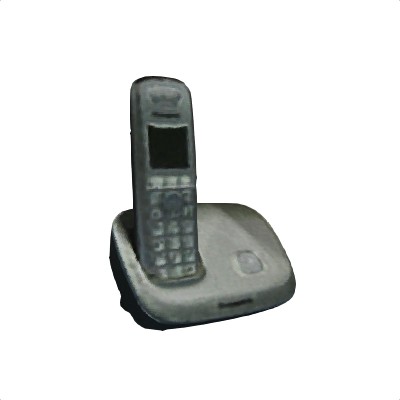}\vspace{-5pt}
    \end{subfigure}%
    \begin{subfigure}[t]{0.2\linewidth}
        \centering
        \includegraphics[width=\textwidth]{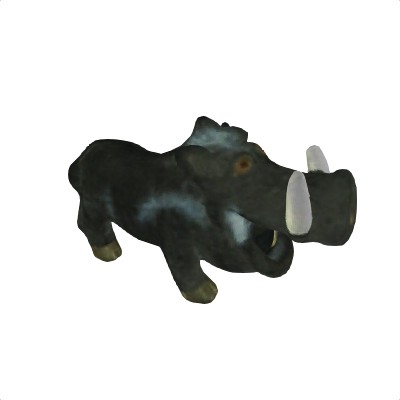}\vspace{-5pt}
    \end{subfigure}%
    \begin{subfigure}[t]{0.2\linewidth}
        \centering
        \includegraphics[width=\textwidth]{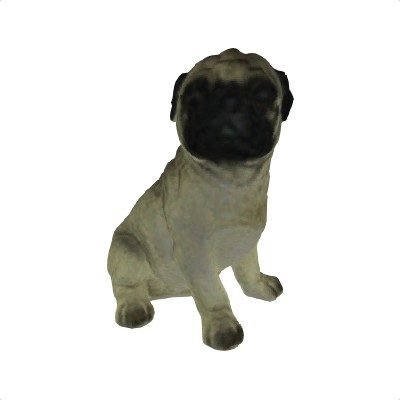}\vspace{-5pt}
    \end{subfigure}%
    \begin{subfigure}[t]{0.2\linewidth}
        \centering
        \includegraphics[width=\textwidth]{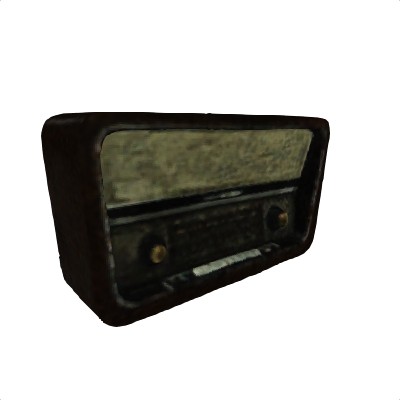}\vspace{-5pt}
    \end{subfigure}%
    \begin{subfigure}[t]{0.2\linewidth}
        \centering
        \includegraphics[width=\textwidth]{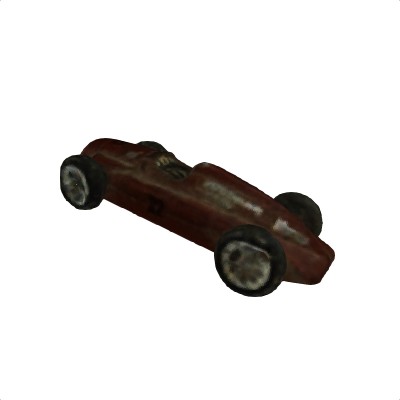}\vspace{-5pt}
    \end{subfigure}

    \begin{subfigure}[t]{0.2\linewidth}
        \centering
        \includegraphics[width=\textwidth]{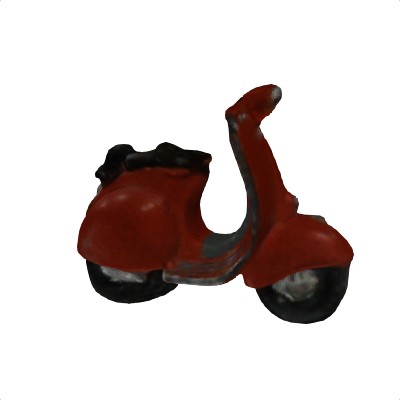}\vspace{-5pt}
    \end{subfigure}%
    \begin{subfigure}[t]{0.2\linewidth}
        \centering
        \includegraphics[width=\textwidth]{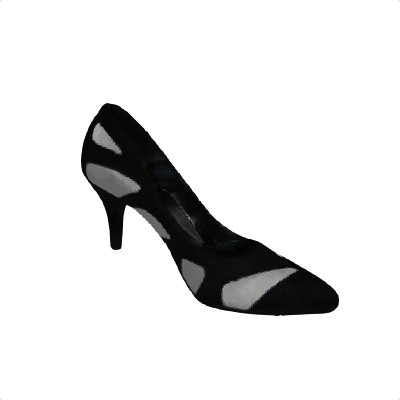}\vspace{-5pt}
    \end{subfigure}%
    \begin{subfigure}[t]{0.2\linewidth}
        \centering
        \includegraphics[width=\textwidth]{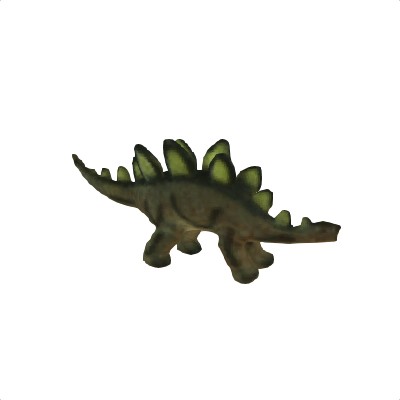}\vspace{-5pt}
    \end{subfigure}%
    \begin{subfigure}[t]{0.2\linewidth}
        \centering
        \includegraphics[width=\textwidth]{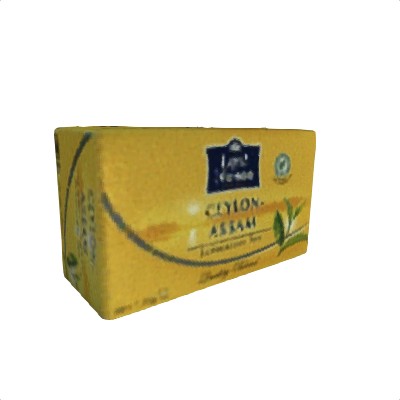}\vspace{-5pt}
    \end{subfigure}%
    \begin{subfigure}[t]{0.2\linewidth}
        \centering
        \includegraphics[width=\textwidth]{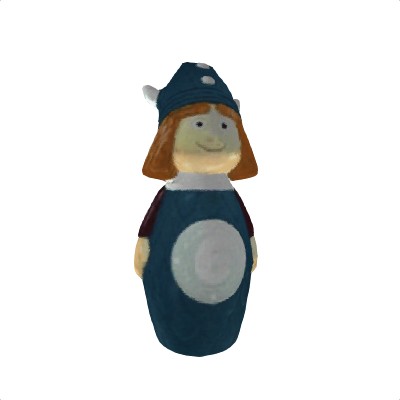}\vspace{-5pt}
    \end{subfigure}

    \caption{\textbf{Additional HomebrewedDB Results.} Best viewed zoomed in.}
\label{fig:supp_hb}
\vspace{-9pt}
\end{figure*}

\begin{figure*}[t]
    \centering
    \vspace{-5pt}
    
    \begin{subfigure}[t]{0.2\linewidth}
        \centering
        \includegraphics[width=\textwidth]{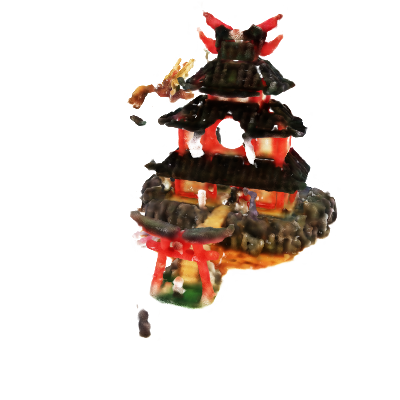}\vspace{-5pt}
        \caption*{Lat 32}
    \end{subfigure}%
    \begin{subfigure}[t]{0.2\linewidth}
        \centering
        \includegraphics[width=\textwidth]{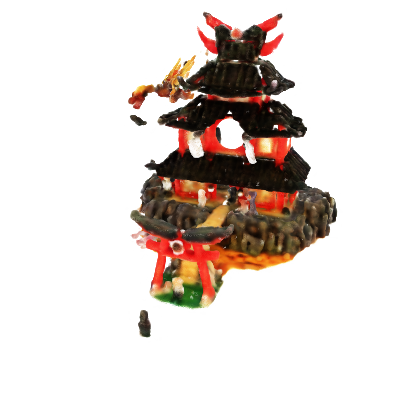}\vspace{-5pt}
        \caption*{Lat 64}
    \end{subfigure}%
    \begin{subfigure}[t]{0.2\linewidth}
        \centering
        \includegraphics[width=\textwidth]{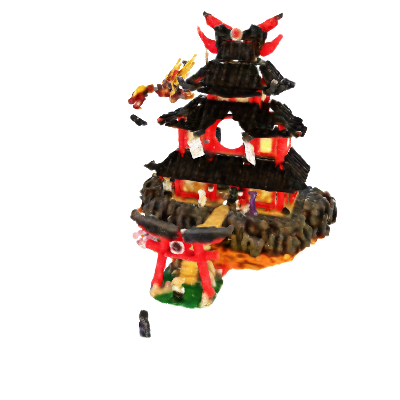}\vspace{-5pt}
        \caption*{Lat 128}
    \end{subfigure}%
    \begin{subfigure}[t]{0.2\linewidth}
        \centering
        \includegraphics[width=\textwidth]{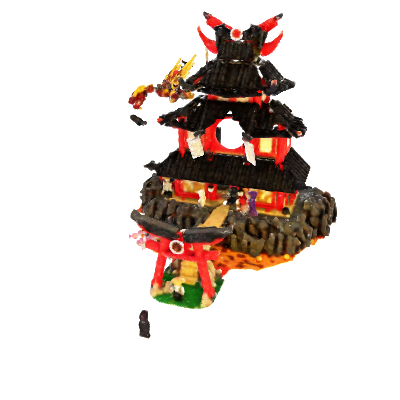}\vspace{-5pt}
        \caption*{Lat 256}
    \end{subfigure}%
    \begin{subfigure}[t]{0.2\linewidth}
        \centering
        \includegraphics[width=\textwidth]{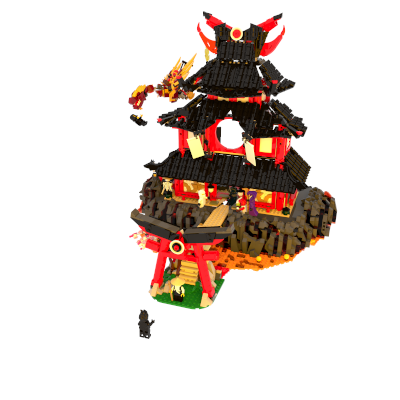}\vspace{-5pt}
        \caption*{GT}
    \end{subfigure}

    \begin{subfigure}[t]{0.2\linewidth}
        \centering
        \includegraphics[width=\textwidth]{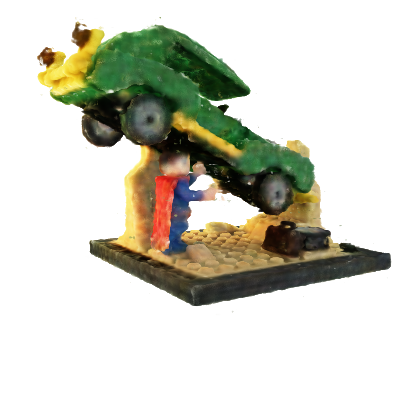}\vspace{-5pt}
        \caption*{Lat 32}
    \end{subfigure}%
    \begin{subfigure}[t]{0.2\linewidth}
        \centering
        \includegraphics[width=\textwidth]{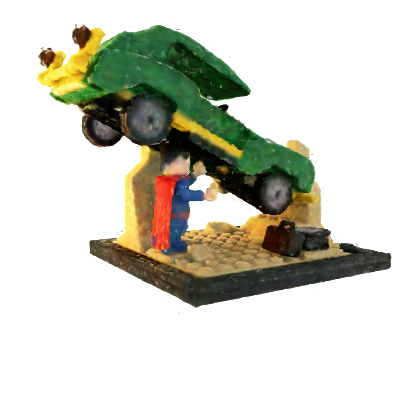}\vspace{-5pt}
        \caption*{Lat 64}
    \end{subfigure}%
    \begin{subfigure}[t]{0.2\linewidth}
        \centering
        \includegraphics[width=\textwidth]{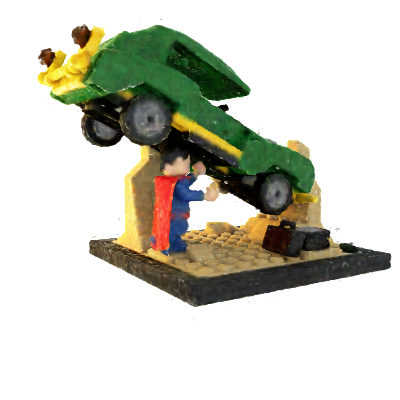}\vspace{-5pt}
        \caption*{Lat 128}
    \end{subfigure}%
    \begin{subfigure}[t]{0.2\linewidth}
        \centering
        \includegraphics[width=\textwidth]{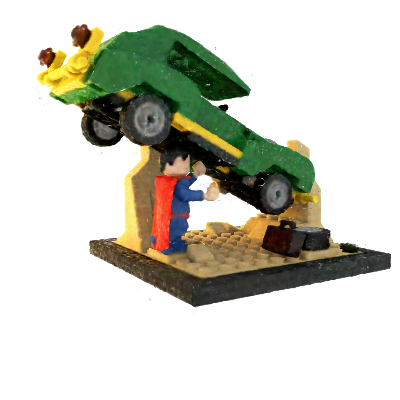}\vspace{-5pt}
        \caption*{Lat 256}
    \end{subfigure}%
    \begin{subfigure}[t]{0.2\linewidth}
        \centering
        \includegraphics[width=\textwidth]{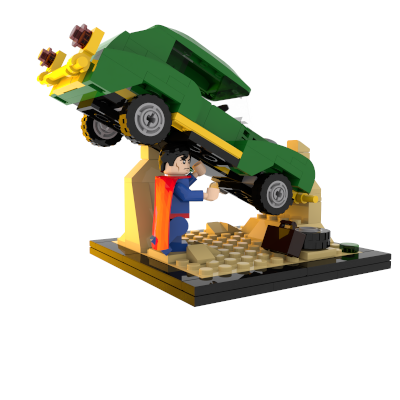}\vspace{-5pt}
        \caption*{GT}
    \end{subfigure}
    
    \begin{subfigure}[t]{0.2\linewidth}
        \centering
        \includegraphics[width=\textwidth]{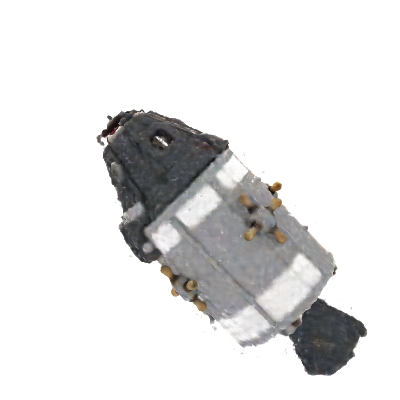}\vspace{-5pt}
        \caption*{Lat 32}
    \end{subfigure}%
    \begin{subfigure}[t]{0.2\linewidth}
        \centering
        \includegraphics[width=\textwidth]{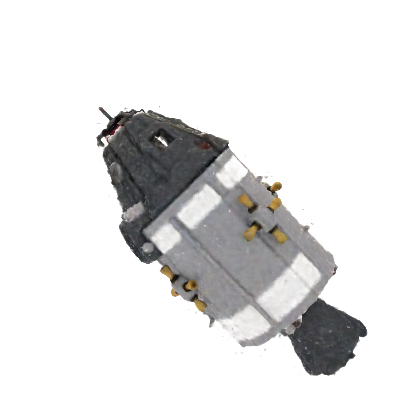}\vspace{-5pt}
        \caption*{Lat 64}
    \end{subfigure}%
    \begin{subfigure}[t]{0.2\linewidth}
        \centering
        \includegraphics[width=\textwidth]{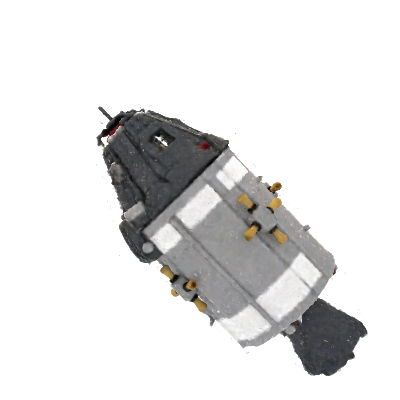}\vspace{-5pt}
        \caption*{Lat 128}
    \end{subfigure}%
    \begin{subfigure}[t]{0.2\linewidth}
        \centering
        \includegraphics[width=\textwidth]{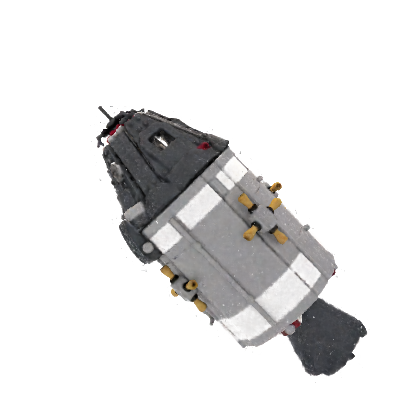}\vspace{-5pt}
        \caption*{Lat 256}
    \end{subfigure}%
    \begin{subfigure}[t]{0.2\linewidth}
        \centering
        \includegraphics[width=\textwidth]{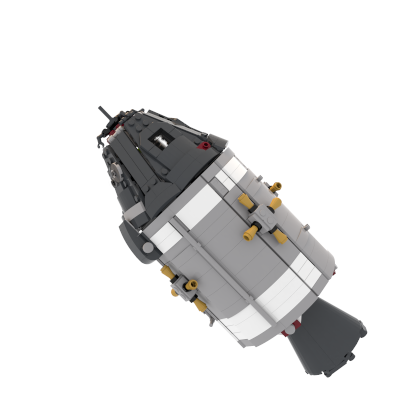}\vspace{-5pt}
        \caption*{GT}
    \end{subfigure}
    
    \begin{subfigure}[t]{0.2\linewidth}
        \centering
        \includegraphics[width=\textwidth]{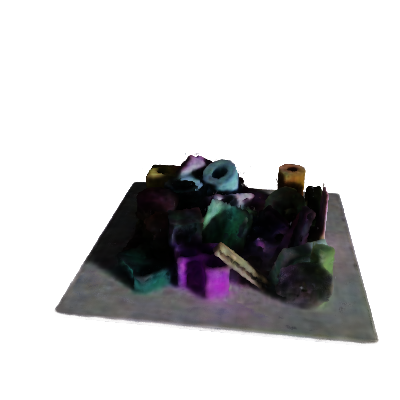}\vspace{-5pt}
        \caption*{Lat 32}
    \end{subfigure}%
    \begin{subfigure}[t]{0.2\linewidth}
        \centering
        \includegraphics[width=\textwidth]{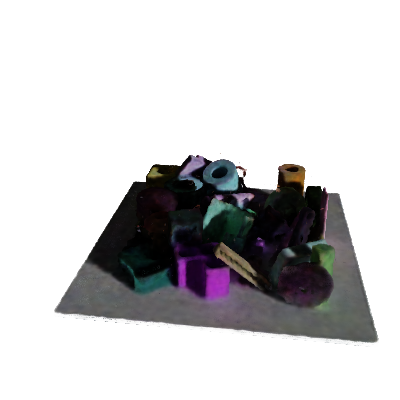}\vspace{-5pt}
        \caption*{Lat 64}
    \end{subfigure}%
    \begin{subfigure}[t]{0.2\linewidth}
        \centering
        \includegraphics[width=\textwidth]{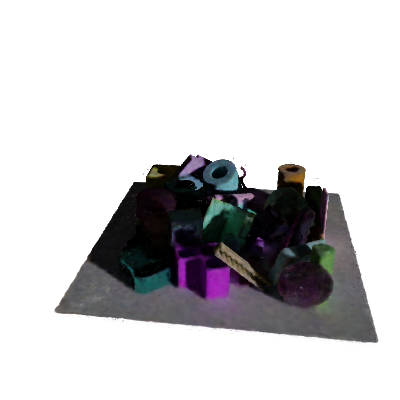}\vspace{-5pt}
        \caption*{Lat 128}
    \end{subfigure}%
    \begin{subfigure}[t]{0.2\linewidth}
        \centering
        \includegraphics[width=\textwidth]{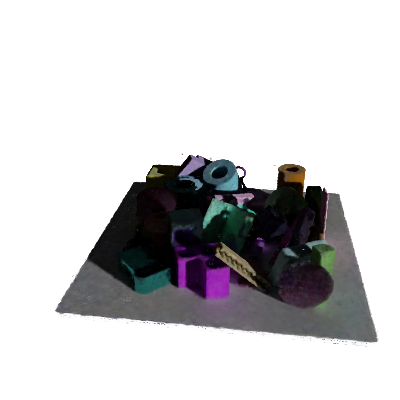}\vspace{-5pt}
        \caption*{Lat 256}
    \end{subfigure}%
    \begin{subfigure}[t]{0.2\linewidth}
        \centering
        \includegraphics[width=\textwidth]{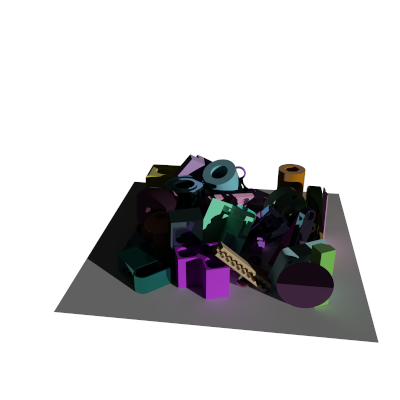}\vspace{-5pt}
        \caption*{GT}
    \end{subfigure}
    
    \begin{subfigure}[t]{0.2\linewidth}
        \centering
        \includegraphics[width=\textwidth]{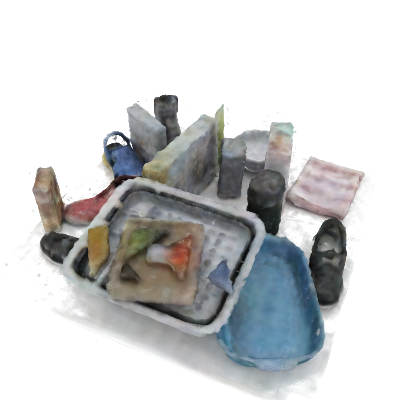}\vspace{-5pt}
        \caption*{Lat 32}
    \end{subfigure}%
    \begin{subfigure}[t]{0.2\linewidth}
        \centering
        \includegraphics[width=\textwidth]{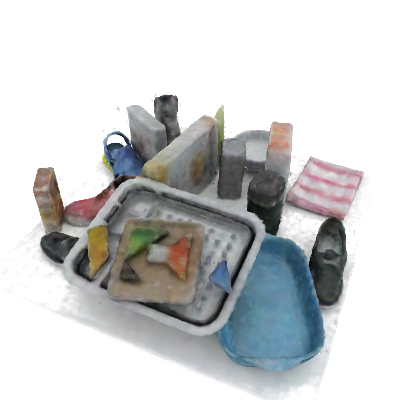}\vspace{-5pt}
        \caption*{Lat 64}
    \end{subfigure}%
    \begin{subfigure}[t]{0.2\linewidth}
        \centering
        \includegraphics[width=\textwidth]{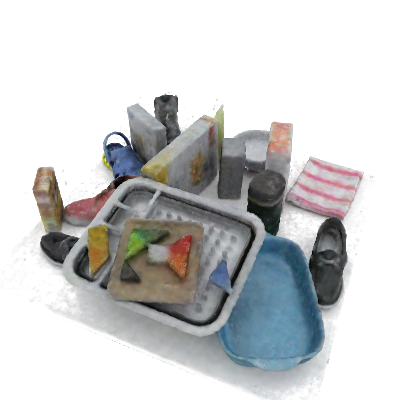}\vspace{-5pt}
        \caption*{Lat 128}
    \end{subfigure}%
    \begin{subfigure}[t]{0.2\linewidth}
        \centering
        \includegraphics[width=\textwidth]{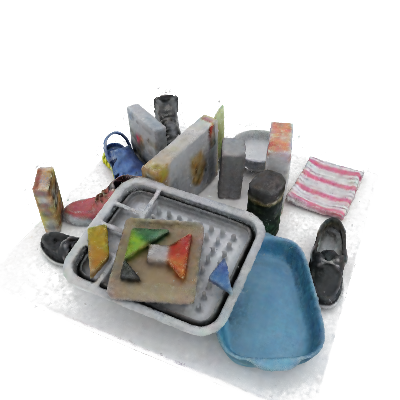}\vspace{-5pt}
        \caption*{Lat 256}
    \end{subfigure}%
    \begin{subfigure}[t]{0.2\linewidth}
        \centering
        \includegraphics[width=\textwidth]{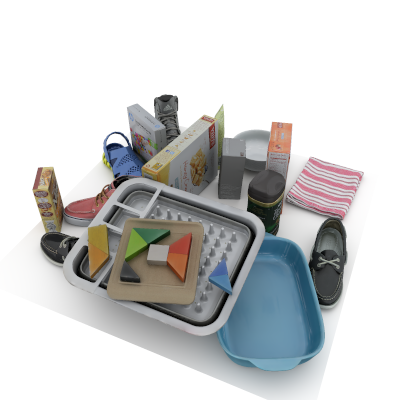}\vspace{-5pt}
        \caption*{GT}
    \end{subfigure}

    \caption{\textbf{Additional RTMV Results.} Best viewed zoomed in.}
\label{fig:supp_rtmv}
\vspace{-9pt}
\end{figure*}

\paragraph{Qualitative Results}
In Figs.~\ref{fig:supp_thingi32} and \ref{fig:supp_shapenet150}, we present additional qualitative results comparing \methodspace against baselines: DeepSDF, Curriculum DeepSDF, and ROAD. We also demonstrate reconstructions of the HomebrewedDB in Fig.~\ref{fig:supp_hb} and additional RTMV qualitative results in Fig.~\ref{fig:supp_rtmv}.

\setlength{\tabcolsep}{8pt}
\begin{table}[t]
  \centering
  \caption{
    \textbf{SIREN vs ReLU MLP.} SIREN-based MLP demonstrates better overall reconstuction results when compared to a ReLU-based MLP.}
  \resizebox{\columnwidth}{!}{
    \begin{tabular}{lccccc}
    \toprule
    \multirow{2}[4]{*}{\textbf{Activation}} & T-Rex (SDF) & \multicolumn{2}{c}{Dog (SDF+RGB)} & \multicolumn{2}{c}{Car (NeRF)} \\
\cmidrule(lr){2-2} \cmidrule(lr){3-4} \cmidrule(lr){5-6}          & CD$\downarrow$    & CD$\downarrow$    & 3D PSNR$\uparrow$ & PSNR$\uparrow$  & SSIM$\uparrow$ \\
    \midrule
    ReLU  & 0.026 & 0.021 & 34.08 & 28.170 & 0.951 \\
    SIREN & 0.025 & 0.020 & 42.16 & 29.130 & 0.962 \\
    \bottomrule
    \end{tabular}%
    }
 \label{tab:siren_relu}
\end{table}%

\section{Applications}
In recent years neural fields have found its use in various domains including robotics and graphics. In recent years, neural fields have found utility in various domains, including robotics and graphics. In robotics, neural fields are actively employed to represent 3D geometry and appearance, with applications in object pose estimation and refinement~\cite{zakharov2020autolabeling,irshad2022shapo}, grasping~\cite{breyer2021volumetric,ichnowski2022dex}, and trajectory planning~\cite{adamkiewicz2022vision}. In graphics, they have been successfully utilized for object reconstruction from sparse and noisy data~\cite{williams2021neural} and for representing high-quality 3D assets~\cite{takikawa2022variable}.

\methodspace employs a recursive hierarchical formulation that leverages object self-similarity, resulting in a highly compressed and efficient shape latent space. We demonstrate that our method achieves impressive results in SDF-based reconstruction (Table 1 of the main paper) and NeRF-based novel-view synthesis (Tables 2 and 3 of the main paper), and features well-clustered latent spaces allowing for smooth interpolation (Figs.~\ref{fig:latent_spaces} and \ref{fig:latent}). We believe that these properties will accelerate the applicability of neural fields in real-world tasks, particularly those involving compression.

% !TEX root = ../supplementary.tex

% !TEX root = ../supplementary.tex

% !TEX root = ../supplementary.tex

% !TEX root = ../supplementary.tex

% \clearpage
% \clearpage

% \bibliographystyle{ACM-Reference-Format}
% \bibliography{egbib}

\end{document}